\documentclass{article} %
\usepackage{iclr2027_conference,times}

\usepackage{amsmath,amsfonts,bm}

\def\eqref#1{equation~\ref{#1}}

\def\1{\bm{1}}

\DeclareMathAlphabet{\mathsfit}{\encodingdefault}{\sfdefault}{m}{sl}
\SetMathAlphabet{\mathsfit}{bold}{\encodingdefault}{\sfdefault}{bx}{n}

\usepackage{hyperref}
\hypersetup{hidelinks}
\usepackage{url}
\usepackage{booktabs}
\usepackage{tikz}
\usepackage{graphicx}
\usepackage{pgfplots}
\usepackage{amsmath}
\usepackage{multirow}
\usepackage{pgfplots}
\usepgfplotslibrary{groupplots}
\usepackage{times}\usepackage{tikz}\usepackage{subcaption}
\pgfplotsset{compat=1.18}
\usetikzlibrary{plotmarks}
\usetikzlibrary{arrows.meta}
\usetikzlibrary{decorations.pathreplacing,positioning}
\pgfdeclareplotmark{cross}{
    \node[red, inner sep=0pt] at (0,0)
    {\large$\boldsymbol{\times}$};
}

\title{Longer Records, Broader Invariance: The Hidden Scaling Problem in Longitudinal Contrastive Learning}

\author{Rameen Mahmood$^{1}$, Xuhai ``Orson'' Xu$^{2}$, Zachary Beattie$^{3}$, Jeffrey Kaye$^{3}$, Danny Yuxing Huang$^{1}$\\
\normalfont $^{1}$New York University\quad $^{2}$Columbia University\quad $^{3}$Oregon Health \& Science University\\
\texttt{\{rameen.mahmood,dhuang\}@nyu.edu, xx2489@cumc.columbia.edu,}\\
\texttt{\{beattiez,kaye\}@ohsu.edu}}

\iclrfinalcopy
\begin{document}

\maketitle
\lhead{Under review as a conference paper at ICLR 2027}

\begin{abstract}
Longitudinal data are valuable because people change. Yet the objectives used
to learn from these data can inadvertently erase that change. In person-level
contrastive learning, observations from the same person are treated as
positives; as records grow, those positives can span increasingly
distant---and increasingly different---behavioral states. More history can
therefore produce not only more data, but broader invariance.
We show that this distinction is fundamental. We separate \emph{record span},
how much history the learner sees, from \emph{supervision span}, how far across
that history positive-pair supervision reaches. Across in-home sensing records spanning up to 2.7 years, broader supervision systematically suppresses recoverable changing-state information, even when the available history is held fixed. At the broadest span, less than 10\% of the information recoverable from an untrained encoder remains. Yet keeping positives local is not sufficient: as records grow, even distant states that are never paired become increasingly similar. Explicitly contrasting other observations from the same person reverses this loss without shortening the record, revealing a second route by which longitudinal scale can broaden invariance. Finally, we prospectively reproduce the supervision-span effect in 199 GLOBEM participants.
Longitudinal scale therefore presents a choice: more history need not
mean more invariance. By controlling what is held invariant as records grow,
we can preserve the change that made the longitudinal data valuable in the
first place.
\end{abstract}

\section{Introduction}

Longitudinal datasets are getting longer. Wearables, ambient sensors, and mobile devices can now accumulate months or years of repeated observations from the same individual \citep{onnela2016harnessing,narayanswamy2025scaling}, creating an appealing new source of scale for self-supervised learning (SSL) \citep{yuan2024self,abbaspourazad2023large,narayanswamy2025scaling}. More data should expose a learner to more of the structure it seeks to model \citep{kaplan2020scaling,hoffmann2022training}. However, longitudinal data have a property that ordinary dataset scaling does not: the dataset grows through time. As more history becomes available, observations from the same person can be separated not only by more samples, but by months or years of behavioral change.

This creates a subtle problem for person-level contrastive learning. Such methods treat observations from the same person as positives \citep{abbaspourazad2023large,diamant2022patient}. In a short record, those observations are typically close in time. In a long record, the same rule can pair states separated by months or seasons. They still share \emph{who} generated them, but may share progressively less of \emph{what is happening now}. Thus, extending the record can silently extend the invariance imposed by the objective: more history becomes not only more data to learn from, but more variation the model may be asked to ignore.

We formalize this distinction through two axes of longitudinal scale. \emph{Record span}, $L$, is the amount of history available to the learner. \emph{Supervision span}, $\tau$, is the temporal reach over which the objective relates observations as positives. Ordinary same-person sampling couples them: as the record grows, increasingly distant states become eligible positives, and $\tau$ grows with $L$. This creates a basic ambiguity in longitudinal scaling. If representations lose changing-state information as records grow, is the additional history itself responsible, or has the supervision silently changed with it?

This question follows from a broader principle in contrastive learning: representations are shaped by what the objective declares equivalent. Positive-pair construction determines which variation should survive \citep{tian2020makes}; imposed invariances can remove features needed downstream \citep{xiao2020should}; and contrastive objectives encourage representations to preserve what paired observations share while suppressing what distinguishes them \citep{wang2020understanding,von2021self,robinson2021can}. Time-series SSL therefore makes consequential choices about temporal relationships. CPC relates observations through future prediction \citep{oord2018representation}; TS2Vec through hierarchical temporal contexts \citep{yue2022ts2vec}; and neighborhood-based methods restrict or weight relationships according to temporal proximity or stationarity \citep{tonekaboni2021tnc,yeche2021neighborhood,lee2024soft}. These methods differ in what they declare invariant, but they share an important assumption: the meaning of that relation does not silently change merely because more data are collected.

Longitudinal same-person learning breaks this assumption. ``Same person'' remains a fixed sampling rule while becoming an increasingly broad temporal relation. The resulting confound is easy to miss because neither the loss nor the sampler needs to change. Only the record grows.

We disentangle these effects in DETECT \citep{detectad2022}, an in-home behavioral sensing study of 57 single-resident participants and 44 multi-resident households with records spanning up to 2.7 years. We first hold record span fixed and vary supervision span. As positive pairs reach farther through time, changing-state information systematically declines, falling below 10\% of the untrained reference at the broadest span. Supervision span is therefore a direct driver of state suppression, distinct from record span itself.

However, keeping supervision local does not solve the problem. As record span grows, changing-state information still declines, and even distant observations that are never paired during training become increasingly similar. Local supervision limits direct alignment across time, but does not guarantee local invariance.

This residual loss reveals a second route to invariance. Person-level contrastive learning separates people, but does not require different states of the same person to remain distinguishable. We test this directly by treating other observations from the same person as negatives. This single change recovers changing-state information to the short-record level without discarding the additional history, while identity remains strongly decodable.\footnote{The recovery does not depend on temporal segmentation: random groupings work equally well, whereas partitioning the record without introducing within-person discrimination does not.} The problem is therefore not long history itself, but an objective that fails to preserve distinctions within a person’s history.

Finally, we test whether the supervision-span effect generalizes to a separate dataset. A prospectively specified test on a later cohort of 199 participants from GLOBEM \citep{xu2022globem}, a mobile phone-sensing study of a different population, reproduces the decline across all five seeds and both prespecified readouts. The effect therefore transfers across population, dataset, and sensing setting.

Together, these experiments reveal two distinct routes to unwanted invariance. Broader supervision directly aligns states across time; longer records create a second route, allowing within-person differences to collapse even when those states are never paired. Importantly, neither is an unavoidable cost of using more history: local positives limit the first, and within-person discrimination reverses the second. Record span determines how much history a learner can use; supervision determines how much it is asked to treat as equivalent. \emph{A year of data need not imply a year of invariance.}

We make three contributions:
\begin{enumerate}
\item We identify \emph{supervision span} as a distinct axis of longitudinal scale, and isolate it experimentally: as records grow, same-person positives can span increasingly distant states, silently broadening the invariance imposed by supervision.

\item We show that \emph{local supervision does not guarantee local invariance}: over long records, person-level contrastive learning induces invariance even between states never paired together. Treating distant same-person states as negatives reverses this loss without shortening the record.

\item We prospectively reproduce the supervision-span effect in a separate dataset, GLOBEM, in 199 participants, demonstrating generalization beyond the discovery setting.
\end{enumerate}

\section{Setup: Measuring Information Beyond Identity}
\label{sec:setup}

\paragraph{Data and prediction problem.}
We study two longitudinal behavioral-sensing cohorts with records spanning up
to 2.7 years: 57 participants living in single-resident homes and 44
multi-resident households.\footnote{The multi-resident data contain 87 subject
records from 44 homes. As co-residents share the same home-level sensor
stream, we retain one record per home and treat the household as the
independent unit. Experiment-specific eligibility yields 51--55
single-resident participants and typically 42 multi-resident households.}
The multi-resident cohort serves as a robustness cohort rather than an
independent-dataset replication. Appendix~\ref{app:cohort} provides cohort
construction and eligibility details.

Each day is represented by 48 behavioral features derived from eight sensing
channels, including motion, steps, sleep, respiration, medication events, and
weight. The data come from DETECT-AD (hereafter DETECT),\footnote{Unless otherwise specified, all analyses use DETECT; GLOBEM is used only for the prospective replication in Section~\ref{sec:globem-transfer}.} an IRB-approved longitudinal
in-home sensing study that uses a continuous home-based
sensing platform \citep{beattie2020cart} to capture everyday behavior over
periods of up to 2.7 years. 
Data are observed at daily resolution; canonical analyses aggregate
consecutive days into non-overlapping 5-day windows
$x_{i,t}\in\mathbb{R}^{48}$, while experiments requiring finer temporal
control operate directly at daily resolution (Appendix~\ref{app:features}).
Aggregation tolerates intermittent sensor dropout: a window is retained when at
least four of its five days meet the daily coverage criterion
(Appendix~\ref{app:retention}).

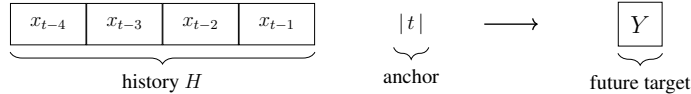
\begin{figure}[!ht]
\centering
\resizebox{0.65\linewidth}{!}{%
\begin{tikzpicture}[
    x=1cm,
    y=1cm,
    every node/.style={font=\large},
    box/.style={
        draw,
        minimum width=1.55cm,
        minimum height=0.85cm,
        inner sep=0pt
    }
]

\node[box] (x4) at (0,0) {$x_{t-4}$};
\node[box, anchor=west] (x3) at (x4.east) {$x_{t-3}$};
\node[box, anchor=west] (x2) at (x3.east) {$x_{t-2}$};
\node[box, anchor=west] (x1) at (x2.east) {$x_{t-1}$};

\draw[
    decorate,
    decoration={brace, amplitude=7pt, mirror, raise=4pt}
]
(x4.south west) -- (x1.south east)
node[midway, below=13pt] {history $H$};

\node (anchor) at (7.45,0) {$\vert\,t\,\vert$};

\draw[
    decorate,
    decoration={brace, amplitude=7pt, mirror, raise=4pt}
]
([xshift=-0.36cm]anchor.south) --
([xshift=0.36cm]anchor.south)
node[midway, below=13pt] {anchor};

\draw[->, line width=0.8pt]
    (8.95,0) -- (10.05,0);

\node[
    draw,
    minimum width=0.9cm,
    minimum height=0.9cm,
    inner sep=0pt,
    font=\Large
] (Y) at (12.15,0) {$Y$};

\draw[
    decorate,
    decoration={brace, amplitude=7pt, mirror, raise=4pt}
]
(Y.south west) -- (Y.south east)
node[midway, below=13pt] {future target};

\end{tikzpicture}%
}
\caption{Prediction setup. History $H$ precedes anchor $t$, and $Y$ is the future target at the specified prediction horizon.}
\label{fig:prediction-setup}
\end{figure}

We frame evaluation as future prediction (Fig.~\ref{fig:prediction-setup}).
For an anchor time $t$, history $H$ precedes the anchor, $z=f(H)$ represents
that history, and $Y$ is the subsequent 5-day behavioral vector. The target is
the window immediately following the history block, so the prediction horizon is
not a separate design choice: it equals the window width, and is one day in the
daily-resolution analyses. Within each experiment, the anchor, future target,
and observation budget are held fixed; only the property of the history under
study is varied.

\paragraph{Measuring changing state beyond identity.}
The central measurement problem is separating \emph{who a person is} from
\emph{how that person is changing}. Longitudinal behavior is strongly
person-specific, so a representation can predict future behavior simply by
encoding a person's stable behavioral profile. High predictive accuracy alone
therefore does not establish that the representation captures changing state.

For participant $i$, let $B_i$ denote mean behavior over the first 60\% of
that participant's windows, providing a fixed person-level reference for typical
behavior. The 60\% fraction is the chronological train/test split point, so
$B_i$ is computed from training-era data alone and never sees the test period.
It is held constant across experiments;
Appendix~\ref{app:trailing-baseline} details its construction and sensitivity.
We then measure how much predictive information the learned representation
$z$ adds beyond this reference by comparing prediction of future behavior
$Y \in \mathbb{R}^D$ from $B_i$ alone with prediction from $B_i$ and $z$:
\[
\Delta I =
\frac{1}{D}\,
\mathbb{E}_{\mathrm{test}}
\left[
\log p_{\hat{\theta}_1}(Y\mid B_i,z)
-
\log p_{\hat{\theta}_0}(Y\mid B_i)
\right].
\]

We report $\Delta I$ in nats per output dimension. Positive $\Delta I$ means
that knowing the representation improves prediction of what the person will do
next beyond knowing what that person is typically like. We use this incremental
predictive information as our operational measure of \emph{changing-state
information}. As $B_i$ is fixed, $\Delta I$ can capture both relatively
local state and slower within-person drift; trailing and recency-matched
baselines preserve the central comparisons
(Appendix~\ref{app:trailing-baseline}).

Formally, $\Delta I$ is conditional predictive $\mathcal{V}$-information
\citep{xu2020theory,hewitt2021conditional}, not Shannon mutual information.
Primary analyses use matched ridge decoders with penalties selected on held-out
validation data. The record-span and supervision-span declines persist with
matched nonlinear MLP decoders, recency-matched baselines, and a residualized
estimator (Appendices~\ref{app:mlp-decoder}, \ref{app:trailing-baseline},
and~\ref{app:estimator}); comparisons against an untrained reference use the
ridge decoder throughout. Solver-selection audits appear in
Appendix~\ref{app:solver-audit}. Evaluation protocol and estimator are specified in
Appendix~\ref{app:protocol}, and architectures, optimization and sampling in
Appendix~\ref{app:implementation}.

\paragraph{Evaluation.}
We split each participant or household chronologically, with training preceding
testing and a buffer preventing histories or prediction targets from crossing
the split. All experimental arms are evaluated on identical anchors.
Uncertainty is estimated by hierarchical resampling over training seeds and
independent units---participants in the single-resident cohort and households
in the multi-resident cohort.
We call an effect consistent across cohorts only when its sign agrees
and its confidence interval excludes zero in both. Leakage checks, controls,
and complete statistical details appear in Appendices~\ref{app:audit}
and~\ref{app:statistics}.

\section{Longitudinal History Is Not Temporally Exchangeable}
\label{non-exchangeable}

Longitudinal learning confronts a fundamental asymmetry: identity persists while behavioral state changes. Two observations months apart may still be unmistakably from the same person while saying very different things about that person's current state. This separation is what makes longer records consequential for person-level self-supervision. As history grows, ``same person'' remains a valid identity relation while spanning an increasingly diverse range of behavioral states. Before asking what a learning objective does across that growing span, we establish that this identity--state separation is already present in the longitudinal record itself.

\subsection{The same history becomes stale}

We first ask whether equally sized histories remain equally informative about what happens next. We hold the anchor, future target, and observation budget fixed and change only \emph{when} the history was observed. Throughout this section, we measure the information preserved by a 32-dimensional bottleneck MLP trained to predict the next 5-day window from the history block; the prediction head is discarded and $z$ denotes the bottleneck representation. Section~\ref{sec:longer-records} then asks what happens when representations are learned through person-level self-supervision.

Every condition contains the same four 5-day windows (20 days total). The history either immediately precedes the anchor (\texttt{recent}), forms a consecutive block sampled earlier in the record (\texttt{shifted}), consists of four windows sampled elsewhere in the available past (\texttt{random}), or comes from the beginning of the record (\texttt{distant}).\footnote{The distant condition uses the study-onset block and is therefore fixed within an observational unit; the anchor-relative distance sweep in Section~\ref{sec:state-decay} separately tests temporal staleness using moving histories.} The shifted block ends at least 20 days before the recent block and is typically much farther away (median 140 days; 10th--90th percentile, 25--420 days).

\begin{table}[!ht]
\centering
\footnotesize
\begin{tabular}{lcc}
\toprule
History source & Single-resident & Households \\
\midrule
Recent ($t-1,\dots,t-4$)      & $0.2818 \pm 0.0054$ & $0.2325 \pm 0.0097$ \\
Shifted (disjoint block)      & $0.0447 \pm 0.0291$ & $0.0158 \pm 0.0043$ \\
Random (four past windows)    & $0.0374 \pm 0.0169$ & $0.0261 \pm 0.0045$ \\
Distant (earliest four windows)& $0.0051 \pm 0.0143$ & $0.0051 \pm 0.0058$ \\
\bottomrule
\end{tabular}
\caption{Usable predictive information ($\Delta I$, nats/dim) by temporal location of history.
Values are mean $\pm$ standard deviation across five seeds. The households column reports the
44 deduplicated household streams, consistent with the observational unit defined in
Appendix~\ref{app:cohort}.}
\label{tab:history-source}
\end{table}

As Table~\ref{tab:history-source} shows, the same amount of history carries dramatically different information depending on when it was observed. Recent history contains \(0.282\) and \(0.233\) nats/dim of changing-state information in the single- and multi-resident cohorts, whereas the same 20-day budget drawn elsewhere contains at most \(0.045\). Older histories carry \(84\)--\(98\%\) less changing-state information than recent history, consistently across both cohorts and all five seeds.

Notably, this recency advantage is not created by the representation. When we bypass the encoder and predict directly from raw history, recent observations remain far more informative about the future than equally sized older histories ($0.344$ and $0.256$ nats/dim in the two cohorts). The temporal asymmetry is therefore already present in the underlying behavior: representation learning can preserve or discard it, but does not create it. Thus, equal amounts of history from the same person need not carry equal information about what comes next.\footnote{The result persists with motion-only features and sequence-aware GRU encoders; leakage and permuted-target audits find no evidence of an evaluation artifact (Appendix~\ref{app:leakage}; Appendix~\ref{app:gru}).}

\subsection{State fades; identity persists}
\label{sec:state-decay}

Older history carries less information about current state. The key question is whether identity fades with it: if not, same-person positives can remain valid even as their states diverge. We therefore track both over temporal distance.

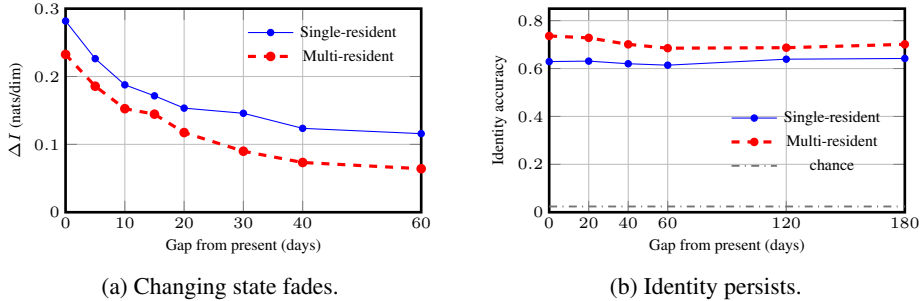
\begin{figure}[!ht]
\centering
\tiny

\begin{subfigure}[t]{0.45\linewidth}
\centering
\begin{tikzpicture}
\begin{axis}[
    width=\linewidth,
    height=0.68\linewidth,
    xlabel={Gap from present (days)},
    ylabel={$\Delta I$ (nats/dim)},
    xmin=0, xmax=60,
    ymin=0, ymax=0.30,
    xtick={0,10,20,30,40,60},
    ytick={0,0.10,0.20,0.30},
    grid=major,
    axis line style={line width=1pt},
    legend style={
        at={(0.97,0.97)},
        anchor=north east,
        draw=none,
        fill=none
    },
]

\addplot[
    blue,
    solid,
    mark=*,
    mark size=1.2pt,
    mark options={
        fill=blue,
        draw=blue
    }
] coordinates {
    (0,0.2818)
    (5,0.2263)
    (10,0.1877)
    (15,0.1715)
    (20,0.1534)
    (30,0.1458)
    (40,0.1236)
    (60,0.1158)
};
\addlegendentry{Single-resident}

\addplot[
    red,
    dashed,
    line width=1.2pt,
    mark=*,
    mark size=1.2pt,
    mark options={
        solid,
        fill=red,
        draw=red
    }
] coordinates {
    (0,0.2325)
    (5,0.1857)
    (10,0.1526)
    (15,0.1445)
    (20,0.1174)
    (30,0.0899)
    (40,0.0733)
    (60,0.0641)
};
\addlegendentry{Multi-resident}

\end{axis}
\end{tikzpicture}
\caption{Changing state fades.}
\label{fig:state-fades}
\end{subfigure}%
\hspace{0.01\linewidth}%
\begin{subfigure}[t]{0.45\linewidth}
\centering
\begin{tikzpicture}
\begin{axis}[
    width=\linewidth,
    height=0.68\linewidth,
    xlabel={Gap from present (days)},
    ylabel={Identity accuracy},
    xmin=0, xmax=180,
    ymin=0, ymax=0.85,
    xtick={0,20,40,60,120,180},
    ytick={0,0.2,0.4,0.6,0.8},
    grid=major,
    axis line style={line width=1pt},
    legend style={
        at={(0.97,0.55)},
        anchor=north east,
        draw=none,
        fill=none
    },
]

\addplot[
    blue,
    solid,
    mark=*,
    mark size=1.2pt,
    mark options={
        fill=blue,
        draw=blue
    }
] coordinates {
    (0,0.629)
    (20,0.631)
    (40,0.620)
    (60,0.614)
    (120,0.639)
    (180,0.642)
};
\addlegendentry{Single-resident}

\addplot[
    red,
    dashed,
    line width=1.2pt,
    mark=*,
    mark size=1pt,
    mark options={
        solid,
        fill=red,
        draw=red
    }
] coordinates {
    (0,0.736)
    (20,0.728)
    (40,0.701)
    (60,0.685)
    (120,0.687)
    (180,0.701)
};
\addlegendentry{Multi-resident}

\addplot[
    black!55,
    dashdotted,
    line width=0.7pt
] coordinates {
    (0,0.024)
    (180,0.024)
};
\addlegendentry{chance}

\end{axis}
\end{tikzpicture}
\caption{Identity persists.}
\label{fig:identity-persists}
\end{subfigure}

\caption{Changing state fades while identity persists with temporal
distance. (a) Usable predictive information about the next window
decreases as the same four-window history block is moved farther from the
present. (b) Identity decoded from displaced histories remains stable
over substantially longer separations.}
\label{fig:state-vs-identity}
\end{figure}

The two quantities evolve on strikingly different timescales
(Fig.~\ref{fig:state-vs-identity}). Changing-state information falls rapidly:
over 60 days, $\Delta I$ declines from $0.282$ to $0.116$ nats/dim in the
single-resident cohort and from $0.233$ to $0.064$ in the multi-resident
cohort---losses of $59\%$ and $72\%$ (Fig.~\ref{fig:state-vs-identity}a).

Identity, by contrast, barely moves over substantially longer separations
(Fig.~\ref{fig:state-vs-identity}b). Decoded from the same anchor-relative raw
history, identity accuracy remains nearly unchanged even at 180 days: $0.629$
stays within $0.614$--$0.642$ in the single-resident cohort, while $0.736$
falls only to $0.701$ in the multi-resident cohort. Observations months apart can therefore remain excellent
evidence of \emph{who} generated them long after becoming weak evidence of
\emph{how that person is now}.\footnote{This separation is robust to decoder choice: linear
and nonlinear identity probes show the same persistence with temporal distance
(Appendix~\ref{app:mlp-decoder}), while sequence-aware GRU analyses preserve
the decline in changing-state information (Appendix~\ref{app:gru}). Identity
here is decoded directly from raw history and is not directly comparable to
identity probes of learned representations in
Section~\ref{sec:longer-records}.}

This raises a natural question: does state information fade simply with elapsed time, or with the behavioral change that occurs across it? The latter appears to matter. When a person's current behavior differs more from their own past, recent history becomes correspondingly more informative than older history. Strikingly, this relationship appears not only across observational units but also \emph{within the same individual over time} in the single-resident cohort (Appendix~\ref{app:drift}). Temporal distance therefore tracks a meaningful change in state: as behavior drifts, older history becomes less informative about what the person is doing now.

\subsection{Recent state lives partly in temporal order}

The information that fades is also not simply a collection of recent measurements. Some of it lies in how behavior unfolds through time. To isolate this structure, we hold the observations themselves fixed and destroy only their order. The \texttt{ordered} condition receives the true sequence. The \texttt{shuffled} condition permutes the order of the same non-overlapping 5-day windows, preserving their contents while destroying order across windows: the same $K$ windows enter the encoder with the same values, and only their arrangement changes ($K{=}4$ for a 20-day history, $K{=}8$ for 40 days). The \texttt{set} condition is an order-free control, a permutation-invariant encoder that cannot represent order at all, so it bounds what any single shuffle could achieve and shows the effect is not an artifact of the particular permutation drawn.

\begin{table}[!ht]
\centering
\footnotesize

\begin{tabular}{llcccc}
\toprule
Cohort & History & Ordered & Shuffled & Set
& Ordered $-$ Shuffled \\
\midrule
Single-resident
& 20d ($K=4$)
& 0.2877 & 0.2469 & 0.2454
& +0.0408 [0.0252, 0.0576] \\

Single-resident
& 40d ($K=8$)
& 0.2773 & 0.2312 & 0.2098
& +0.0461 [0.0285, 0.0649] \\

Multi-resident
& 20d ($K=4$)
& 0.2480 & 0.2252 & 0.2253
& +0.0228 [0.0142, 0.0313] \\

Multi-resident
& 40d ($K=8$)
& 0.2421 & 0.2055 & 0.2064
& +0.0365 [0.0249, 0.0485] \\
\bottomrule
\end{tabular}

\caption{Effect of temporal order on usable predictive information. All arms receive the same recent windows at matched parameter budget. Brackets show 95\% bootstrap confidence intervals for the ordered$-$shuffled difference; all comparisons hold across 5/5 seeds.}
\label{tab:temporal-order}

\end{table}

As Table~\ref{tab:temporal-order} shows, shuffling reduces changing-state
information by $0.023$--$0.046$ nats/dim across both cohorts and history
lengths, a relative loss of $9$--$17\%$. A meaningful fraction of recent-state
information therefore resides in temporal order rather than merely in which
observations are present. The set representation closely tracks the shuffled condition,
showing that the loss comes from removing temporal structure rather than from
the particular permutation mechanism. At daily resolution, the
ordered--shuffled advantage is $1.6$--$3\times$ larger
(Appendix~\ref{app:order-daily}).

Together, these results establish the asymmetry that makes longitudinal person-level learning fundamentally different from ordinary dataset scaling. Identity persists across time; behavioral state does not. As records grow, the learner therefore encounters observations that become increasingly separated in behavioral state while remaining valid matches under a same-person objective. The additional history does not merely provide more examples of the same underlying relation---it expands the range of behavioral variation contained within that relation.

This is where the learning problem begins. A person-level objective must decide what to preserve when the same identity spans many different states. Does the representation change simply because more history becomes available, or because supervision now reaches across more of that variation? We answer this by separating \emph{record span} from \emph{supervision span}.
\section{Longitudinal Scale Broadens Learned Invariance}
\label{sec:longer-records}

We separate record span from supervision span to ask how each changes what the representation preserves (Fig.~\ref{fig:longitudinal-scale}). Under ordinary same-person sampling, the two grow together: longer records provide more history, but also make increasingly distant states eligible as positive pairs (Fig.~\ref{fig:longitudinal-scale}a). We disentangle these effects by keeping supervision local as record span grows (Fig.~\ref{fig:longitudinal-scale}b), holding record span fixed while broadening supervision (Fig.~\ref{fig:longitudinal-scale}c), and finally introducing within-person discrimination to recover the state information that remains lost (Fig.~\ref{fig:longitudinal-scale}d).

\definecolor{spanRed}{HTML}{C0392B}
\definecolor{spanBlue}{HTML}{1F6FB4}
\definecolor{spanGreen}{HTML}{238B45}
\definecolor{spanInk}{HTML}{1A1A1A}
\definecolor{spanGray}{HTML}{666666}
\definecolor{spanEdge}{HTML}{B9B9B9}
\definecolor{spanWin}{HTML}{FAFAFA}

\tikzset{
  win/.style={
    draw=spanEdge,
    fill=spanWin,
    line width=.25pt
  },
  anchorwin/.style={
    draw=spanInk,
    fill=spanInk,
    line width=.25pt
  },
  flabel/.style={
    font=\scriptsize,
    text=spanGray
  },
  ptitle/.style={
    font=\scriptsize,
    align=center,
    text=spanInk
  },
  outcome/.style={
    font=\scriptsize,
    align=center,
    text=spanInk
  }
}

\newcommand{\BW}{0.18}
\newcommand{\BH}{0.27}
\newcommand{\BG}{0.025}

\newcommand{\HStrip}[4]{%
  \foreach \i in {1,...,#3}{%
    \pgfmathsetmacro{\xx}{#1+(\i-1)*(\BW+\BG)}%
    \ifnum\i=#3\relax
      \draw[anchorwin]
        (\xx,#2) rectangle ++(\BW,\BH);
    \else
      \draw[win]
        (\xx,#2) rectangle ++(\BW,\BH);
    \fi
    \coordinate (#4-\i)
      at (\xx+.5*\BW,#2+\BH);
  }%
}

\newcommand{\PArc}[4][0.28]{%
  \draw[#4,line width=.7pt,-{Stealth[length=1.9pt]}]
    (#2) .. controls +(0,#1) and +(0,#1) .. (#3);
}

\begin{figure*}[!ht]
\centering

\resizebox{\textwidth}{!}{%
\begin{tikzpicture}[x=1cm,y=1cm]

\begin{scope}[shift={(0,0)}]

\node[ptitle,text width=3.5cm] at (1.75,4.40)
  {(a) Coupled scaling\\
   {\normalfont $L\uparrow,\;\tau\uparrow$}};

\node[flabel,anchor=east] at (.64,3.55) {$L=30$ d};

\HStrip{.80}{3.41}{4}{aS}

\PArc[.18]{aS-4}{aS-2}{spanRed}
\PArc[.13]{aS-4}{aS-3}{spanRed}

\draw[spanRed,line width=.55pt]
  (.80,3.13) -- (1.58,3.13);
\draw[spanRed,line width=.55pt]
  (.80,3.07) -- (.80,3.19);
\draw[spanRed,line width=.55pt]
  (1.58,3.07) -- (1.58,3.19);

\node[flabel] at (1.19,2.93)
  {$\tau\approx20$ d};

\node[flabel,anchor=east] at (.64,2.36) {$L=$ full};

\HStrip{.80}{2.22}{14}{aL}

\PArc[.34]{aL-14}{aL-2}{spanRed}
\PArc[.27]{aL-14}{aL-5}{spanRed}
\PArc[.21]{aL-14}{aL-8}{spanRed}
\PArc[.15]{aL-14}{aL-11}{spanRed}

\draw[spanRed,line width=.55pt]
  (.80,1.94) -- (3.44,1.94);
\draw[spanRed,line width=.55pt]
  (.80,1.88) -- (.80,2.00);
\draw[spanRed,line width=.55pt]
  (3.44,1.88) -- (3.44,2.00);

\node[flabel] at (2.12,1.74)
  {$\tau\approx245$ d};

\node[outcome] at (2.02,1.22)
  {State $\downarrow$\qquad Identity $\uparrow$};

\end{scope}

\begin{scope}[shift={(4.50,0)}]

\node[ptitle,text width=3.5cm] at (1.75,4.40)
  {(b) Record span\\
   {\normalfont $L\uparrow,\;\tau$ local}};

\node[flabel,anchor=east] at (.64,3.55) {$L=30$ d};

\HStrip{.80}{3.41}{4}{bS}

\PArc[.18]{bS-4}{bS-2}{spanBlue}
\PArc[.13]{bS-4}{bS-3}{spanBlue}

\draw[spanBlue,line width=.55pt]
  (.80,3.13) -- (1.58,3.13);
\draw[spanBlue,line width=.55pt]
  (.80,3.07) -- (.80,3.19);
\draw[spanBlue,line width=.55pt]
  (1.58,3.07) -- (1.58,3.19);

\node[flabel] at (1.19,2.93)
  {$\tau\approx20$ d};

\node[flabel,anchor=east] at (.64,2.36) {$L=$ full};

\HStrip{.80}{2.22}{14}{bL}

\PArc[.18]{bL-14}{bL-12}{spanBlue}
\PArc[.13]{bL-14}{bL-13}{spanBlue}

\draw[spanBlue,line width=.55pt]
  (2.86,1.94) -- (3.44,1.94);
\draw[spanBlue,line width=.55pt]
  (2.86,1.88) -- (2.86,2.00);
\draw[spanBlue,line width=.55pt]
  (3.44,1.88) -- (3.44,2.00);

\node[flabel] at (3.15,1.74)
  {$\tau\approx20$ d};

\node[outcome] at (2.02,1.22)
  {State $\downarrow$\qquad Identity $\uparrow$};

\end{scope}

\begin{scope}[shift={(8.75,0)}]

\node[ptitle,text width=3.5cm] at (1.75,4.40)
  {(c) Supervision span\\
   {\normalfont $L$ fixed,\;$\tau\uparrow$}};

\node[flabel,anchor=east] at (.70,3.58)
  {$\tau=2$ d};

\HStrip{.84}{3.44}{13}{c1}

\PArc[.12]{c1-13}{c1-12}{spanGreen}
\PArc[.16]{c1-13}{c1-11}{spanGreen}

\node[flabel,anchor=east] at (.70,2.83)
  {$\tau=20$ d};

\HStrip{.84}{2.69}{13}{c2}

\PArc[.16]{c2-13}{c2-10}{spanGreen}
\PArc[.20]{c2-13}{c2-7}{spanGreen}
\PArc[.13]{c2-7}{c2-5}{spanGreen}

\node[flabel,anchor=east] at (.70,2.08)
  {$\tau=128$ d};

\HStrip{.84}{1.94}{13}{c3}

\PArc[.30]{c3-13}{c3-2}{spanGreen}
\PArc[.24]{c3-13}{c3-5}{spanGreen}
\PArc[.17]{c3-13}{c3-8}{spanGreen}

\node[outcome] at (2.02,1.22)
  {State $\downarrow$ as $\tau\uparrow$};

\end{scope}

\begin{scope}[shift={(12.60,0)}]

\node[ptitle,text width=3.5cm] at (1.75,4.40)
  {(d) Within-person\\discrimination};

\draw[spanGray,line width=.45pt,<->]
  (.55,3.52) -- (3.40,3.52);

\node[flabel] at (1.98,3.74)
  {same person over time};

\foreach \i in {0,...,2}{
  \pgfmathsetmacro{\xx}{.58+\i*.34}
  \draw[
    draw=spanBlue!65,
    fill=spanBlue!8,
    line width=.35pt
  ]
    (\xx,2.95) rectangle ++(.29,.34);
}

\foreach \i in {0,...,3}{
  \pgfmathsetmacro{\xx}{1.72+\i*.34}
  \draw[
    draw=spanRed!60,
    fill=spanRed!7,
    line width=.35pt
  ]
    (\xx,2.95) rectangle ++(.29,.34);
}

\foreach \i in {0,...,2}{
  \pgfmathsetmacro{\xx}{3.20+\i*.34}
  \draw[
    draw=spanGreen!60,
    fill=spanGreen!7,
    line width=.35pt
  ]
    (\xx,2.95) rectangle ++(.29,.34);
}

\node[font=\scriptsize] at (4.45,3.12)
  {$\cdots$};

\draw[
  spanBlue,
  line width=.7pt,
  -{Stealth[length=2pt]}
]
  (1.22,2.89)
    .. controls +(0,-.29) and +(0,-.29) ..
  (.62,2.89);

\draw[
  spanRed,
  line width=.75pt,
  -{Stealth[length=2pt]}
]
  (2.76,2.89)
    .. controls +(0,-.47) and +(0,-.47) ..
  (1.76,2.89);

\draw[
  spanGreen,
  line width=.7pt,
  -{Stealth[length=2pt]}
]
  (3.82,2.89)
    .. controls +(0,-.29) and +(0,-.29) ..
  (3.24,2.89);

\node[
  font=\scriptsize,
  text=spanBlue
] at (.92,2.39)
  {positive};

\node[
  font=\scriptsize,
  text=spanRed,
  align=center
] at (2.26,2.16)
  {within-person\\negative};

\node[
  font=\scriptsize,
  text=spanGreen
] at (3.54,2.39)
  {positive};

\node[outcome] at (2.02,1.22)
  {State $\uparrow$\qquad Identity $\approx$ preserved};

\end{scope}

\end{tikzpicture}%
}

\caption{Disentangling longitudinal scale and learned invariance.
(a) Under ordinary same-person sampling, record span $L$ and
supervision span $\tau$ grow together.
(b) Keeping supervision local isolates the effect of increasing
record span.
(c) Fixing the record while broadening supervision isolates the
effect of supervision span.
(d) Within-person discrimination tests whether preserving
distinctions among states can recover information lost over long records.}
\label{fig:longitudinal-scale}

\end{figure*}
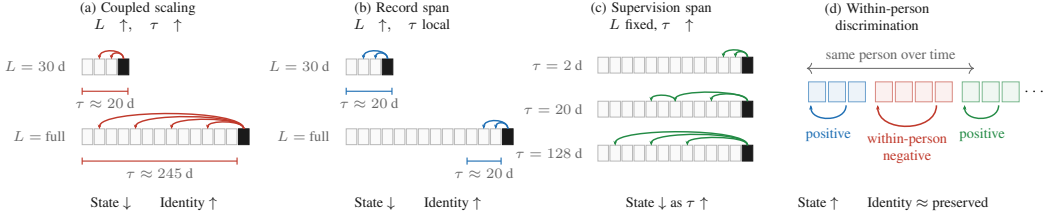

\subsection{Local supervision does not guarantee local invariance}

We first isolate record span by increasing $L$ while keeping supervision span
$\tau$ local (Fig.~\ref{fig:longitudinal-scale}b). Training histories are nested
and share an end date, so increasing $L$ adds progressively older observations
without extending the reach of positive pairs. Each participant contributes
the same number of training anchors across record spans.

Keeping supervision local substantially attenuates, but does not eliminate,
the effect of longer records. From 30 days to the full record, changing-state
information falls from $0.1410$ to $0.0944$ nats/dim in the single-resident
cohort ($-33\%$) and from $0.1025$ to $0.0539$ in households ($-47\%$).
Meanwhile, identity decodability rises from $0.546$ to $0.613$ and from
$0.668$ to $0.757$, respectively.

More surprisingly, the learned invariance itself does not remain local. As the record grows, observations from the same person separated by more than 180 days become increasingly similar despite never appearing as positive pairs. Thus, long records can make distant states converge even when the objective never directly aligns them.

\subsection{Broader supervision directly suppresses changing state}
\label{sec:supervision-span}

We next isolate supervision span by holding the available record fixed and
varying only $\tau$ (Fig.~\ref{fig:longitudinal-scale}c). Every condition sees
the same history; only the temporal reach of same-person supervision changes.

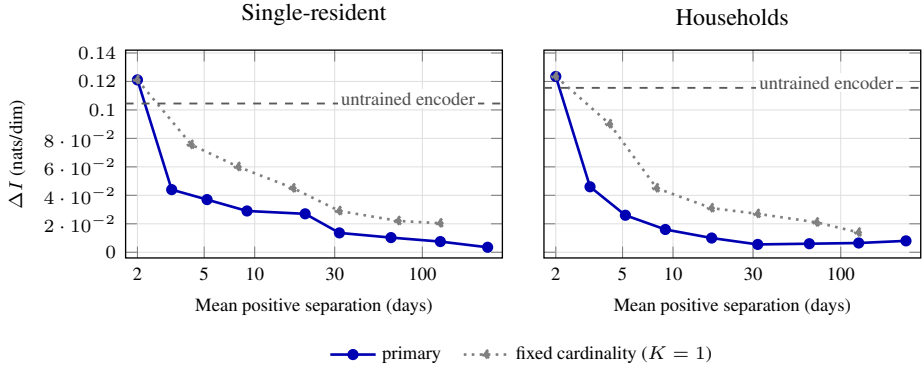
\begin{figure}[!ht]
\centering

\begin{tikzpicture}

\begin{groupplot}[
    group style={
        group size=2 by 1,
        horizontal sep=0.55cm,
        ylabels at=edge left,
        yticklabels at=edge left,
    },
    width=0.47\linewidth,
    height=0.31\linewidth,
    xmode=log,
    log basis x=10,
    xmin=1.7,
    xmax=300,
    ymin=-0.004,
    ymax=0.142,
    xtick={2,5,10,30,100},
    xticklabels={2,5,10,30,100},
    ytick={0,0.02,0.04,0.06,0.08,0.10,0.12,0.14},
    grid=major,
    grid style={
        line width=.25pt,
        draw=gray!25
    },
    axis line style={
        line width=.5pt
    },
    tick style={
        line width=.4pt
    },
    xlabel={Mean positive separation (days)},
    ylabel={$\Delta I$ (nats/dim)},
    title style={
        font=\small
    },
    label style={
        font=\scriptsize
    },
    tick label style={
        font=\scriptsize
    },
    legend style={
        font=\scriptsize,
        draw=none,
        fill=none,
        /tikz/every even column/.append style={
            column sep=6pt
        }
    },
    legend columns=2,
]

\nextgroupplot[
    legend to name=supspanlegend,
    title={Single-resident}
]

\addplot[
    blue!70!black,
    solid,
    line width=1pt,
    mark=*,
    mark size=1.6pt
] coordinates {
    (2.0,0.1211)
    (3.2,0.0440)
    (5.2,0.0370)
    (9.0,0.0290)
    (20.0,0.0270)
    (32.0,0.0136)
    (65.0,0.0103)
    (128.0,0.0075)
    (245.0,0.0035)
};

\addlegendentry{primary}

\addplot[
    gray,
    dotted,
    line width=1pt,
    mark=diamond*,
    mark size=1.8pt
] coordinates {
    (2.0,0.1211)
    (4.2,0.0755)
    (8.0,0.0600)
    (17.0,0.0450)
    (32.0,0.0290)
    (72.0,0.0220)
    (128.0,0.0204)
};

\addlegendentry{fixed cardinality ($K=1$)}

\addplot[
    black!65,
    dashed,
    line width=.7pt,
    forget plot
] coordinates {
    (1.7,0.1045)
    (300,0.1045)
};

\node[
    font=\scriptsize,
    anchor=east,
    text=black!70,
    fill=white,
    inner sep=1pt
] at (axis cs:215,0.108)
{untrained encoder};

\nextgroupplot[
    title={Households}
]

\addplot[
    blue!70!black,
    solid,
    line width=1pt,
    mark=*,
    mark size=1.6pt
] coordinates {
    (2.0,0.1235)
    (3.2,0.0460)
    (5.2,0.0260)
    (9.0,0.0160)
    (17.0,0.0100)
    (32.0,0.0055)
    (65.0,0.0060)
    (128.0,0.0065)
    (245.0,0.0080)
};

\addplot[
    gray,
    dotted,
    line width=1pt,
    mark=diamond*,
    mark size=1.8pt
] coordinates {
    (2.0,0.1235)
    (4.2,0.0900)
    (8.0,0.0450)
    (17.0,0.0310)
    (32.0,0.0270)
    (72.0,0.0210)
    (128.0,0.0139)
};

\addplot[
    black!65,
    dashed,
    line width=.7pt,
    forget plot
] coordinates {
    (1.7,0.1155)
    (300,0.1155)
};

\node[
    font=\scriptsize,
    anchor=east,
    text=black!70,
    fill=white,
    inner sep=1pt
] at (axis cs:215,0.119)
{untrained encoder};

\end{groupplot}

\node[
    anchor=north,
    inner sep=0pt
] at
($(group c1r1.south)!0.5!(group c2r1.south)+(0,-1.02cm)$)
{\pgfplotslegendfromname{supspanlegend}};

\end{tikzpicture}

\vspace{1.5mm}

\caption{Broader supervision suppresses changing-state information at fixed record span. As mean positive-pair separation increases, $\Delta I$ declines sharply in both cohorts, including when each anchor has exactly one positive ($K=1$). Dashed lines denote matched untrained encoders.}
\label{fig:4.2}

\end{figure}

As Fig.~\ref{fig:4.2} shows, changing-state information declines sharply as
positive pairs reach farther through time. At the narrowest span, the learned
representation preserves approximately the information available in a matched
untrained encoder ($1.16\times$ and $1.07\times$ in the two cohorts); at the
broadest span, less than $10\%$ remains.

This decline is not explained by the number of available positives. With
exactly one positive per anchor ($K=1$), increasing $\tau$ from 2 to 128 days
still reduces changing-state information by $83\%$ and $89\%$ in the
single-resident and household cohorts, respectively
(Fig.~\ref{fig:4.2}). Thus, broadening supervision alone removes most
recoverable changing-state information, even though the available history is
unchanged. PCA and temporally local TNC preserve substantially more
state, while trained representations move far from initialization
(Appendix~\ref{app:tnc}).

Broader supervision therefore directly suppresses changing-state information. But it does not explain the full record-span effect: even with local supervision, longer records make distant, never-paired states increasingly similar.

\subsection{Within-person discrimination recovers the lost state}

Person-level contrastive learning distinguishes people from one another, but
does not require different states of the same person to remain distinguishable.
That omission matters increasingly as the record grows and exposes the learner
to a broader range of within-person variation.

We test this directly at the full record span. Positive supervision remains
local, but observations from other groups within the same person's history can
now serve as negatives (Fig.~\ref{fig:longitudinal-scale}d). The available
history and positive pairs are unchanged; only whether within-person
differences matter to the objective is altered.

The effect is striking: within-person discrimination fully reverses the long-record loss. Without shortening the record or changing positive supervision, changing-state information increases by 54--56\% in the single-resident cohort ($0.0944 \rightarrow 0.1453$--$0.1474$ nats/dim) and 96--121\% in households ($0.0539 \rightarrow 0.1058$--$0.1189$), returning to the 30-day level in both cohorts while identity remains strongly decodable.

Controls isolate what drives the recovery. Random rather than temporal groupings work equally well, while partitioning the record without contrasting across groups changes nothing. The intervention also provides no benefit on short records. Together, these results show that recovery comes not from temporal segmentation itself, but from requiring the objective to preserve distinctions within a person's history.

Together, these experiments separate two effects of longitudinal scale: broader supervision directly suppresses changing state, while within-person discrimination reverses the additional loss that emerges over long records.

\subsection{The supervision-span effect transfers prospectively}
\label{sec:globem-transfer}

So far, we have established the supervision-span effect within DETECT. To test whether it generalizes, we use a separate GLOBEM cohort \citep{xu2022globem} to define and fix the state-selection rule, then apply it unchanged to a later cohort of 199 participants spanning Bluetooth, location, and screen sensing. The two datasets are analyzed at different temporal resolutions: DETECT canonical analyses use 5-day windows, with daily-resolution analyses where finer control is needed, whereas GLOBEM is analyzed daily throughout. If supervision span---rather than the particular dataset---drives state suppression, broadening supervision at fixed record span should again reduce changing-state information.

The prediction holds. Broadening realized supervision span from $3.2$ to $24.6$ days---a $7.7\times$ increase---reduces changing-state information by $0.0070$ nats/dim, with all five seeds agreeing in direction. A prespecified secondary readout independently reproduces the effect ($-0.0064$ nats/dim).\footnote{The smaller absolute effect does not reflect a lack of changing-state signal: PCA recovers $20.3\%$ of baseline prediction error, compared with $0.14\%$ under unrestricted person-contrastive learning.}

GLOBEM further reinforces the distinction between record span and supervision span. Its shorter records constrain the record-span sweep to realized separations of $14.4$--$24.5$ days, missing the shorter-separation range where most of the decline occurs. Directly varying supervision spans $3.2$--$24.6$ days and reproduces the effect; where the two sweeps overlap, their $\Delta I$ values are comparable. Thus, the GLOBEM results transfer the supervision-span effect across population, dataset, and sensing setting while further showing that temporal reach---not record length alone---is the relevant axis.

\section{Discussion}

Longitudinal scale changes the learning problem in a way that ordinary dataset scale does not. As a record grows, the learner sees not only more observations, but more variation within the same person. A fixed rule such as ``same person'' can therefore encompass an increasingly diverse set of states without any change to the objective itself. More data can silently become more invariance.

Our experiments make this hidden change measurable. By separating record span from supervision span, we show that they are distinct axes of longitudinal scale: broader supervision directly suppresses changing-state information, while longer records broaden invariance even when supervision remains local. The latter effect reaches states that are never paired during training. Requiring those states to remain distinguishable recovers the lost information without shortening the record, showing that the tradeoff is not between more history and more state. It is between how much variation the learner can observe and how much of that variation the objective allows it to preserve.

This changes how longitudinal scaling should be understood. Record length is a property of the data; invariance is a property of the learning problem imposed on those data. They need not scale together. More generally, whenever self-supervision defines relationships among observations, increasing dataset scale may change what falls within those relationships even when the loss and sampling rule remain fixed. Scaling such systems therefore requires tracking not only how much data is added, but how the meaning of supervision changes with it.

\paragraph{Limitations.}
Our experiments establish these effects for person-level contrastive learning in behavioral sensing, not self-supervised learning generally. The two DETECT cohorts provide within-study consistency rather than independent replication. GLOBEM externally reproduces the direct supervision-span effect; its shorter within-person records, however, do not allow record span to vary enough to reproduce the long-record loss that the within-person intervention is designed to reverse. We evaluate multiple controls, objectives, and encoder capacities, but do not establish whether the same behavior holds at substantially larger model scales or across other forms of self-supervision.

Longitudinal data are valuable precisely because they capture change. Our results show that using more of that history does not require becoming invariant to more of it. \emph{Scale the record, not the invariance.}

\section*{Ethics and data availability.}
DETECT data were collected under an IRB-approved longitudinal study
(DETECT-AD), with informed consent from participants.
Data are available upon request from the data-holding institution.
\section*{Reproducibility Statement}
The evaluation protocol, chronological splits, leakage audits and the $\Delta I$
estimator are specified in Appendix~\ref{app:protocol}. Architectures,
optimization settings, positive and negative sampling, and compute are given in
Appendix~\ref{app:implementation}. Cohort construction, household
deduplication and feature definitions are in Appendix~\ref{app:data}. All
reported effects are averaged over five seeds with unit-clustered bootstrap
intervals. DETECT and GLOBEM are credentialed-access datasets and cannot be
redistributed.

\section*{AI Use Statement}
In accordance with the ICLR 2027 policy on AI use, we disclose that large language
model tools were used for two purposes: to draft sections of the paper, and to aid
or polish writing. All AI-assisted content was reviewed, verified, and edited by the
authors, who take full responsibility for the final manuscript.

\bibliography{bibliography}
\bibliographystyle{iclr2027_conference}

\appendix

\section*{Appendix Overview}

The appendix is organized around the paper's central empirical claims.
Appendices~\ref{app:data}--\ref{app:protocol} document the data, evaluation protocol, and
estimator. Appendix~\ref{app:availability} characterizes the changing-state information available
in the longitudinal record. Appendices~\ref{app:recordspan}--\ref{app:globem-appendix} provide
supporting analyses for the four experiments of Section~\ref{sec:longer-records}: record span
under local supervision, supervision span at fixed record length, within-person discrimination,
and prospective replication in GLOBEM. Appendix~\ref{app:robustness} collects cross-cutting
robustness analyses, Appendix~\ref{app:scope} reports additional and negative results that
delimit the scope of the conclusions, and Appendix~\ref{app:implementation} gives implementation
details.

\section{DETECT: Data, Cohorts, and Preprocessing}
\label{app:data}
\label{app:features}

This appendix documents DETECT, the in-home behavioral sensing study used
throughout. GLOBEM, used only for the replication in
Section~\ref{sec:globem-transfer}, is described in Appendix~\ref{app:globem}.

\subsection{Daily feature construction}
\label{app:feature-construction}

Each retained day is represented by a 48-dimensional feature vector derived
from eight sensing channels:
\[
\begin{aligned}
\{\;&\text{motion events},\;
     \text{motion areas},\;
     \text{steps},\;
     \text{bed activity},\\
    &\text{respiratory rate},\;
     \text{medication events},\;
     \text{time in bed},\;
     \text{weight observations}\;\}.
\end{aligned}
\]

For each channel $c$, we compute three daily statistics:
\[
\log(1+\mathrm{total}_c),
\qquad
\frac{\mathrm{activity}_{c,00{:}00\text{--}05{:}59}}
     {\mathrm{activity}_{c,\mathrm{day}}},
\qquad
H(p_{c,1},\ldots,p_{c,24}),
\]
where the second quantity is the fraction of activity occurring between
00:00 and 05:59 inclusive, that is, in the six hours before 06:00, and $H$ is the entropy of the normalized 24-hour activity
distribution. These statistics contribute $8\times3=24$ dimensions.

For motion events, we additionally retain the normalized 24-bin hourly
profile,
\[
(p_{\mathrm{mot},1},\ldots,p_{\mathrm{mot},24}),
\]
giving 24 additional dimensions and 48 dimensions in total. Thirty of the
48 dimensions are therefore motion-derived.

\subsection{Retention and temporal aggregation}
\label{app:retention}

We retain days with at least 18 hours of motion-sensor coverage. Five-day
windows are retained when at least four of their five constituent days satisfy
the daily coverage criterion. Retained daily feature vectors are averaged
within non-overlapping 5-day windows.

Feature standardization parameters are estimated using training-era windows
only and subsequently applied unchanged to validation and test data. No
statistics from test-era windows are used to construct the normalization
applied to training data.

\subsection{Cohort construction and household deduplication}
\label{app:cohort}

The raw panel contains 57 single-resident subject records from 57 distinct
homes and 87 multi-resident subject records from 44 homes. Among the
multi-resident homes, 43 contribute two co-resident subject records and one
contributes a single record.

Individual experiments retain fewer units than the panel, because each
comparison requires enough windows to construct its history block, buffer, and
prediction target. Across the experiments reported here, 51--55 single-resident
participants and 42 households satisfy the eligibility criteria; each results table states the
count for the experiment it reports. We report panel counts here and per-experiment counts where
the distinction matters.

This distinction matters because the ambient sensors are installed at the
home level rather than attached to an individual. Consequently, co-residents
can share the same dominant ambient sensing stream even though they appear as
separate subject records in the raw panel.

We quantify this dependence directly by comparing sensing channels between
co-resident records from the same home. Motion sensing is identical across all
43 co-resident pairs (within-home correlation 1.000, 43/43), as are weight
observations, indicating a shared scale. Bed-derived channels are also strongly
correlated wherever both records carry them (medians 0.38--0.89), consistent
with a shared bed. Only steps (median 0.206) and medication events (median
0.011) behave as genuinely individual channels. Several channels are present
with variance in both records for only 8--14 of the 43 pairs, so those medians
rest on few homes.

Table~\ref{tab:coresident-correlation} reports the channel-level results.

\begin{table}[!ht]
\centering
\caption{Within-home similarity of co-resident sensing streams. Motion and
weight channels are duplicated across all 43 paired homes, and bed-derived
channels are near-duplicated, motivating household-level deduplication. Only
steps and medication events behave as individual channels.}
\label{tab:coresident-correlation}
\small
\begin{tabular}{lcc}
\toprule
Channel & Median correlation & Pairs with variance in both \\
\midrule
Motion events & 1.000 & 43/43 \\
Motion areas & 1.000 & 43/43 \\
Steps & 0.206 & 14/43 \\
Bed activity & 0.382 & 8/43 \\
Respiratory rate & 0.850 & 8/43 \\
Medication events & 0.011 & 8/43 \\
Time in bed & 0.890 & 8/43 \\
Weight observations & 1.000 & 43/43 \\
\bottomrule
\end{tabular}
\end{table}

Treating all 87 multi-resident records as independent participants would
duplicate the dominant home-level sensing stream and artificially increase
the apparent sample size. We therefore define the independent observational
unit in this cohort to be the household.

For each home, we retain one record, prioritizing first the record with the
largest number of available sensing modalities and then the record with the
largest number of recorded hours. This procedure yields 44 independent
household streams.

The household cohort is consequently used to assess consistency of
\emph{stream-level} findings. We do not use it to make claims about
individual-level behavioral state changes, because the mapping between the
ambient stream and a particular co-resident is not identifiable.

We reran the headline analyses before and after household deduplication.
Table~\ref{tab:dedup-sensitivity} reports the comparison. Deduplication
reduces the effective sample size and widens uncertainty, but does not reverse
the headline conclusions. In particular, the recent-history advantage remains
large: the corresponding gap changes from approximately 0.245 nats per
dimension under the record-level analysis to approximately 0.215 nats per
dimension after household deduplication.

\begin{table}[!ht]
\centering
\caption{Sensitivity of headline results to household deduplication. The
record-level analysis treats the 87 subject records separately; the
household-level analysis retains one stream from each of 44 homes.}
\label{tab:dedup-sensitivity}
\small
\begin{tabular}{lcc}
\toprule
Analysis & 87 records & 44 households \\
\midrule
Recent $\Delta I$ & 0.2746 & 0.2325 \\
Shifted $\Delta I$ & 0.0306 & 0.0158 \\
Random $\Delta I$ & 0.0235 & 0.0261 \\
Distant $\Delta I$ & 0.0324 & 0.0051 \\
\midrule
Recent $-$ shifted & 0.2441 & 0.2167 \\
Recent $-$ random & 0.2511 & 0.2064 \\
Recent $-$ distant & 0.2422 & 0.2274 \\
\bottomrule
\end{tabular}
\end{table}

\subsection{Feature observation and motion-only robustness}
\label{app:feature-observation}

Several non-motion channels are intermittently observed. Absence is encoded
as zero in the constructed daily vector, making observation frequency an
important property of the feature representation. Table~\ref{tab:feature-summary}
therefore reports the mean, standard deviation, and empirical zero rate for
every input dimension.

All motion-derived dimensions have a zero rate of 0.00. Non-motion channels
range from 0.14 (steps) through 0.32--0.33 (bed activity, respiratory rate,
time in bed) to 0.47 (medication-event entropy) and 0.69 (weight-observation
entropy). In addition, the weight-observation volume feature and the
overnight-fraction and entropy dimensions derived from motion areas are
near-degenerate, with standard deviations of 0.270, 0.016, and 0.071
respectively.
As missingness or observation frequency could itself carry temporal
information, we accompany analyses using the full 48-dimensional feature set
with a 30-dimensional motion-only robustness analysis where relevant.

\begin{table}[!ht]
\centering
\caption{Summary of the 48 daily input features, with the 24 hourly-profile bins collapsed to their range. Zero rate denotes the
fraction of retained days on which the corresponding feature is zero.
Exact feature-level summary statistics are computed from the final
preprocessing output.}
\label{tab:feature-summary}
\small
\begin{tabular}{llrrr}
\toprule
Channel & Statistic & Mean & SD & Zero rate \\
\midrule
Motion events & $\log(1{+}\mathrm{total})$ & 7.818 & 0.787 & 0.00 \\
Motion areas & $\log(1{+}\mathrm{total})$ & 4.964 & 0.430 & 0.00 \\
Steps & $\log(1{+}\mathrm{total})$ & 6.323 & 2.949 & 0.14 \\
Bed activity & $\log(1{+}\mathrm{total})$ & 3.673 & 3.048 & 0.32 \\
Respiratory rate & $\log(1{+}\mathrm{total})$ & 2.657 & 2.222 & 0.33 \\
Medication events & $\log(1{+}\mathrm{total})$ & 1.077 & 0.568 & 0.11 \\
Time in bed & $\log(1{+}\mathrm{total})$ & 1.280 & 1.084 & 0.32 \\
Weight observations & $\log(1{+}\mathrm{total})$ & 0.489 & 0.270 & 0.15 \\
Motion events & Overnight fraction & 0.094 & 0.061 & 0.00 \\
Motion areas & Overnight fraction & 0.225 & 0.016 & 0.00 \\
Steps & Overnight fraction & 0.021 & 0.046 & 0.34 \\
Bed activity & Overnight fraction & 0.240 & 0.218 & 0.35 \\
Respiratory rate & Overnight fraction & 0.298 & 0.266 & 0.35 \\
Medication events & Overnight fraction & 0.458 & 0.350 & 0.21 \\
Time in bed & Overnight fraction & 0.295 & 0.261 & 0.35 \\
Weight observations & Overnight fraction & 0.393 & 0.372 & 0.32 \\
Motion events & Hourly entropy & 2.683 & 0.160 & 0.00 \\
Motion areas & Hourly entropy & 3.088 & 0.071 & 0.00 \\
Steps & Hourly entropy & 1.650 & 0.798 & 0.14 \\
Bed activity & Hourly entropy & 1.082 & 0.958 & 0.33 \\
Respiratory rate & Hourly entropy & 1.189 & 1.037 & 0.33 \\
Medication events & Hourly entropy & 0.154 & 0.232 & 0.47 \\
Time in bed & Hourly entropy & 1.215 & 1.046 & 0.33 \\
Weight observations & Hourly entropy & 0.056 & 0.096 & 0.69 \\
Motion events & Hourly profile, 24 bins & 0.013--0.064 & 0.012--0.032 & 0.00 \\
\bottomrule
\end{tabular}
\end{table}

As absence is encoded as zero, several non-motion dimensions double as
indicators of device availability, and device availability is itself
autocorrelated in time. A recency advantage could therefore partly reflect
sensor uptime predicting sensor uptime rather than behavior predicting
behavior. We repeat the history-source analysis restricted to the 30
motion-derived dimensions, every one of which has a zero rate of 0.00 and so
cannot encode device presence. Both the history block and the prediction
target are restricted.

Table~\ref{tab:motion-only} reports the result. The recent-history advantage
persists in both cohorts. We do not interpret the reduction in magnitude as an
estimate of the confounded share, because the 18 excluded dimensions carry
behavioral information as well as missingness, and the two cannot be separated
by this comparison.

\begin{table}[!ht]
\centering
\caption{History-source results using only the 30 motion-derived features.
Values are $\Delta I$ in nats per output dimension.}
\label{tab:motion-only}
\small
\begin{tabular}{lcc}
\toprule
History arm & Single-resident & Households \\
\midrule
Recent & 0.1335 & 0.1171 \\
Shifted & 0.0241 & 0.0425 \\
Random & 0.0409 & 0.0499 \\
Distant & 0.0553 & 0.0559 \\
\bottomrule
\end{tabular}
\end{table}

The same motion-only panel is used in Appendix~\ref{app:motion-only-scale} to test whether
changing sensor availability drives the scale effect.

\section{Evaluation and Statistical Protocol}
\label{app:protocol}
\label{app:estimator}

\subsection{Chronological splits and leakage prevention}
\label{app:splits}
\label{app:audit}
\label{app:leakage}

All data splits are chronological within the independent observational unit:
participant for the single-resident cohort and household for the
household cohort. Training and test periods are separated by an
experiment-specific buffer sufficient to cover the prediction horizon and
the maximum history displacement required by the comparison.

Schematically,
\[
\boxed{\text{training anchors}}
\quad\big|\quad
\boxed{\text{buffer}}
\quad\big|\quad
\boxed{\text{test anchors}}.
\]

All arms within a comparison use the same eligible test-anchor list. Thus,
changing the source or ordering of history does not change the future targets
on which the competing conditions are evaluated.

The person-specific baseline $B_i$ is computed exclusively from training-era
windows and is fixed across all experimental arms.

We explicitly audit the constructed datasets for chronology violations,
target leakage, inconsistent eligibility, and baseline contamination.
Table~\ref{tab:leakage-audit} summarizes these checks.

These checks concern the \emph{prediction} pathway: the features from which $\Delta I$ is
estimated never contain the target window, in any arm. They say nothing about the
\emph{self-supervised} pathway, where a same-person positive block may overlap the target
even though the encoder's evaluation input does not. The two exposures are distinct and we
audit them separately. \textsuperscript{\dag}Appendix~\ref{app:target-overlap} quantifies
the second, which is non-zero and cap-dependent, and reports the corrected experiment.

\begin{table}[!ht]
\centering
\caption{Adversarial audit of temporal splitting and history construction.}
\label{tab:leakage-audit}
\small
\begin{tabular}{lc}
\toprule
Check & Violations \\
\midrule
Training chronology violations & 0 \\
Test chronology violations & 0 \\
Target contained in \emph{prediction} input & 0 \\
Target contained in an SSL \emph{positive} block & audited separately\textsuperscript{\dag} \\
Mismatched anchor eligibility across arms & 0 \\
$B_i$ uses post-training data & 0 \\
\bottomrule
\end{tabular}
\end{table}

As a negative control, we destroy the correspondence between history and
future behavior by permuting prediction targets while preserving the
marginal feature distributions and estimator pipeline. If $\Delta I$ reflects
genuine predictive dependence rather than estimator bias, the history
advantage should collapse under this manipulation.

Across arms and cohorts, the resulting estimates have magnitude at most
$0.0002$ nats per dimension, under $0.1\%$ of the recent-history value of
$0.282$ that the control is meant to rule out. The history advantage therefore
collapses when the history--future correspondence is destroyed, so it does not
arise from estimator bias and the conclusions of
Section~\ref{non-exchangeable} are unaffected.

\begin{table}[!ht]
\centering
\caption{Permuted-target negative control. All estimates are $\Delta I$ in
nats per output dimension.}
\label{tab:permuted-target}
\small
\begin{tabular}{lcc}
\toprule
History arm & Single-resident & Households \\
\midrule
Recent & $-0.0004$ & $-0.0001$ \\
Shifted & $-0.0001$ & $+0.0002$ \\
Random & $-0.0001$ & $+0.0002$ \\
Distant & $-0.0002$ & $+0.0002$ \\
\bottomrule
\end{tabular}
\end{table}

The chronological train--test buffer is set to 8 windows, which preserves the required
temporal separation while keeping anchor eligibility stable as temporal distance grows. A
stricter 24-window buffer changes which anchors remain eligible at each distance and induces an
apparent zero crossing at 20 days that is absent under the 8-window construction on the same
data; Table~\ref{tab:buffer-sensitivity} reports both. Buffer width therefore changes the
absolute level of the distance curve but not its monotone decline, so the qualitative
conclusion that state information fades with temporal distance is unaffected by this choice.

\begin{table}[!ht]
\centering
\caption{Sensitivity of the temporal-distance analysis to train--test buffer
width. The comparison isolates the effect of anchor eligibility on the
estimated distance curve.}
\label{tab:buffer-sensitivity}
\small
\begin{tabular}{lcc}
\toprule
Temporal distance & 24-window buffer & 8-window buffer \\
\midrule
5 days & $+0.117$ & $+0.285$ \\
10 days & $+0.082$ & $+0.235$ \\
15 days & $+0.044$ & $+0.194$ \\
20 days & $+0.006$ & $+0.173$ \\
25 days & $-0.016$ & $+0.153$ \\
30 days & $-0.033$ & $+0.126$ \\
35 days & $-0.050$ & $+0.113$ \\
40 days & $-0.050$ & $+0.112$ \\
\bottomrule
\end{tabular}
\end{table}

The shifted-history arm is constructed to be disjoint from the recent block: its four windows
are drawn from strictly earlier positions, so no window is shared between the two conditions. A
construction that permits overlap inflates the shifted arm, and with the disjoint construction
the four history conditions separate into two levels rather than four, with history immediately
preceding prediction on one side and histories drawn outside that local period on the other.

\subsection{Conditional usable predictive information}
\label{app:deltaI}

For participant or household $i$, let $B_i$ denote the stable baseline and
let $z$ denote the representation of the history under evaluation. We fit two
predictors of the same future target $Y$. The baseline predictor receives
$B_i$ alone,
\[
\hat{Y}_0=g_{\hat{\theta}_0}(B_i),
\]
while the history-conditioned predictor receives both the baseline and the
history representation,
\[
\hat{Y}_1=g_{\hat{\theta}_1}([B_i,z]).
\]

We define usable predictive information as
\[
\Delta I
=
\frac{1}{D}\,
\mathbb{E}_{\mathrm{test\ anchors}}
\left[
\log p_{\hat{\theta}_1}(Y\mid B_i,z)
-
\log p_{\hat{\theta}_0}(Y\mid B_i)
\right],
\]
where $D=48$ is the number of output dimensions, so $\Delta I$ is reported in nats per output
dimension and the log-likelihoods are joint over the $D$ target dimensions.

Both predictors belong to the same ridge predictive family and are evaluated
on identical held-out targets. Positive $\Delta I$ therefore indicates that
the history representation contains information about future behavior that is
usable beyond the stable person- or household-specific baseline.

For each output dimension $d$, likelihood is evaluated under a Gaussian
predictive model,
\[
\log p(Y_d\mid X)
=
-\frac{1}{2}
\left[
\log(2\pi\sigma_d^2)
+
\frac{(Y_d-\hat{Y}_d)^2}{\sigma_d^2}
\right].
\]
Predictive variances $\sigma_d^2$ are calibrated on a held-out validation
slice. We average the resulting log-likelihood differences over output
dimensions and report $\Delta I$ in nats per dimension.

The canonical estimator conditions directly on the stable baseline:
\[
X_{\mathrm{canonical}}=[B_i,z].
\]

As $B_i$ and $z$ can encode overlapping stable information, we also
evaluate a residualized construction. We fit a predictor $r$ of $z$ from
$B_i$ using training data and define
\[
z_i^{\perp}=z_i-r(B_i).
\]
The corresponding history-conditioned input is
\[
X_{\mathrm{residualized}}=[B_i,z_i^{\perp}].
\]

Table~\ref{tab:canonical-residualized} compares the two constructions for each history arm,
both cohorts, and both learned and mean representations. The two estimators agree closely for
the learned representation: no arm differs by more than $0.0022$ nats/dim, under $1\%$ of the
recent-history estimate and roughly $5\%$ of the record-span decline the estimator is used to
measure. Recent history gives $0.2818$ against $0.2796$ in the single-resident cohort, a
difference of under $1\%$ of the estimate and therefore essentially unchanged. Differences are
larger for the mean summary, where the shifted and random arms fall slightly below zero under
residualization, but those arms carry almost no information under either construction. The
distant arm is at or near zero throughout ($0.0051$ learned in both cohorts, under $2\%$ of the
recent-history value), indicating little incremental information on the scale of the effects
studied here, consistent with the corrected history-source results.

\subsection{Bootstrap inference and seed aggregation}
\label{app:bootstrap}
\label{app:statistics}

Our inferential unit is the independent observational stream rather than the
training seed. For the single-resident cohort, we bootstrap participants with
replacement. For the household cohort, we bootstrap households with
replacement. All observations belonging to a sampled participant or
household are retained together within a bootstrap replicate.

We use 2000 bootstrap replicates and construct 95\% percentile confidence
intervals from the resulting distribution of per-replicate means.

\paragraph{Seeds are resampled, not averaged.} Every interval on a trained arm resamples
\emph{hierarchically}: within each replicate we draw participants (or households) with
replacement \emph{and} seeds with replacement, rather than averaging the five seeds into a point
estimate and then resampling participants around it. Averaging first treats the seed draw as
free, which it is not. Between-seed standard deviations here are $0.006$--$0.008$ against
between-participant standard deviations of $0.009$--$0.032$, so the seed term is smaller but not
negligible, and omitting it narrows intervals by roughly a fifth.

The choice of resampling matters for one arm. The separation-matched households arm reads
$+0.0037$ $[+0.0004,+0.0075]$ under seed-averaged resampling and $+0.0037$ $[-0.0031,+0.0128]$
under hierarchical resampling, and we report it as a null. The other arms are unaffected: both
uniform arms still
decline ($-0.0211$ $[-0.0337,-0.0101]$ and $-0.0170$ $[-0.0299,-0.0069]$) and the single-resident
matched arm was already null.

A stream-level result is considered replicated when the estimated effect has
the same sign in both cohorts and its bootstrap confidence interval excludes
zero in each cohort.

Learned-representation experiments are repeated with five random seeds.
Table~\ref{tab:seed-results} reports the mean and standard deviation across
seeds for the headline comparisons, together with the number of seeds
favoring each condition.

Agreement across seeds is treated as a consistency check rather than as the
primary inferential procedure. With five paired seeds, the minimum attainable
two-sided Wilcoxon signed-rank $p$-value is
\[
p_{\min}=0.0625.
\]
Thus, even perfect directional agreement across five seeds cannot yield
$p<0.05$ under this test. Inferential claims therefore rely on bootstrap
uncertainty over independent observational units.

\begin{table*}[!ht]
\centering
\caption{Seed variation for headline learned-representation comparisons.
``Seeds'' counts how many of five favor the first-named condition.}
\label{tab:seed-results}
\small
\begin{tabular}{lllcc}
\toprule
Experiment & Cohort & Condition & Mean $\pm$ SD & Seeds \\
\midrule
History source & Single-resident & Recent & $0.2818 \pm 0.0054$ & 5/5 \\
History source & Single-resident & Shifted & $0.0447 \pm 0.0291$ & --- \\
History source & Single-resident & Random & $0.0374 \pm 0.0169$ & --- \\
History source & Single-resident & Distant & $0.0051 \pm 0.0143$ & --- \\
History source & Households & Recent & $0.2325 \pm 0.0097$ & 5/5 \\
History source & Households & Shifted & $0.0158 \pm 0.0043$ & --- \\
History source & Households & Random & $0.0261 \pm 0.0045$ & --- \\
History source & Households & Distant & $0.0051 \pm 0.0058$ & --- \\
\midrule
Temporal order, $K{=}4$ & Single-resident & Ordered & $0.2877 \pm 0.0030$ & 5/5 \\
Temporal order, $K{=}4$ & Single-resident & Shuffled & $0.2469 \pm 0.0058$ & --- \\
Temporal order, $K{=}8$ & Single-resident & Ordered & $0.2773 \pm 0.0054$ & 5/5 \\
Temporal order, $K{=}8$ & Single-resident & Shuffled & $0.2312 \pm 0.0059$ & --- \\
Temporal order, $K{=}4$ & Households & Ordered & $0.2480 \pm 0.0041$ & 5/5 \\
Temporal order, $K{=}4$ & Households & Shuffled & $0.2252 \pm 0.0065$ & --- \\
Temporal order, $K{=}8$ & Households & Ordered & $0.2421 \pm 0.0118$ & 5/5 \\
Temporal order, $K{=}8$ & Households & Shuffled & $0.2055 \pm 0.0099$ & --- \\
\bottomrule
\end{tabular}
\end{table*}

\subsection{Identity probing}
\label{app:identity-probe}

As $\Delta I$ is defined incrementally over the person baseline $B_i$, a
history representation could in principle score near zero simply by being
uninformative, or by carrying only information $B_i$ already supplies. We
therefore probe participant identity directly on the residualized
representation actually used for scoring, that is, on $z^{\perp} = z - r(B_i)$
rather than on $z$.

We report multiclass identification accuracy against explicit chance, not AUC.
Chance is $1/54 = 0.0185$ in the single-resident cohort and $1/42 = 0.0238$ in
the household cohort. We use two probes of different capacity, multinomial
logistic regression and a small MLP, because a linear probe alone can
understate recoverable identity.

\begin{table}[!ht]
\centering
\caption{Participant identification accuracy from the residualized
representation used for scoring. Values are accuracy, not AUC.}
\label{tab:identity-probe}
\small
\begin{tabular}{lcccc}
\toprule
& \multicolumn{2}{c}{Single-resident (chance $0.0185$)}
& \multicolumn{2}{c}{Households (chance $0.0238$)} \\
\cmidrule(lr){2-3}\cmidrule(lr){4-5}
Residualized $z$ & Logistic & MLP & Logistic & MLP \\
\midrule
Recent  & 0.019 & 0.183 & 0.022 & 0.157 \\
Shifted & 0.036 & 0.510 & 0.022 & 0.541 \\
Distant & 0.895 & 0.895 & 0.888 & 0.888 \\
\bottomrule
\end{tabular}
\end{table}

We do not treat this as evidence for temporal persistence of identity. The distant block is
the participant's earliest windows, so it is \emph{the same block for every anchor} of a given
participant: near-perfect identifiability follows largely by construction rather than from
anything about temporal distance. The clean evidence for persistence is the anchor-relative
probe of Section~\ref{non-exchangeable}, where the block varies with the anchor and identity
accuracy stays roughly flat out to 180 days while state information falls to the level of the
person baseline. This
table is a diagnostic of the residualized representation rather than a measurement of how
identity survives over time. Read that way, distant history remains far above chance as an identity cue,
at $0.895$ and $0.888$, roughly
$48\times$ and $37\times$ chance, while contributing little incremental
predictive information ($\Delta I = 0.0051$ in both cohorts, under $2\%$ of the
recent-history value, Table~\ref{tab:history-source}). Recent history is the reverse: it is barely
identifying under a linear probe, at $0.019$ and $0.022$, essentially chance,
while carrying the largest $\Delta I$ of any arm.

Shifted history is the informative intermediate case. Under a linear probe it
looks unidentifying, but under the MLP probe it reaches $0.510$ and $0.541$,
higher than recent history under the same probe. Identity is therefore present
in the shifted representation in a form a linear readout misses, which is why we
do not treat a linear identity probe alone as evidence that identity has been
removed.

\subsection{Numerical conditioning and ridge selection}
\label{app:conditioning}
\label{app:solver-audit}
\label{app:ridge-sensitivity}

The conditional design matrix is severely ill-conditioned because $B_i$ and
the history representation can be highly collinear. Across the primary arms,
condition numbers for $[B_i,z]$ are typically
\[
1.3\times10^{15}
\text{--}
2.1\times10^{15},
\]
and reach approximately
\[
8\times10^{25}
\]
for distant history.

\begin{table}[!ht]
\centering
\caption{Condition numbers of the conditional design matrix $[B_i,z]$ by
experimental arm.}
\label{tab:condition-numbers}
\small
\begin{tabular}{lcc}
\toprule
History arm & Single-resident & Households \\
\midrule
Recent & $1.9\times10^{15}$ & $2.5\times10^{8}$ \\
Shifted & $1.8\times10^{15}$ & $2.4\times10^{8}$ \\
Random & $1.9\times10^{15}$ & $2.4\times10^{8}$ \\
Distant & $8.0\times10^{25}$ & $9.7\times10^{15}$ \\
\bottomrule
\end{tabular}
\end{table}

Under the extreme collinearity described above, this estimator was
numerically unstable. In particular, the distant-history arm produced
\texttt{LinAlgError: SVD did not converge}. Casting the design matrices to
64-bit floating point did not resolve the failure.

We therefore use \texttt{Ridge(solver='cholesky')} and select the
regularization penalty on a held-out validation slice.

\paragraph{Why the penalty is selected on held-out data.} At these condition numbers the
estimate depends on how the ridge penalty is chosen. Internal efficient leave-one-out on the
fitting rows is optimistic under the autocorrelation of longitudinal windows: adjacent rows are
near-duplicates, so a held-out row is nearly present in the fit, and the rule selects penalties at
the grid floor. We therefore select the penalty by explicit validation MSE on rows the fit never
sees. Holding the estimator, solver and working precision fixed and varying only the selection
rule reproduces the entire difference between the two pipelines, so the selection rule,
not the solver, is what matters.

\paragraph{The held-out penalty also predicts better.} The argument above concerns conditioning,
which is not by itself a reason to prefer one penalty over another: the question is whether the
better-conditioned choice forecasts better on data neither rule has seen. We therefore select $\lambda$ both ways on the
same design, split and seeds, and score both on the same untouched test rows
(Table~\ref{tab:lambda-predictive}). Selection is honest on each side, since internal
leave-one-out sees only the fitting rows and held-out selection only the validation rows.

Held-out selection predicts better in 8 of 8 arms across both cohorts, every interval
excluding zero, by 3.7\% to 25.1\% of the leave-one-out test MSE. Internal leave-one-out
selects $\lambda$ between 0.68 and 1.60, near the grid floor, against 62 to 127 for
held-out selection, and the corresponding design condition numbers differ by roughly an order of
magnitude. The preference for held-out selection therefore does not rest on conditioning alone:
the penalty it chooses is the one that generalizes, and the penalty internal leave-one-out
chooses is both worse conditioned and worse at prediction.

\begin{table}[!ht]
\centering
\caption{Does the held-out-selected penalty predict better on held-out data? Penalties are
selected two ways on the same design, split and seeds: \texttt{RidgeCV}'s internal efficient
leave-one-out on the fitting rows, and explicit validation MSE on the held-out validation rows.
Neither sees the test rows. Condition numbers are of the regularized design at the selected
penalty, averaged over seeds. Neither rule uses the test rows for fitting or for
selection: internal leave-one-out sees only the fitting rows, held-out selection sees only the
validation rows, and the test rows enter once, after both penalties are fixed, solely to compare
how the two generalize. The final columns give the paired difference in test MSE, LOO minus
held-out-selected, so positive favors held-out selection; brackets are $95\%$ bootstrap
intervals over units, five seeds.}

\label{tab:lambda-predictive}
\small
\resizebox{\linewidth}{!}{%
\begin{tabular}{llrrrrrrrl}
\toprule
& & \multicolumn{2}{c}{Penalty} & \multicolumn{2}{c}{Design cond.} &
\multicolumn{2}{c}{Test MSE} & \multicolumn{2}{c}{Test MSE, LOO $-$ HO} \\
\cmidrule(lr){3-4}\cmidrule(lr){5-6}\cmidrule(lr){7-8}\cmidrule(lr){9-10}
Cohort & Arm & LOO & HO & LOO & HO & LOO & HO & $\Delta$ & $95\%$ CI \\
\midrule
Single-resident & Uniform, 30\,d & $0.71$ & $78$ & $810$ & $13$ & $1.1551$ & $0.8656$ & $+0.2895$ & $[+0.0726,+0.5975]$ \\
 & Uniform, full & $1.60$ & $62$ & $95$ & $14$ & $0.9152$ & $0.8813$ & $+0.0339$ & $[+0.0106,+0.0606]$ \\
 & Matched, 30\,d & $1.12$ & $78$ & $112$ & $13$ & $0.9246$ & $0.8686$ & $+0.0560$ & $[+0.0243,+0.0942]$ \\
 & Matched, full & $1.16$ & $111$ & $129$ & $11$ & $0.9384$ & $0.8698$ & $+0.0685$ & $[+0.0308,+0.1122]$ \\
\midrule
Households & Uniform, 30\,d & $0.89$ & $62$ & $109$ & $13$ & $0.8896$ & $0.8062$ & $+0.0834$ & $[+0.0253,+0.1702]$ \\
 & Uniform, full & $1.12$ & $78$ & $100$ & $12$ & $0.8686$ & $0.8199$ & $+0.0487$ & $[+0.0119,+0.0972]$ \\
 & Matched, 30\,d & $0.88$ & $127$ & $142$ & $9$ & $0.9283$ & $0.8083$ & $+0.1200$ & $[+0.0349,+0.2283]$ \\
 & Matched, full & $0.68$ & $94$ & $132$ & $11$ & $0.8945$ & $0.8076$ & $+0.0869$ & $[+0.0264,+0.1750]$ \\
\bottomrule
\end{tabular}}
\end{table}
This also settles the solver question on its own terms. Recomputing the strong fixed-anchor
cells with an SVD/pseudoinverse ridge, $\beta = V\,\mathrm{diag}\!\left(s/(s^2+\lambda)\right)U^{\top}y$,
which never forms $X^{\top}X$, reproduces the Cholesky estimates to $10^{-15}$ and leaves every
sign and interval unchanged. We note for completeness that the normal-equation form squares the
condition number and is the less stable formulation in principle; it is immaterial here only
because the selected penalty dominates the conditioning.

The decoder penalty is selected on a held-out validation split rather than by internal
leave-one-out, which leaves the penalty grid and the choice of validation rows free. Re-running
the record-span contrast over three grids ($\log$-spaced $10^{-3}$--$10^{4}$, $10^{-6}$--$10^{8}$
and $10^{-8}$--$10^{10}$, with $20$, $40$ and $60$ points) crossed with four validation splits
(the last $15\%$, $25\%$ and $10\%$ of the training era, and a random $15\%$) gives twelve
specifications per arm and cohort. The uniform decline is negative in $24/24$ cells across both
cohorts, with an interval excluding zero in $12/12$ single-resident cells and $8/12$ household
cells; the cells that lose the interval are the random split and the last $25\%$. The matched
arms never show a consistent effect: the single-resident arm is positive in $12/12$ cells and
significant in none, and the household arm takes both signs across splits and is significant in
both directions depending on the split, which is a second reason for reporting it as a null
(Appendices~\ref{app:record-span-builds} and~\ref{app:bootstrap}). We therefore quote the
single-resident uniform decline as robust to decoder specification, the household decline as
robust in sign but split-sensitive in significance, and no matched arm as showing an effect.

\section{Availability of Changing-State Information}
\label{app:availability}

\subsection{History sources and representation-free baselines}
\label{app:repfree}

Every $\Delta I$ elsewhere is measured through a learned encoder. Two baselines separate what
the data contain from what the encoder contributes: the same history block fed directly to the
ridge decoder, and an untrained encoder of identical architecture and hidden width, at 20
random initializations. Anchors, $B_i$, targets, splits, ridge grid and the participant
bootstrap are unchanged.

\begin{table}[!ht]
\centering
\caption{History-source arms under three representations. The untrained column is the mean
over 20 initializations of the same architecture \texttt{train()} uses.}
\label{tab:repfree}
\small
\resizebox{\linewidth}{!}{%
\begin{tabular}{lcccccc}
\toprule
& \multicolumn{3}{c}{Single-resident} & \multicolumn{3}{c}{Households} \\
\cmidrule(lr){2-4}\cmidrule(lr){5-7}
Arm & Learned & Untrained & Raw block & Learned & Untrained & Raw block \\
\midrule
Recent  & $+0.2818$ & $+0.0705$ & $+0.3437$ & $+0.2325$ & $+0.0476$ & $+0.2559$ \\
Shifted & $+0.0447$ & $-0.0016$ & $+0.0340$ & $+0.0158$ & $+0.0006$ & $+0.0017$ \\
Random  & $+0.0374$ & $-0.0010$ & $+0.0176$ & $+0.0261$ & $+0.0008$ & $+0.0175$ \\
Distant & $+0.0051$ & $-0.0032$ & $+0.0597$ & $+0.0051$ & $-0.0030$ & $+0.0040$ \\
\midrule
Recent $-$ shifted & $+0.2371$ & $+0.0721$ & $+0.3097$ & $+0.2167$ & $+0.0469$ & $+0.2543$ \\
Recent $-$ random  & $+0.2444$ & $+0.0715$ & $+0.3261$ & $+0.2064$ & $+0.0468$ & $+0.2384$ \\
\bottomrule
\end{tabular}}
\end{table}

Three things follow (Table~\ref{tab:repfree}). The recency separation exists without any
learning: the raw-block gaps ($+0.3097$, $+0.2543$) are \emph{larger} than the learned ones
($+0.2371$, $+0.2167$). A random projection preserves the ordering but roughly a quarter of the
magnitude, which is what compressing a $192$-dimensional block to $32$ dimensions with no
objective would do. And training recovers most of what the projection loses while still ending
below the raw block. Temporal non-exchangeability is therefore a property of the data that
representation learning neither creates nor fully preserves.

One caveat on the distant arm: raw distant is $+0.0597$ $[+0.0156,+0.1004]$ in the
single-resident cohort, nonzero unlike its learned ($+0.0051$) and untrained ($-0.0032$)
counterparts. That is the fixed-block artifact of Appendix~\ref{app:identity-probe} reappearing
--- the earliest block is constant within a unit, so an unbottlenecked decoder can exploit it
directly. It is a further reason to prefer the anchor-relative far arm.

\paragraph{The same bracket across the record-span configurations.} Repeating this on the
builds used for the record-span sweep, the separation-matched control and the pair-pool
experiments gives raw-history $\Delta I$ of $0.3001$ and $0.2236$, against untrained floors of
$0.0907$ and $0.0983$ and a best learned value of $0.2299$ and $0.1906$. No learned
representation in this paper exceeds the raw history block fed to the same decoder. The
comparison is not capacity-matched --- raw supplies 96 dimensions against the learned
representations' 32 --- so we treat it as context for the untrained comparison rather than as a
claim that encoders are unnecessary.

\subsection{Behavioral-drift operationalizations}
\label{app:drift}

The temporal non-exchangeability analyses ask whether history becomes less
informative as it moves away from the prediction target. A related question
is whether the advantage of recent history is larger when behavior itself is
changing. We operationalize behavioral drift in three ways and distinguish
between pooled and within-participant stratification.

\begin{figure}[!ht]
\centering

\begin{tikzpicture}

\begin{axis}[
    name=leftplot,
    width=0.47\linewidth,
    height=0.36\linewidth,
    title={(a) Across participants},
    title style={font=\small},
    xlabel={Unit's average displacement},
    ylabel={$\Delta I$: recent $-$ random (nats/dim)},
    symbolic x coords={Low,Medium,High},
    xtick=data,
    ymin=0.05,
    ymax=0.47,
    ytick={0.1,0.2,0.3,0.4},
    grid=major,
    tick label style={font=\footnotesize},
    label style={font=\footnotesize},
    legend style={
        font=\scriptsize,
        at={(0.02,0.98)},
        anchor=north west,
        draw=none,
        fill=none
    },
]

\addplot[
    blue,
    solid,
    line width=1.2pt,
    mark=*,
    mark size=2pt,
    mark options={
        solid,
        fill=blue,
        draw=blue
    },
    error bars/.cd,
        y dir=both,
        y explicit
]
table[
    x=group,
    y=mean,
    y error minus=errminus,
    y error plus=errplus
] {
group   mean    errminus   errplus
Low     0.1410  0.0353     0.0313
Medium  0.2038  0.0323     0.0312
High    0.3837  0.0559     0.0609
};
\addlegendentry{Single-resident}

\addplot[
    gray,
    solid,
    line width=1.2pt,
    mark=square*,
    mark size=2pt,
    mark options={
        solid,
        fill=gray,
        draw=gray
    },
    error bars/.cd,
        y dir=both,
        y explicit
]
table[
    x=group,
    y=mean,
    y error minus=errminus,
    y error plus=errplus
] {
group   mean    errminus   errplus
Low     0.1089  0.0306     0.0296
Medium  0.1748  0.0270     0.0294
High    0.3116  0.0627     0.0608
};
\addlegendentry{Multi-resident (households)}

\end{axis}

\begin{axis}[
    at={(leftplot.east)},
    anchor=west,
    xshift=1.0cm,
    width=0.47\linewidth,
    height=0.36\linewidth,
    title={(b) Within a participant},
    title style={font=\small},
    xlabel={Displacement from own past},
    symbolic x coords={Low,Medium,High},
    xtick=data,
    ymin=0.05,
    ymax=0.47,
    ytick={0.1,0.2,0.3,0.4},
    yticklabels={},
    grid=major,
    tick label style={font=\footnotesize},
    label style={font=\footnotesize},
]

\addplot[
    blue,
    solid,
    line width=1.2pt,
    mark=*,
    mark size=2pt,
    mark options={
        solid,
        fill=blue,
        draw=blue
    },
    error bars/.cd,
        y dir=both,
        y explicit
]
table[
    x=group,
    y=mean,
    y error minus=errminus,
    y error plus=errplus
] {
group   mean    errminus   errplus
Low     0.1492  0.0246     0.0236
Medium  0.2359  0.0389     0.0413
High    0.3483  0.0457     0.0481
};

\node[
    anchor=north west,
    font=\tiny
] at (rel axis cs:0.04,0.96)
{Spearman $\rho=0.191$,\ 95\% CI $[0.122,0.257]$};

\end{axis}

\end{tikzpicture}

\caption{Recency advantage as a function of behavioral displacement. \textbf{(a)} Across single-resident participants and multi-resident households. \textbf{(b)} Within single-resident participants, relative to each participant's own past. Points show tercile means; error bars show bootstrap 95\% confidence intervals.}
\label{fig:drift}

\end{figure}
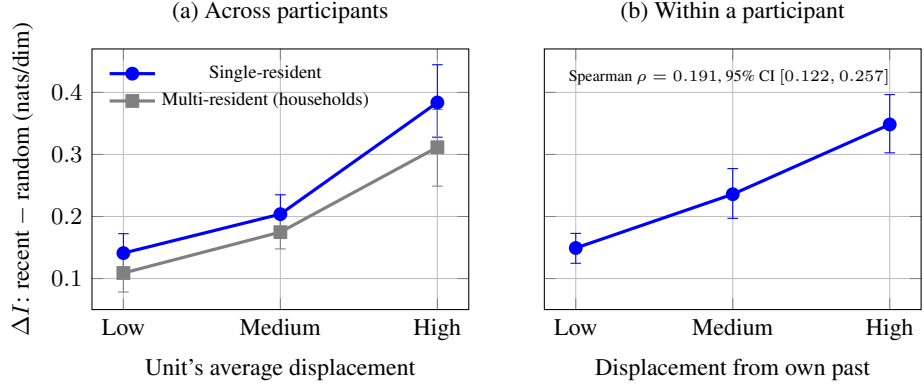

The recency advantage is largest for units whose behavior has changed most, both between units
and, for the contemporaneous measure, within the same person over time (Fig.~\ref{fig:drift};
$\rho=0.191$, 95\% CI $[0.122,0.257]$ within single-resident participants).

Let $x_{i,t}$ denote the behavioral window for participant $i$ at time $t$.
We evaluate three definitions of drift:

\paragraph{Local drift.}
The deviation of the block immediately preceding the history block from all
earlier history. For anchor index $p$ and history length $N$,
\[
D^{\mathrm{local}}_{i,p}
=
\big\lVert
\operatorname{mean}(x_{i,p-2N:p-N})-\operatorname{mean}(x_{i,0:p-2N})
\big\rVert_2 .
\]
This is computed from windows that no history arm receives as input, so it
cannot mechanically favor the recent condition.

\paragraph{Baseline-relative drift.}
The deviation of that same preceding block from the stable baseline,
\[
D^{\mathrm{baseline}}_{i,p}
=
\big\lVert \operatorname{mean}(x_{i,p-2N:p-N})-B_i \big\rVert_2 .
\]

\paragraph{Person-level drift.}
A single late-versus-early contrast per unit, where $c_i$ marks the end of the
training era,
\[
D^{\mathrm{person}}_{i}
=
\big\lVert \operatorname{mean}(x_{i,c_i:})-B_i \big\rVert_2 ,
\]
which is constant within a unit and therefore degenerate under within-unit
stratification.

Drift is stratified into terciles two ways. Pooled terciles are formed over all eligible
participant--time observations and ask whether higher-drift observations show a larger
recent-history advantage at the population level; because they can sort participants rather than
moments, we also form terciles within each participant, which asks whether a given person's
high-drift periods differ from their own low-drift periods. The two are different estimands and
need not agree. Person-level drift is constant within a unit by construction, so it is reported
only for the pooled analysis.

Table~\ref{tab:drift-results} reports the full set of analyses across drift
definitions, stratification procedures, and cohorts. For each tercile we
report the estimated predictive-information quantity and bootstrap confidence
interval, together with a direct test of whether the low- and high-drift
intervals are disjoint.

Person-level interpretations are restricted to the single-resident cohort.
Results from the household cohort are interpreted only at the household
stream level.

\begin{table*}[!ht]
\centering
\caption{Predictive-information results stratified by behavioral drift.
Within-participant stratification is not defined for the person-level drift
measure.}
\label{tab:drift-results}
\small
\resizebox{\linewidth}{!}{%
\begin{tabular}{lllcccc}
\toprule
Cohort & Drift definition & Stratification &
Low & Middle & High & Low/high disjoint \\
\midrule
Single-resident & Local & Pooled & +0.1817 [.131,.234] & +0.2305 & +0.3215 [.262,.369] & yes \\
Single-resident & Local & Within & +0.2367 [.185,.281] & +0.2246 & +0.2723 [.227,.316] & no \\
Single-resident & Baseline & Pooled & +0.1793 [.128,.235] & +0.2390 & +0.3151 [.256,.363] & yes \\
Single-resident & Baseline & Within & +0.2422 [.189,.285] & +0.2253 & +0.2661 [.223,.307] & no \\
Single-resident & Person & Pooled & +0.1739 [.138,.210] & +0.2182 & +0.3412 [.288,.405] & yes \\
Households & Local & Pooled & +0.1164 [.094,.138] & +0.1870 & +0.3163 [.259,.361] & yes \\
Households & Local & Within & +0.1558 [.121,.189] & +0.1962 & +0.2672 [.208,.315] & yes \\
Households & Baseline & Pooled & +0.1106 [.088,.136] & +0.1977 & +0.3110 [.243,.359] & yes \\
Households & Baseline & Within & +0.1550 [.118,.191] & +0.1925 & +0.2718 [.216,.317] & yes \\
Households & Person & Pooled & +0.1436 [.112,.174] & +0.1650 & +0.3091 [.230,.380] & yes \\
\bottomrule
\end{tabular}}
\end{table*}

A drift measure can capture two different things: how \emph{far} a unit's current behavior sits
from its own past, and how \emph{fast} it is currently changing. These come apart, and only the
first holds. For each unit we correlate the recency advantage ($\Delta I$ recent $-$ random)
across that unit's anchors with three measures --- displacement of the current block from the
unit's earlier history, the same displacement lagged one block, and the rate of change between
consecutive blocks --- and average the Spearman correlations over units
(Table~\ref{tab:drift-mechanism}).

\begin{table}[!ht]
\centering
\caption{Within-unit Spearman correlation between the recency advantage and
three measures of behavioral change. Brackets show bootstrap 95\% confidence
intervals over units.}
\label{tab:drift-mechanism}
\small
\begin{tabular}{lcc}
\toprule
Within-unit measure & Single-resident ($n=50$) & Households ($n=41$) \\
\midrule
Displacement from own past & $+0.191$ $[+0.122,+0.257]$ & $+0.231$ $[+0.147,+0.313]$ \\
Displacement, lagged one block & $+0.037$ $[-0.034,+0.114]$ & $+0.130$ $[+0.067,+0.194]$ \\
Rate of change & $+0.008$ $[-0.060,+0.075]$ & $-0.033$ $[-0.105,+0.031]$ \\
\bottomrule
\end{tabular}
\end{table}

Only displacement from the unit's own past replicates: it is positive with an
interval excluding zero in both cohorts. The rate measure is null in both, and
the lagged measure holds in the household cohort but not in the
single-resident cohort. This resolves the apparent within-participant null in
Table~\ref{tab:drift-results}, whose local and baseline definitions are both
lagged by one block: the recency advantage is larger when a unit has moved far
from where it used to be, not when it happens to be moving quickly.

We therefore restrict the within-person claim to displacement and do not claim
that the within-person relationship holds across all three drift definitions.

\paragraph{Temporal-distance sweep under both estimators.}
The comparison above covers the four history-source arms. As the monotone decay of
Section~\ref{non-exchangeable} is the result most likely to be sensitive to how $z$ is
conditioned on $B_i$, we recompute the full distance sweep under both constructions.
Table~\ref{tab:resid-sweep} reports all eight gaps in both cohorts.

\begin{table}[!ht]
\centering
\caption{Temporal-distance sweep under the canonical and residualized estimators. A fixed
20-day block is moved progressively further from the anchor with the target held at $t+5$
days. Brackets are 95\% bootstrap intervals over units; five seeds per cell.}
\label{tab:resid-sweep}
\small
\resizebox{\linewidth}{!}{%
\begin{tabular}{lcccc}
\toprule
& \multicolumn{2}{c}{Single-resident} & \multicolumn{2}{c}{Households} \\
\cmidrule(lr){2-3}\cmidrule(lr){4-5}
Gap & Canonical & Residualized & Canonical & Residualized \\
\midrule
0 days  & $+0.2818$ $[+0.2289,+0.3385]$ & $+0.2796$ $[+0.2272,+0.3363]$ & $+0.2325$ $[+0.1810,+0.2883]$ & $+0.2324$ $[+0.1807,+0.2881]$ \\
5 days  & $+0.2263$ $[+0.1808,+0.2760]$ & $+0.2236$ $[+0.1782,+0.2732]$ & $+0.1857$ $[+0.1360,+0.2396]$ & $+0.1846$ $[+0.1346,+0.2376]$ \\
10 days & $+0.1877$ $[+0.1465,+0.2352]$ & $+0.1846$ $[+0.1434,+0.2321]$ & $+0.1526$ $[+0.1061,+0.2047]$ & $+0.1514$ $[+0.1046,+0.2026]$ \\
15 days & $+0.1715$ $[+0.1328,+0.2162]$ & $+0.1668$ $[+0.1283,+0.2105]$ & $+0.1445$ $[+0.0990,+0.1953]$ & $+0.1420$ $[+0.0963,+0.1931]$ \\
20 days & $+0.1534$ $[+0.1169,+0.1928]$ & $+0.1498$ $[+0.1129,+0.1893]$ & $+0.1174$ $[+0.0723,+0.1664]$ & $+0.1141$ $[+0.0689,+0.1624]$ \\
30 days & $+0.1458$ $[+0.1085,+0.1854]$ & $+0.1427$ $[+0.1053,+0.1825]$ & $+0.0899$ $[+0.0493,+0.1328]$ & $+0.0876$ $[+0.0451,+0.1324]$ \\
40 days & $+0.1236$ $[+0.0882,+0.1592]$ & $+0.1207$ $[+0.0856,+0.1565]$ & $+0.0733$ $[+0.0393,+0.1100]$ & $+0.0687$ $[+0.0341,+0.1068]$ \\
60 days & $+0.1158$ $[+0.0807,+0.1517]$ & $+0.1120$ $[+0.0772,+0.1480]$ & $+0.0641$ $[+0.0307,+0.0987]$ & $+0.0584$ $[+0.0237,+0.0941]$ \\
\bottomrule
\end{tabular}}
\end{table}

Both curves are monotone decreasing in both cohorts. Over 60 days the canonical estimator
gives $-59\%$ and $-72\%$ and the residualized estimator $-60\%$ and $-75\%$. No paired
value differs by more than $0.006$ nats per dimension and every pair of intervals overlaps
almost entirely. The decay is therefore a property of the data rather than of the
conditioning choice.

\subsection{Conditioned and unconditioned persistence}
\label{app:uncond}

Throughout, $\Delta I$ is \emph{incremental} over the person baseline: how much a history block
adds beyond what $B_i$ already supplies. We therefore quantify what that conditioning does to
the measured timescale, recomputing the temporal-distance sweep without any
conditioning, comparing
\[
  \Delta I(d) = \mathrm{NLL}(Y \mid B_i) - \mathrm{NLL}(Y \mid B_i, z_d),
  \qquad
  U(d) = \mathrm{NLL}(Y \mid \text{const}) - \mathrm{NLL}(Y \mid z_d),
\]
on identical encoders, lags, splits, seeds, ridge grid and unit bootstrap.

\begin{table}[!ht]
\centering
\caption{Persistence with and without conditioning on $B_i$, normalized to the value at $d=0$.
Raw values in nats per output dimension.}
\label{tab:uncond}
\small
\resizebox{\linewidth}{!}{%
\begin{tabular}{lcccccc}
\toprule
& \multicolumn{3}{c}{Single-resident} & \multicolumn{3}{c}{Households} \\
\cmidrule(lr){2-4}\cmidrule(lr){5-7}
Gap & $\Delta I(d)$ & $U(d)$ & $U/\Delta I$ & $\Delta I(d)$ & $U(d)$ & $U/\Delta I$ \\
\midrule
0 days  & $0.2818$ ($1.000$) & $0.4829$ ($1.000$) & $1.7$ & $0.2325$ ($1.000$) & $0.4453$ ($1.000$) & $1.9$ \\
5 days  & $0.2263$ ($0.803$) & $0.4244$ ($0.879$) & $1.9$ & $0.1857$ ($0.799$) & $0.3993$ ($0.897$) & $2.2$ \\
10 days & $0.1877$ ($0.666$) & $0.3893$ ($0.806$) & $2.1$ & $0.1526$ ($0.656$) & $0.3624$ ($0.814$) & $2.4$ \\
20 days & $0.1534$ ($0.544$) & $0.3566$ ($0.738$) & $2.3$ & $0.1174$ ($0.505$) & $0.3298$ ($0.741$) & $2.8$ \\
30 days & $0.1458$ ($0.517$) & $0.3427$ ($0.710$) & $2.4$ & $0.0899$ ($0.387$) & $0.3025$ ($0.679$) & $3.4$ \\
40 days & $0.1236$ ($0.439$) & $0.3353$ ($0.694$) & $2.7$ & $0.0733$ ($0.315$) & $0.2870$ ($0.645$) & $3.9$ \\
60 days & $0.1158$ ($0.411$) & $0.3217$ ($0.666$) & $2.8$ & $0.0641$ ($0.276$) & $0.2562$ ($0.575$) & $4.0$ \\
\bottomrule
\end{tabular}}
\end{table}

Conditioning on $B_i$ isolates changing-state information; without conditioning, predictive
information persists much longer (Table~\ref{tab:uncond}). Normalized to its value at $d=0$,
the conditioned quantity falls to $0.411$ and $0.276$ by 60 days while the unconditioned one
retains $0.666$ and $0.575$, and the ratio between them widens monotonically with distance
($1.7\to2.8$ and $1.9\to4.0$). For scale, the baseline alone is worth $0.2018$ and $0.2371$
nats, against an unconditioned curve that plateaus near $0.32$ and $0.26$: most of what a
distant block still predicts is stable person-level information that $B_i$ already carries.

We therefore read $\Delta I$ as a measure of \emph{deviation from a person's own baseline}
rather than of behavioral predictability in general. Behavior at large remains highly
predictable from months-old data because people are stable; what decays on the short timescales
reported here is the departure-from-baseline component. Statements elsewhere in this appendix
about state becoming stale should be understood in that sense.

\section{Record Span with Local Supervision}
\label{app:recordspan}

Holding positive supervision local substantially attenuates, but does not eliminate, the effect
of record span. Across nested histories, increasing the available record from $30$ days to the
full record reduces changing-state information by $33\%$ in single-resident participants and
$47\%$ in households, while identity becomes more decodable. This appendix describes the
construction, shows that the effect extends to same-person states that are never paired, and
tests alternative explanations for the residual decline.

\subsection{Nested-history construction}
\label{app:nested-construction}

Histories are nested and end at the same cut. At record span $L$, a training anchor is eligible if
it lies within the last $L$ days of its unit's training era, so the $L$-day pool is contained in
every longer one and the most recent history is present in all conditions. Positives are drawn
with separation capped at the $30$-day reference at every $L$: realized mean separation is
$20.1$--$20.7$ days in all conditions, and the largest distance between the separation
distribution at any $L$ and at $L=30$ is $0.03$ on a draw-weighted Kolmogorov--Smirnov statistic.
Positives are also confined to the training era; no positive block ends after a unit's cut, which
we verified exhaustively ($0$ of $27{,}096$ and $0$ of $21{,}904$ drawn positives).

Each unit contributes the same number of anchors per epoch at every $L$, equal to its count in the
$30$-day pool, so longer records do not buy a unit more gradient weight. The step budget, batch
size, encoder, optimizer, negatives, targets, splits and person baselines $B_i$ are identical
across $L$ (Appendix~\ref{app:implementation}). Anchors are resampled from the unit's $L$-day span
each epoch; Appendix~\ref{app:record-span-builds} reports variants that hold the set of distinct
anchors fixed, that weight units by record length, and that use the base-capacity encoder.

\subsection{Changing state and identity across record span}
\label{app:nested-results}

Changing-state information falls as the record grows, by $33\%$ and $47\%$ from a $30$-day record
to the full record, with $5/5$ seeds declining in both cohorts and intervals excluding zero
(Table~\ref{tab:nested-recordspan}). Most of the decline arrives by $120$ days and the curve is
then flat, which matches the separation-driven decline of Appendix~\ref{app:supspan} in shape but
arises with positive separation pinned near $20$ days throughout.

Identity moves the other way over the same sweep, from $0.546$ to $0.613$ and from $0.668$ to
$0.757$. Longer records therefore make the representation more person-identifiable even though no
positive pair ever spans more than a month, which is the pairing of effects that
Appendix~\ref{app:nested-geometry} explains geometrically.

\begin{table}[!ht]
\centering
\caption{Changing-state information and identity as the available record grows under local
supervision, for the construction used in Section~\ref{sec:longer-records}. Nested histories end
at the same cut, positive separation stays capped at the $30$-day reference at every $L$, and each
unit contributes the same number of anchors per epoch. Intervals are paired unit bootstraps on the
$30$-day to full-record change (Appendix~\ref{app:bootstrap}); ``seeds'' counts how many of five
decline.}
\label{tab:nested-recordspan}
\small
\resizebox{\linewidth}{!}{%
\begin{tabular}{lccccccc}
\toprule
& \multicolumn{3}{c}{$\Delta I$ at record span $L$ (days)} & & & & \\
\cmidrule(lr){2-4}
Cohort & 30 & 120 & full & Change & 95\% CI & Seeds & Identity \\
\midrule
Single-resident & $0.1410$ & $0.0963$ & $0.0944$ & $-33\%$ & $[-0.0639,-0.0320]$ & 5/5 & $0.546\!\to\!0.613$ \\
Households & $0.1025$ & $0.0539$ & $0.0539$ & $-47\%$ & $[-0.0641,-0.0359]$ & 5/5 & $0.668\!\to\!0.757$ \\
\bottomrule
\end{tabular}}
\end{table}
\subsection{Geometry of distant same-person states}
\label{app:nested-geometry}

What changes with the record is how the representation allocates variance. On the fixed
evaluation anchors, between-person dispersion rises and within-person temporal variation falls
monotonically with $L$, so their ratio grows from $2.40$ to $3.10$ and from $3.55$ to $4.74$
(Table~\ref{tab:nested-geometry}). Across all $25$ cells of the $L\times$seed grid, that ratio and
$\Delta I$ are strongly anticorrelated, $\rho=-0.74$ and $-0.73$: the cells with the most
person-dominated geometry are the cells that lose the most changing-state information.

The contraction reaches pairs the objective never touches. A participant's own windows more than
$180$ days apart, which are never eligible as positives at any $L$, move closer together relative
to the distance between participants, from $0.637$ to $0.576$ and from $0.539$ to $0.458$. In the
household cohort the far pairs contract more than the near pairs ($-15\%$ against $-6\%$); in the
single-resident cohort the contraction is nearly uniform across separations, so it reads there as
a global compression of within-person structure rather than a specifically long-range effect.
Local positives therefore bound which observations are explicitly equated, but not how similar
distant states become.

\begin{table}[!ht]
\centering
\caption{Representation geometry on the fixed evaluation anchors as the record grows under local
supervision, $L_2$-normalized. Between-person is the dispersion of participant centroids and
within-person is the mean spread of a participant's own embeddings. The last two rows give the
mean distance between a participant's own windows, divided by the mean distance to other
participants' windows, for near and far pairs; the far pairs are never used as positives at any
$L$.}
\label{tab:nested-geometry}
\small
\begin{tabular}{lccccc}
\toprule
Record span $L$ (days) & 30 & 60 & 120 & 240 & full \\
\midrule
\multicolumn{6}{l}{\emph{Single-resident}} \\
$\Delta I$ & $0.141$ & $0.136$ & $0.096$ & $0.091$ & $0.094$ \\
identity & $0.546$ & $0.546$ & $0.571$ & $0.585$ & $0.613$ \\
between-person & $0.333$ & $0.329$ & $0.357$ & $0.373$ & $0.371$ \\
within-person & $0.139$ & $0.133$ & $0.124$ & $0.122$ & $0.120$ \\
between/within & $2.397$ & $2.478$ & $2.876$ & $3.069$ & $3.097$ \\
own windows $<30$\,d apart & $0.431$ & $0.424$ & $0.405$ & $0.398$ & $0.391$ \\
own windows $>180$\,d apart & $0.637$ & $0.633$ & $0.604$ & $0.585$ & $0.576$ \\
\multicolumn{6}{l}{\emph{Households}} \\
$\Delta I$ & $0.102$ & $0.085$ & $0.054$ & $0.042$ & $0.054$ \\
identity & $0.668$ & $0.683$ & $0.719$ & $0.745$ & $0.757$ \\
between-person & $0.432$ & $0.454$ & $0.459$ & $0.486$ & $0.484$ \\
within-person & $0.122$ & $0.114$ & $0.101$ & $0.100$ & $0.102$ \\
between/within & $3.548$ & $4.003$ & $4.556$ & $4.873$ & $4.738$ \\
own windows $<30$\,d apart & $0.360$ & $0.349$ & $0.339$ & $0.332$ & $0.337$ \\
own windows $>180$\,d apart & $0.539$ & $0.508$ & $0.470$ & $0.455$ & $0.458$ \\
\bottomrule
\end{tabular}
\end{table}
\subsection{Alternative explanations for the residual decline}
\label{app:nested-controls}

We next test four alternative explanations for the residual decline.

\paragraph{Not the number of distinct training examples.} Holding the count of distinct anchors at
its $30$-day value, spaced evenly across the growing span, leaves the decline at $-38\%$ and
$-51\%$ (Table~\ref{tab:nested-variants}), so it is not a consequence of longer records supplying
more distinct examples at the same compute.

\paragraph{Not the age of the training data.} Sliding a $30$-day block of anchors backwards in
time, with anchor count and unit composition fixed, costs $-9\%$ and $-4\%$ at an age of $255$
days, both intervals spanning zero. Intermediate ages are non-monotone and vary by roughly
$\pm20\%$ between blocks, which is the period-to-period variability of the panel rather than a
trend.

\paragraph{Not cross-period discrimination inside the batch.} Drawing each batch so that every
unit contributes anchors from a single $30$-day slice, so no participant is discriminated across
their own periods within a gradient step, leaves the decline essentially unchanged: $-27\%$
against $-33\%$ in the single-resident cohort and $-44\%$ against $-47\%$ in households. The
local-batch advantage at the full record is $+0.0088$ $[+0.0043,+0.0139]$ and $+0.0038$
$[-0.0004,+0.0086]$, so at most a fifth of the residual is attributable to within-batch
cross-period contrast.

\paragraph{Not a shift in the inputs, nor within-person heterogeneity.} Units whose older training
inputs sit further from their test-period inputs do not lose more ($\rho$ between $-0.12$ and
$+0.11$ across arms and cohorts); the broad-history arms' training inputs are in fact slightly
closer to the test-period mean than the recent block's. Units whose behavior varies more across
their own $30$-day periods also do not lose more, whether heterogeneity is measured by centroid
dispersion, a between-to-within ratio, or an energy distance between period distributions ($\rho$
between $-0.12$ and $+0.19$, every interval containing zero, $n=54$ and $42$). These tests have
power only against strong relationships.

\subsection{Alternative record-span constructions}
\label{app:record-span-builds}

\begin{table}[!ht]
\centering
\caption{Variants of the nested-history construction: a fixed set of distinct anchors rather than a fresh draw each epoch, and the base-capacity encoder. Unrestricted pairing on the same build is shown for contrast.
Nested histories end at the same cut; positives are capped at the $30$-day reference in every
condition, so realized separation stays near $20$ days throughout, and each unit contributes the
same number of anchors per epoch at every $L$. ``Local'' caps positive separation; ``unrestricted''
lets positives come from anywhere in the available record and is shown for contrast. Intervals are
paired unit bootstraps on the $30$-day to full-record change; the last column gives identity
accuracy at the two endpoints.}
\label{tab:nested-variants}
\small
\resizebox{\linewidth}{!}{%
\begin{tabular}{lll ccccc cccc}
\toprule
& & & \multicolumn{5}{c}{$\Delta I$ at record span $L$ (days)} & & & & \\
\cmidrule(lr){4-8}
Cohort & Encoder & Supervision & 30 & 60 & 120 & 240 & full & Change & 95\% CI & Seeds & Identity \\
\midrule
Single-resident & $160$k & local & $0.1419$ & $0.1356$ & $0.1056$ & $0.0946$ & $0.0878$ & $-38\%$ & $[-0.0692,-0.0412]$ & 5/5 & $0.54\!\to\!0.59$ \\
Single-resident & $160$k & unrestricted & $0.1386$ & $0.1063$ & $0.0673$ & $0.0452$ & $0.0346$ & $-75\%$ & $[-0.1281,-0.0833]$ & 5/5 & $0.53\!\to\!0.72$ \\
Single-resident & $40$k & local & $0.0863$ & $0.0898$ & $0.0927$ & $0.0913$ & $0.0771$ & $-11\%$ & $[-0.0202,-0.0000]$ & 4/5 & $0.46\!\to\!0.48$ \\
Single-resident & $40$k & unrestricted & $0.0911$ & $0.0765$ & $0.0711$ & $0.0379$ & $0.0196$ & $-78\%$ & $[-0.0900,-0.0534]$ & 5/5 & $0.47\!\to\!0.58$ \\
Households & $160$k & local & $0.1081$ & $0.0899$ & $0.0572$ & $0.0621$ & $0.0533$ & $-51\%$ & $[-0.0702,-0.0409]$ & 5/5 & $0.67\!\to\!0.74$ \\
Households & $160$k & unrestricted & $0.1136$ & $0.0690$ & $0.0543$ & $0.0342$ & $0.0306$ & $-73\%$ & $[-0.1010,-0.0665]$ & 5/5 & $0.68\!\to\!0.79$ \\
Households & $40$k & local & $0.0965$ & $0.0898$ & $0.0965$ & $0.1007$ & $0.0893$ & $-8\%$ & $[-0.0142,-0.0004]$ & 3/5 & $0.63\!\to\!0.63$ \\
Households & $40$k & unrestricted & $0.0945$ & $0.0789$ & $0.0623$ & $0.0398$ & $0.0292$ & $-69\%$ & $[-0.0792,-0.0521]$ & 5/5 & $0.63\!\to\!0.71$ \\
\bottomrule
\end{tabular}}
\end{table}

Three distinct constructions in this paper vary something called record span, and they measure
different things. The distinction that matters is between the record span $L$, the history
available to the encoder, and the supervision span $\tau$, the temporal reach over which the
objective is trained to treat two observations as interchangeable. The three constructions sit
differently with respect to it: the expanding sweep grows $L$ and $\tau$ together and is
therefore descriptive; the fixed-anchor sweep holds the record fixed and broadens $\tau$ alone,
which is the direct evidence that broad supervision suppresses changing state; and the
nested-history sweep grows $L$ while holding $\tau$ local, which is the only one of the three
that asks what more history costs by itself.

\paragraph{Fixed-anchor construction ($L$ fixed in substance, $\tau$ varied).} The record length $L$
enters in exactly one place: the pool from which a positive may be drawn,
$\{q : |q-p| \le L/\text{WD}\}$. The anchor list, the encoder input, the negatives, the
prediction targets, the train/validation/test partition and the person baselines $B_i$ are
identical at every $L$, and candidate positives are drawn from the unit's whole record at every
$L$. Growing $L$ therefore does not change how much history the encoder sees; it changes only how
far supervision may reach, which is the definition of supervision span. We report it here as a
supervision-reach sweep and not as a record-span manipulation. Its separation-matched arm is
correspondingly uninformative about record span: the matched pool does not depend on $L$, so
that arm is the shortest-record condition redrawn, and its null follows from the construction
rather than from the data. Read as a supervision-span sweep it is informative, and
Appendix~\ref{app:tau} runs the same manipulation with $\tau$ as an explicit parameter on a
record that is full throughout, which is the cleaner form of the same experiment.

\paragraph{Expanding-prediction-time construction ($L$ and $\tau$ grow together).} Here $L$
does two things at once. It gates a mask over training anchors --- an anchor is admitted if it
lies within the last $L$ days of its own unit's training era, measured from that unit's final
training anchor --- and it also caps the separation at which a positive may be drawn, exactly as
in the fixed-anchor construction. Growing the record therefore admits anchors from further back
in time, so the encoder is fitted on a different distribution of prediction times at each $L$,
\emph{and} widens supervision reach at the same time. Because the two move together, this sweep
is descriptive of the combined manipulation and cannot separate them. The number of training anchors is \emph{not} a confound: the pool is subsampled to
$N_{\text{fix}} = \min_L |\text{pool}(L)|$ at every length, so the anchor count and the epoch
budget are constant. What varies is the composition of the anchor population, not its size.

\paragraph{Nested-history construction (the record-span experiment).} Here the histories are
nested and end at the same cut: at each $L$, training anchors may be drawn from anywhere in the
last $L$ days of the unit's training era, while positives stay local, capped at the $30$-day
reference so that realized separation is pinned near $20$ days at every $L$ and never leaves the
training era. Each unit contributes the same number of anchors per epoch at every $L$, and the
step budget, encoder, optimizer, targets, splits and evaluation are unchanged. This is the
construction in which more history genuinely reaches the learner under fixed local supervision.

\paragraph{What the three give.} All three decline, in both cohorts.

\begin{table}[!ht]
\centering
\caption{The record-span decline under every construction reported in this paper.
$\Delta I$ at the shortest and longest
record. The first three blocks use uniform positive sampling: the expanding block varies the
prediction population as well as supervision reach, and the two fixed-anchor blocks vary
supervision reach alone, at base and at $4\times$ capacity. The last two blocks are the
nested-history construction with positives held local, which is the one that varies the history
available to the encoder. Absolute levels are not comparable across constructions, since the
anchor sets differ; only the within-construction change is.}
\label{tab:record-span-builds}
\small
\resizebox{\linewidth}{!}{%
\begin{tabular}{lllcccc}
\toprule
Construction & Prediction population & Encoder & Cohort & 30 days & Full & Change \\
\midrule
Expanding    & varies with $L$ & $40$k & Single-resident & $0.0971$ & $0.0198$ & $-80\%$ \\
Expanding    & varies with $L$ & $40$k & Households      & $0.0958$ & $0.0171$ & $-82\%$ \\
\midrule
Fixed-anchor & held fixed      & $40$k & Single-resident & $0.0664$ & $0.0243$ & $-63\%$ \\
Fixed-anchor & held fixed      & $40$k & Households      & $0.0396$ & $0.0138$ & $-65\%$ \\
\midrule
Fixed-anchor & held fixed      & $160$k & Single-resident & $0.0578$ & $0.0385$ & $-33\%$ \\
Fixed-anchor & held fixed      & $160$k & Households      & $0.0475$ & $0.0304$ & $-36\%$ \\
\midrule
Nested history, local positives & held fixed & $40$k  & Single-resident & $0.0863$ & $0.0771$ & $-11\%$ \\
Nested history, local positives & held fixed & $40$k  & Households      & $0.0965$ & $0.0893$ & $-7\%$ \\
\midrule
Nested history, local positives & held fixed & $160$k & Single-resident & $0.1410$ & $0.0944$ & $-33\%$ \\
Nested history, local positives & held fixed & $160$k & Households      & $0.1025$ & $0.0539$ & $-47\%$ \\
\bottomrule
\end{tabular}}
\end{table}

These constructions answer different questions. The expanding and fixed-anchor sweeps confound or
isolate changes in supervision reach, whereas the nested-history construction changes the history
available to the encoder while holding positive supervision local. Accordingly, the main-text
record-span claim rests on the nested-history result: at the reported encoder capacity, $\Delta I$
declines by $33\%$ and $47\%$ despite positive separation remaining near $20$ days. The smaller
and cohort-dependent effect at base capacity indicates that capacity modulates the magnitude of
this residual decline but does not account for its presence at the capacity used in the main
analysis. Absolute levels are not comparable across constructions, since the anchor sets differ;
only within-construction changes are.

\paragraph{The separation-matched null of the fixed-anchor sweep.} Under separation-matching at
$4\times$ capacity the fixed-anchor sweep shows no decline in either cohort, $+0.0007$
$[-0.0054,+0.0074]$ and $+0.0037$ $[-0.0031,+0.0128]$. Because the matched pool in that
construction does not depend on $L$, this null establishes that the sweep isolates supervision
reach; it is not evidence that locality protects a growing record, which is what the
nested-history construction tests. A seed-averaged interval would put the households arm
$+10.2\%$ above its $30$-day value, but it is null under the hierarchical resampling of
Appendix~\ref{app:bootstrap} and changes sign across validation splits
(Appendix~\ref{app:ridge-sensitivity}), so we report it as a null and no claim rests on a matched
arm gaining. The $4\times$ encoder itself is described in
Appendix~\ref{app:strong-contrastive}.

\paragraph{Record-span sweep under both estimators.} We repeat the check on the record-span
experiment of Section~\ref{sec:longer-records}, retraining with identical build, arms, spans and
seeds, which reproduces every value exactly (maximum absolute deviation $0$ across all forty
cells).

\begin{table}[!ht]
\centering
\caption{Record-span sweep, canonical / residualized $\Delta I$. The headline collapse and the
direction of every arm are unchanged by the estimator.}
\label{tab:resid-span}
\small
\resizebox{\linewidth}{!}{%
\begin{tabular}{llccccccc}
\toprule
Cohort & Objective & 30d & 60d & 120d & 240d & Full & Canon. & Resid. \\
\midrule
\multirow{4}{*}{Single}
 & Person-contrastive   & $.0971/.0965$ & $.0697/.0686$ & $.0565/.0557$ & $.0460/.0456$ & $.0198/.0195$ & $-80\%$ & $-80\%$ \\
 & \quad sep.-matched   & $.0896/.0887$ & $.1014/.1010$ & $.0827/.0821$ & $.1129/.1121$ & $.1004/.0998$ & $+12\%$ & $+13\%$ \\
 & Predictive           & $.1983/.1979$ & $.2129/.2124$ & $.2136/.2128$ & $.2237/.2239$ & $.2299/.2297$ & $+16\%$ & $+16\%$ \\
 & Masked               & $.2040/.2034$ & $.2042/.2038$ & $.1980/.1975$ & $.2030/.2022$ & $.2049/.2041$ & $+0\%$  & $+0\%$ \\
\midrule
\multirow{4}{*}{Households}
 & Person-contrastive   & $.0958/.0954$ & $.0744/.0741$ & $.0422/.0423$ & $.0222/.0212$ & $.0171/.0162$ & $-82\%$ & $-83\%$ \\
 & \quad sep.-matched   & $.0876/.0788$ & $.0862/.0858$ & $.0868/.0867$ & $.0755/.0690$ & $.0634/.0632$ & $-28\%$ & $-20\%$ \\
 & Predictive           & $.1770/.1764$ & $.1746/.1738$ & $.1779/.1800$ & $.1900/.1895$ & $.1906/.1899$ & $+8\%$  & $+8\%$ \\
 & Masked               & $.1806/.1798$ & $.1845/.1848$ & $.1885/.1881$ & $.1859/.1856$ & $.1864/.1858$ & $+3\%$  & $+3\%$ \\
\bottomrule
\end{tabular}}
\end{table}

The two estimators agree throughout (Table~\ref{tab:resid-span}): the maximum absolute
difference is $0.0011$ in the single-resident cohort and $0.0088$ in the household cohort,
the latter falling entirely on the separation-matched arm. In absolute terms the headline
figure is $\Delta I = 0.0971$ at 30 days falling to $0.0198$ at the full record under the
canonical estimator, and $0.0965$ to $0.0195$ under the residualized one.

\begin{table}[!ht]
\centering
\caption{Canonical and residualized estimates of $\Delta I$ under the overlap-free history
construction, the Cholesky solver and the deduplicated household cohort.}
\label{tab:canonical-residualized}
\small
\begin{tabular}{lllcc}
\toprule
Cohort & Representation & History & Canonical & Residualized \\
\midrule
Single-resident & Learned & Recent & $0.2818$ & $0.2796$ \\
Single-resident & Mean & Recent & $0.2818$ & $0.2696$ \\
Single-resident & Learned & Shifted & $0.0447$ & $0.0442$ \\
Single-resident & Mean & Shifted & $0.0071$ & $-0.0038$ \\
Single-resident & Learned & Random & $0.0374$ & $0.0357$ \\
Single-resident & Mean & Random & $0.0050$ & $-0.0027$ \\
Single-resident & Learned & Distant & $0.0051$ & $0.0071$ \\
Single-resident & Mean & Distant & $0.0196$ & $0.0139$ \\
Households & Learned & Recent & $0.2325$ & $0.2324$ \\
Households & Mean & Recent & $0.2310$ & $0.2223$ \\
Households & Learned & Shifted & $0.0158$ & $0.0166$ \\
Households & Mean & Shifted & $0.0061$ & $0.0072$ \\
Households & Learned & Random & $0.0261$ & $0.0259$ \\
Households & Mean & Random & $0.0159$ & $0.0137$ \\
Households & Learned & Distant & $0.0051$ & $0.0037$ \\
Households & Mean & Distant & $-0.0031$ & $-0.0003$ \\
\bottomrule
\end{tabular}
\end{table}

\subsection{Recency-matched baselines}
\label{app:trailing-baseline}

Every $\Delta I$ reported in the main text conditions on $B_i$, each
participant's mean over the first 60\% of their record. That reference is
\emph{stale}: it does not move as the participant does. A representation could
therefore appear informative merely by carrying recent information the baseline
lacks, and, more damaging to the record-span result, a short-record encoder could
appear to beat a long-record one because of how stale the shared baseline is
rather than because of what either encoder learned.

\paragraph{Construction.}
We replace $B_i$ with a trailing baseline,
\[
B_{i,p}^{(W)} = \operatorname{mean}\big(x_{i,\,p-N_B-W\,:\,p-N_B}\big),
\]
for $W \in \{30, 60\}$ days. The window ends at $p-N_B$ rather than at $p$, so it
is recency-matched but strictly disjoint from the encoder's own input block
$[p-N_B, p)$. This matters: a baseline overlapping the encoder input would drive
$\Delta I$ toward zero by construction, since the comparison would be conditioning
on $z$'s own input and calling it a baseline. Anchors are restricted to those at
which every baseline is defined, so all three columns are computed on identical
rows. Both decoder families of Appendix~\ref{app:mlp-decoder} are reported, so the
two robustness checks compose rather than confound one another.

\paragraph{Result.}
Table~\ref{tab:trailing-baseline} reports the sweep for uniform positives. The decline from a
30-day record to the full record is present in all twelve baseline $\times$ decoder
$\times$ cohort combinations, ranging from $55\%$ to $84\%$. The direction and
approximate scale of the record-span effect are therefore not explained by a stale
participant baseline.

\paragraph{The separation-matched arm under the same baselines.}
Table~\ref{tab:trailing-baseline-matched} repeats the sweep with separation-matched positives,
which the main text cites for the claim that recency matching preserves the qualitative
contrast. It does, but not equally in the two cohorts. In the single-resident cohort the matched
arm is close to flat under every baseline and decoder ($+3\%$ to $-22\%$, against $-61\%$ to
$-84\%$ for uniform). In the household cohort it declines substantially ($-17\%$ to
$-53\%$), still less than uniform ($-55\%$ to $-80\%$) but not negligibly. The contrast between
the arms therefore survives recency matching, while the matched arm itself is a null only in the
single-resident cohort.

\begin{table}[!ht]
\centering
\caption{Record-span sweep for \emph{uniform} positives under a fixed and two recency-matched
baselines, each with both decoder families. Values are $\Delta I$ in nats per output dimension,
computed on the anchors at which all three baselines are defined. The separation-matched arm on
the same anchors is Table~\ref{tab:trailing-baseline-matched}.}
\label{tab:trailing-baseline}
\small
\begin{tabular}{lllccccc r}
\toprule
Cohort & Baseline & Decoder & 30d & 60d & 120d & 240d & Full & 30d$\to$full \\
\midrule
\multirow{6}{*}{Single}
 & \multirow{2}{*}{Fixed $B_i$} & Ridge & .0940 & .0689 & .0525 & .0384 & .0151 & $-84\%$ \\
 &                              & MLP   & .1687 & .1317 & .0995 & .0872 & .0563 & $-67\%$ \\
 & \multirow{2}{*}{Trailing 30d}& Ridge & .0366 & .0278 & .0234 & .0104 & .0143 & $-61\%$ \\
 &                              & MLP   & .0362 & .0244 & .0241 & .0181 & .0112 & $-69\%$ \\
 & \multirow{2}{*}{Trailing 60d}& Ridge & .0430 & .0292 & .0220 & .0062 & .0074 & $-83\%$ \\
 &                              & MLP   & .0504 & .0316 & .0313 & .0261 & .0152 & $-70\%$ \\
\midrule
\multirow{6}{*}{Households}
 & \multirow{2}{*}{Fixed $B_i$} & Ridge & .0911 & .0697 & .0371 & .0244 & .0187 & $-80\%$ \\
 &                              & MLP   & .1261 & .0997 & .0440 & .0509 & .0451 & $-64\%$ \\
 & \multirow{2}{*}{Trailing 30d}& Ridge & .0266 & .0217 & .0144 & .0096 & .0071 & $-73\%$ \\
 &                              & MLP   & .0333 & .0281 & .0159 & .0133 & .0150 & $-55\%$ \\
 & \multirow{2}{*}{Trailing 60d}& Ridge & .0147 & .0113 & .0049 & .0044 & .0043 & $-71\%$ \\
 &                              & MLP   & .0376 & .0287 & .0050 & .0029 & .0094 & $-75\%$ \\
\bottomrule
\end{tabular}
\end{table}

\begin{table}[!ht]
\centering
\caption{The same sweep for the separation-matched arm, computed on the same anchors,
baselines and decoders as Table~\ref{tab:trailing-baseline}. Values are $\Delta I$ in nats per
output dimension.}
\label{tab:trailing-baseline-matched}
\small
\begin{tabular}{lllccccc r}
\toprule
Cohort & Baseline & Decoder & 30d & 60d & 120d & 240d & Full & 30d$\to$full \\
\midrule
\multirow{6}{*}{Single} & \multirow{2}{*}{Fixed $B_i$} & Ridge & 0.0950 & 0.0893 & 0.0818 & 0.1016 & 0.0982 & $+3\%$ \\
 &  & MLP & 0.1629 & 0.1488 & 0.1426 & 0.1587 & 0.1437 & $-12\%$ \\
 & \multirow{2}{*}{Trailing 30d} & Ridge & 0.0392 & 0.0351 & 0.0314 & 0.0385 & 0.0367 & $-6\%$ \\
 &  & MLP & 0.0360 & 0.0234 & 0.0343 & 0.0370 & 0.0330 & $-9\%$ \\
 & \multirow{2}{*}{Trailing 60d} & Ridge & 0.0433 & 0.0393 & 0.0307 & 0.0405 & 0.0406 & $-6\%$ \\
 &  & MLP & 0.0536 & 0.0303 & 0.0480 & 0.0508 & 0.0416 & $-22\%$ \\
\midrule
\multirow{6}{*}{Households} & \multirow{2}{*}{Fixed $B_i$} & Ridge & 0.0820 & 0.0895 & 0.0833 & 0.0626 & 0.0644 & $-22\%$ \\
 &  & MLP & 0.1287 & 0.1162 & 0.1108 & 0.0986 & 0.0838 & $-35\%$ \\
 & \multirow{2}{*}{Trailing 30d} & Ridge & 0.0209 & 0.0229 & 0.0186 & 0.0171 & 0.0173 & $-17\%$ \\
 &  & MLP & 0.0368 & 0.0345 & 0.0217 & 0.0253 & 0.0216 & $-41\%$ \\
 & \multirow{2}{*}{Trailing 60d} & Ridge & 0.0152 & 0.0169 & 0.0099 & 0.0075 & 0.0072 & $-53\%$ \\
 &  & MLP & 0.0376 & 0.0326 & 0.0208 & 0.0233 & 0.0184 & $-51\%$ \\
\bottomrule
\end{tabular}
\end{table}
\paragraph{What does change.}
Recency matching costs $60$ to $80\%$ of the absolute level of $\Delta I$. This is
the expected price of a harder test rather than a failure of it: a trailing mean
already supplies much of the recent information the representation was contributing
against a stale reference. It does mean that statements about how much information
a representation carries, as opposed to how that quantity varies with record span,
are specific to the choice of baseline and should be read as such.
\section{Supervision Span at Fixed Record Span}
\label{app:supspan}

Section~\ref{sec:longer-records} varies the available record. This appendix holds the record fixed and
varies only the maximum temporal separation permitted between an anchor and its same-person
positives. All conditions train on the full record with identical anchors, identical training
sets, identical person baselines and identical negatives; only the positive pool changes. We
verified this by hashing: the anchor array, negative array, baselines, targets and split
indices are each a single object shared across all caps, and only the positive array differs.

\subsection{Five-day windows}
\label{app:supspan-wd5}

\begin{table}[!ht]
\centering
\caption{Changing-state information and identity decodability against the maximum permitted
positive-pair separation, at fixed record span. Realized mean separation is reported because
the cap is an upper bound. No cap produced an empty pool, so none silently reverted to
unrestricted sampling.}
\label{tab:supspan-wd5}
\small
\begin{tabular}{lcccccc}
\toprule
& \multicolumn{3}{c}{Single-resident} & \multicolumn{3}{c}{Households} \\
\cmidrule(lr){2-4}\cmidrule(lr){5-7}
Cap & Realized & $\Delta I$ & Identity & Realized & $\Delta I$ & Identity \\
\midrule
10 days  & $10.0$  & $+0.1250$ & $0.593$ & $10.0$  & $+0.0982$ & $0.743$ \\
20 days  & $15.0$  & $+0.0871$ & $0.615$ & $15.0$  & $+0.0539$ & $0.763$ \\
30 days  & $20.1$  & $+0.0728$ & $0.617$ & $20.1$  & $+0.0408$ & $0.762$ \\
60 days  & $34.5$  & $+0.0628$ & $0.634$ & $34.7$  & $+0.0348$ & $0.761$ \\
90 days  & $49.4$  & $+0.0554$ & $0.648$ & $48.9$  & $+0.0366$ & $0.771$ \\
180 days & $90.1$  & $+0.0476$ & $0.686$ & $90.1$  & $+0.0302$ & $0.798$ \\
Unrestricted & $240.2$ & $+0.0266$ & $0.845$ & $239.3$ & $+0.0166$ & $0.863$ \\
\bottomrule
\end{tabular}
\end{table}

At fixed record span, changing-state information decreases as supervision span increases
(Table~\ref{tab:supspan-wd5}): $-79\%$ from the tightest cap to unrestricted in the
single-resident cohort and $-83\%$ in the household cohort. Identity decodability moves in
the opposite direction, $0.593 \to 0.845$ and $0.743 \to 0.863$. The single-resident series is
monotone across all seven levels; the household cohort has one inversion at 90 days
($0.0366$ against $0.0348$ at 60 days) well inside its interval, so we describe it as monotone
up to noise rather than strictly monotone.

The shortest cap here is 10 days, which is the minimum the geometry permits: windows are five
days and a positive must sit at least two windows from the anchor, since a closer one would
overlap the anchor block.

\subsection{Daily windows}
\label{app:supspan-wd1}

To probe separations below 10 days we rebuild the panel at daily resolution, giving two-day
history blocks and a two-day minimum separation. Minimum separation equals block length, so
this necessarily trades context for resolution and absolute values are not comparable to
Table~\ref{tab:supspan-wd5}. All positives whose block overlaps the target are excluded
(Appendix~\ref{app:target-overlap}); the untrained reference is recomputed under this geometry
from 20 initializations.

\begin{table}[!ht]
\centering
\caption{Daily-window supervision-span sweep, target-overlapping positives excluded. Five seeds
per cell. The untrained reference under this geometry is $0.1058 \pm 0.0081$ and
$0.1524 \pm 0.0102$, from 20 initializations.}
\label{tab:supspan-wd1}
\small
\begin{tabular}{lcccccc}
\toprule
& \multicolumn{3}{c}{Single-resident} & \multicolumn{3}{c}{Households} \\
\cmidrule(lr){2-4}\cmidrule(lr){5-7}
Cap & Realized & $\Delta I$ & Identity & Realized & $\Delta I$ & Identity \\
\midrule
2 days & $2.0$ & $+0.1515$ & $0.537$ & $2.0$ & $+0.2027$ & $0.690$ \\
4 days & $3.2$ & $+0.0909$ & $0.529$ & $3.2$ & $+0.0945$ & $0.703$ \\
7 days & $4.7$ & $+0.0774$ & $0.536$ & $4.7$ & $+0.0790$ & $0.703$ \\
10 days & $6.2$ & $+0.0726$ & $0.531$ & $6.3$ & $+0.0689$ & $0.702$ \\
20 days & $11.2$ & $+0.0670$ & $0.538$ & $11.2$ & $+0.0546$ & $0.702$ \\
30 days & $16.2$ & $+0.0685$ & $0.541$ & $16.2$ & $+0.0373$ & $0.704$ \\
60 days & $30.9$ & $+0.0651$ & $0.546$ & $30.9$ & $+0.0232$ & $0.716$ \\
Unrestricted & $236.4$ & $+0.0503$ & $0.796$ & $236.5$ & $+0.0230$ & $0.846$ \\
\bottomrule
\end{tabular}
\end{table}
At daily resolution the benefit is concentrated at the shortest, two-day supervision span
(Table~\ref{tab:supspan-wd1}). Between 4 and 60 days the single-resident series is nearly flat
($0.0651$--$0.0909$), and the two-day condition alone is $67\%$ above the four-day one. Only the
two-day arm reaches its untrained reference; every other cap in this table sits below it.
We report this as a description of where the benefit lies, not as validation of any rule for
choosing the supervision span.

\paragraph{The two-day condition is the only arm that reaches its untrained reference.} At five
seeds it gives $0.1515 \pm 0.0294$ against a floor of $0.1058 \pm 0.0081$ (gap $+0.0457$,
$95\%$ CI $[+0.0197, +0.0717]$, 5/5 seeds above) and $0.2027 \pm 0.0157$ against
$0.1524 \pm 0.0102$ (gap $+0.0503$, CI $[+0.0358, +0.0648]$, 5/5 seeds above). Every other cap
in either table sits at or below its floor.

These two gaps are scored at each arm's own embedding scale, and they do not survive matching it.
Contrastive training normalizes inside its loss and leaves $\lVert z \rVert$ free, so a trained
arm and an untrained one reach the decoder at different scales (Appendix~\ref{app:tnc}).
Re-scoring both after $L_2$ normalization leaves the two-day arm indistinguishable from the
untrained reference, $+0.0169$ $[-0.0104,+0.0449]$ and $+0.0082$ $[-0.0246,+0.0372]$, both
intervals containing zero, while unrestricted sampling
falls significantly \emph{below} it, $-0.1006$ $[-0.1230,-0.0801]$ and $-0.1074$
$[-0.1351,-0.0829]$. We therefore claim only that the tightest
supervision span matches what the architecture already carries, not that it exceeds it; the
result that survives scale matching is the suppression at wide $\tau$, not a gain at narrow
$\tau$. An earlier three-seed version of this sweep was optimistic
in the single-resident cohort ($0.1631$ at two days, falling to $0.1515$ once two further seeds
were added, with the standard deviation nearly doubling); Table~\ref{tab:supspan-wd1} is now
five-seed throughout. Under a stricter mean-plus-two-standard-deviation criterion single-resident
clears 4/5 seeds while the household cohort clears 5/5.

\subsection{Supervision span as an explicit parameter}
\label{app:tau}

Section~\ref{app:supspan-wd1} varies the cap in a grid chosen for the geometry. Here we treat
the maximum permitted separation $\tau$ as an explicit parameter and sweep it over
$\{2,4,8,16,32,64,128,\infty\}$ days at five seeds. Positives are drawn uniformly from
$\{x_{t'} : \text{same unit},\, 0<|t-t'|\le\tau\}$ subject to the leak-free rule. Every
condition receives the full record; anchors, negatives, splits, budget and evaluation are
identical, verified by hashing.

\emph{Prespecified hypothesis.} $\tau^\ast$ should approximate the timescale over which
behavioral state remains informative, as measured independently in
Section~\ref{non-exchangeable}, with broad $\tau$ hurting because positives stop sharing
state. We did not assume smaller is always better.

\begin{table}[!ht]
\centering
\caption{Changing-state information against the maximum permitted positive-pair separation, at
fixed record span. Realized mean separation is reported because the cap is an upper bound; no cap
produced an empty pool. $\times$floor is $\Delta I$ as a multiple of a same-architecture untrained
encoder: values above $1$ mean the objective adds information the architecture did not already
carry, values below $1$ mean it removes some. Every representation is $L_2$-normalized before
scoring, so trained and untrained arms are compared at matched embedding scale
(Appendix~\ref{app:tnc}).}
\label{tab:tau}
\small
\begin{tabular}{lcccccc}
\toprule
& \multicolumn{3}{c}{Single-resident} & \multicolumn{3}{c}{Households} \\
\cmidrule(lr){2-4}\cmidrule(lr){5-7}
$\tau$ & Realized & $\Delta I$ ($L_2$) & $\times$floor & Realized & $\Delta I$ ($L_2$) & $\times$floor \\
\midrule
2 days & $2.0$ & $+0.1211$ & $1.16$ & $2.0$ & $+0.1235$ & $1.07$ \\
4 days & $3.2$ & $+0.0444$ & $0.43$ & $3.2$ & $+0.0463$ & $0.40$ \\
8 days & $5.2$ & $+0.0368$ & $0.35$ & $5.2$ & $+0.0254$ & $0.22$ \\
16 days & $9.2$ & $+0.0292$ & $0.28$ & $9.2$ & $+0.0153$ & $0.13$ \\
32 days & $17.2$ & $+0.0272$ & $0.26$ & $17.2$ & $+0.0096$ & $0.08$ \\
64 days & $32.9$ & $+0.0130$ & $0.12$ & $32.9$ & $+0.0054$ & $0.05$ \\
128 days & $62.5$ & $+0.0110$ & $0.11$ & $62.8$ & $+0.0060$ & $0.05$ \\
Unrestricted & $243.4$ & $+0.0036$ & $0.03$ & $244.1$ & $+0.0079$ & $0.07$ \\
\midrule
Untrained floor & --- & $+0.1042$ & $1.00$ & --- & $+0.1153$ & $1.00$ \\
\bottomrule
\end{tabular}
\end{table}
\textbf{The hypothesis is not supported.} There is no interior optimum
(Table~\ref{tab:tau}). $\Delta I$ is maximized at the shortest attainable $\tau$ and declines
from $\tau=4$ onward, even though the independently measured availability curve still retains
about half its peak at 20 days. The single-resident series declines monotonically across all
eight caps; the household series declines monotonically through $\tau=64$ and is then flat to
within $0.003$ nats per dimension, so we describe it as monotone up to noise rather than
strictly monotone. Only $\tau=2$ reaches the untrained floor, and every wider cap sits below
it. Identity is flat from $\tau=4$ to $128$ ($0.520$--$0.586$ and
$0.696$--$0.727$) and rises only at unrestricted sampling ($0.791$, $0.847$), so within the
capped range state is lost without identity being gained.

\paragraph{Audits.} Candidate pools are strictly nested with sizes $1,5,13,29$ at
$\tau=2,4,8,16$; $\tau=2$ admits only $\{p-2\}$ because $p+2$'s block contains the target and
is excluded, which is geometry rather than an indexing error. Realized separations broaden
monotonically ($1,3,7,15,31,63,127,868$ distinct lags). Empirical selected-lag frequencies
match the pool-implied expectation to within $0.002$. No positive block contains the target in
$3{,}697{,}277$ checks. In \texttt{run()}, $\tau$ enters only through the positive array.

\paragraph{Why the shortest separation wins.} Two candidate explanations for
the two-day advantage were tested and rejected. It is not target leakage: excluding overlapping
positives leaves it at $0.1631$
against $0.1641$, despite $50\%$ of that pool having been contaminated. And it is not a
predictive effect arising from the positive being the next adjacent block, since pairing with
the immediately preceding, immediately following, or either block gives similar results
($+0.1203$, $+0.1050$, $+0.1240$ single-resident; $+0.1610$, $+0.1717$, $+0.1490$ households),
with the direction contrast changing sign between cohorts. What remains is that a positive
drawn from within a very short state-persistence interval shares local state with the anchor.
Appendix~\ref{app:pool-breadth} separates this from the pool-size confound the cap introduces.

\subsection{Fixed-cardinality control}
\label{app:fixcard}

Increasing $\tau$ enlarges both the separation and the cardinality of the equivalence class.
To separate them we give every anchor exactly \emph{one} positive at every $\tau$: the eligible
candidate whose lag is closest to $\tau$, so the selected separation moves outward with $\tau$
rather than collapsing to the nearest candidate. Realized separation then equals $\tau$ exactly.

\begin{table}[!ht]
\centering
\caption{Fixed cardinality (one positive per anchor) against cap-sampling, scale-matched and
paired on realized separation. Every value is $\Delta I$ after $L_2$-normalizing the
representation (Appendix~\ref{app:tnc}); the fixed-cardinality arm realizes a separation of
exactly $\tau$, so each row is paired with the cap-sampled arm whose realized separation is
closest, given in parentheses. Raw fixed-cardinality values are shown for continuity with the
earlier scoring.}
\label{tab:fixcard}
\small
\resizebox{\linewidth}{!}{%
\begin{tabular}{lcccccc}
\toprule
& \multicolumn{3}{c}{Single-resident} & \multicolumn{3}{c}{Households} \\
\cmidrule(lr){2-4}\cmidrule(lr){5-7}
$\tau$ & Fixed-card. ($L_2$ / raw) & Cap-sampled ($L_2$) & Gap & Fixed-card. ($L_2$ / raw) & Cap-sampled ($L_2$) & Gap \\
\midrule
2 days & $0.1211$ / $0.1644$ & $0.1211$ (2.0) & $+0.0000$ & $0.1235$ / $0.2065$ & $0.1235$ (2.0) & $+0.0000$ \\
4 days & $0.0757$ / $0.1171$ & $0.0444$ (3.2) & $+0.0314$ & $0.0898$ / $0.1552$ & $0.0463$ (3.2) & $+0.0435$ \\
8 days & $0.0599$ / $0.0849$ & $0.0292$ (9.2) & $+0.0307$ & $0.0452$ / $0.1165$ & $0.0153$ (9.2) & $+0.0298$ \\
16 days & $0.0444$ / $0.0774$ & $0.0272$ (17.2) & $+0.0173$ & $0.0300$ / $0.0895$ & $0.0096$ (17.2) & $+0.0204$ \\
32 days & $0.0283$ / $0.0549$ & $0.0130$ (32.9) & $+0.0153$ & $0.0265$ / $0.0676$ & $0.0054$ (32.9) & $+0.0211$ \\
64 days & $0.0219$ / $0.0582$ & $0.0110$ (62.5) & $+0.0110$ & $0.0209$ / $0.0479$ & $0.0060$ (62.8) & $+0.0149$ \\
128 days & $0.0204$ / $0.0505$ & $0.0110$ (62.5) & $+0.0094$ & $0.0139$ / $0.0390$ & $0.0060$ (62.8) & $+0.0079$ \\
\bottomrule
\end{tabular}}
\end{table}
\textbf{Separation hurts on its own.} With cardinality pinned at one and representation scale
matched, $\Delta I$ falls from $+0.1211$ to $+0.0204$ and from $+0.1235$ to $+0.0139$ between
$\tau=2$ and $\tau=128$ (Table~\ref{tab:fixcard}), that is by $83\%$ and $89\%$, and from
$1.16\times$ and $1.07\times$ the untrained floor to $0.20\times$ and $0.12\times$. The temporal
claim therefore does not depend on broader $\tau$ producing more heterogeneous pools.

\textbf{Cardinality adds to it in both cohorts.} Paired on realized separation and scored at
matched scale, the fixed-cardinality arm is above cap-sampling at every separation in both
cohorts, by $+0.0094$ to $+0.0314$ (single-resident) and $+0.0079$ to $+0.0435$ (households), so
restricting the pool helps beyond restricting separation. Pairing the arms at matched \emph{nominal} $\tau$ and scoring them raw instead makes the gap
turn negative at $\tau \ge 32$ in the single-resident cohort and the cardinality effect look
cohort-dependent. Both are artifacts of that pairing: cap-sampling realizes roughly $\tau/2$, so
the nominal pairing gives it the
shorter distance, and the two arms reach the decoder at different embedding scales
(Appendix~\ref{app:tnc}). Corrected, the two cohorts agree, which also resolves the tension with
the pool-size result of Appendix~\ref{app:pool-breadth}.

\paragraph{Independent cross-check.} The fixed-cardinality curve reproduces the separately
implemented $K{=}1$ sweep of Appendix~\ref{app:pool-breadth} at matched realized separation:
at four days, $0.1187$ against $0.1370$ (single-resident) and $0.1500$ against $0.1501$
(households); near 60 days, $0.0612$ against $0.0571$ and $0.0455$ against $0.0512$. Two
independently written constructions of the same question give the same qualitative decline.

\subsection{Distance-weighted positive sampling}
\label{app:softspan}

A hard cap is one way to keep supervision local; a smooth kernel over temporal distance is
another, and is what soft-assignment and time-dependent samplers for longitudinal time series use.
Replacing the cap with a kernel does not avoid the trade-off. Drawing positives with probability
proportional to $\exp(-|\Delta t|/\lambda)$ over the same candidates, changing-state information
falls with $\lambda$ exactly as it falls with $\tau$, and the two rules coincide when compared at
the separation they actually realize (Table~\ref{tab:soft-sampling}). At a realized separation
near $5$ days the kernel gives $0.0397$ against $0.0414$ for the cap in the single-resident cohort;
near $10$ days it gives $0.0301$ against the $0.0292$ of the reported cap sweep at $9.2$ days.
The same holds in households.

What the kernel cannot do is reproduce the tightest cap. An exponential kernel keeps a tail of
distant positives at every $\lambda$, so its realized separation never falls below $4.2$ days, and
no kernel arm reaches the untrained floor: the best is $0.38\times$ and $0.28\times$, against
$1.16\times$ and $1.07\times$ for the $2$-day cap. The quantity that governs the outcome is the
realized temporal reach of supervision, not whether the rule that produces it is a threshold or a
weighting.

\begin{table}[!ht]
\centering
\caption{Hard caps and distance-weighted sampling on the same build, with positives restricted to
the training era in both. A hard cap draws positives uniformly within $\tau$; exponential decay
draws them with probability proportional to $\exp(-|\Delta t|/\lambda)$ over the same candidates.
Realized separation is the draw-weighted mean over training anchors. $\times$floor is $\Delta I$
as a multiple of a same-architecture untrained encoder, scored after $L_2$ normalization.}
\label{tab:soft-sampling}
\small
\begin{tabular}{lcccc}
\toprule
Sampling rule & Realized sep.\ (d) & $\Delta I$ (raw) & $\Delta I$ ($L_2$) & $\times$floor \\
\midrule
\multicolumn{5}{l}{\emph{Single-resident}, untrained floor $0.1042$} \\
hard cap $\tau$ = 2\,d & $2.0$ & $0.1644$ & $0.1211$ & $1.16$ \\
hard cap $\tau$ = 8\,d & $5.2$ & $0.0693$ & $0.0414$ & $0.40$ \\
hard cap $\tau$ = 32\,d & $17.3$ & $0.0642$ & $0.0207$ & $0.20$ \\
hard cap $\tau$ = unrestricted & $141.9$ & $0.0470$ & $0.0083$ & $0.08$ \\
exponential decay $\lambda$ = 2\,d & $4.2$ & $0.0783$ & $0.0397$ & $0.38$ \\
exponential decay $\lambda$ = 8\,d & $10.2$ & $0.0642$ & $0.0301$ & $0.29$ \\
exponential decay $\lambda$ = 32\,d & $31.9$ & $0.0582$ & $0.0136$ & $0.13$ \\
exponential decay $\lambda$ = 128\,d & $83.9$ & $0.0541$ & $0.0103$ & $0.10$ \\
\multicolumn{5}{l}{\emph{Households}, untrained floor $0.1153$} \\
hard cap $\tau$ = 2\,d & $2.0$ & $0.2065$ & $0.1235$ & $1.07$ \\
hard cap $\tau$ = 8\,d & $5.2$ & $0.0763$ & $0.0262$ & $0.23$ \\
hard cap $\tau$ = 32\,d & $17.3$ & $0.0444$ & $0.0104$ & $0.09$ \\
hard cap $\tau$ = unrestricted & $142.0$ & $0.0236$ & $0.0068$ & $0.06$ \\
exponential decay $\lambda$ = 2\,d & $4.2$ & $0.0870$ & $0.0328$ & $0.28$ \\
exponential decay $\lambda$ = 8\,d & $10.2$ & $0.0529$ & $0.0138$ & $0.12$ \\
exponential decay $\lambda$ = 32\,d & $32.1$ & $0.0331$ & $0.0057$ & $0.05$ \\
exponential decay $\lambda$ = 128\,d & $84.9$ & $0.0224$ & $0.0056$ & $0.05$ \\
\bottomrule
\end{tabular}
\end{table}
\subsection{Target-overlap audit}
\label{app:target-overlap}

A history block spanning windows $[q-N_B, q)$ contains the prediction target $p$ whenever
$q-N_B \le p < q$. Under the candidate rule $|q-p| \ge N_B$, exactly one eligible positive
satisfies this, $q = p + N_B$. Such a positive is a legitimate same-person block, but the
encoder sees the target window among its training inputs.

This matters here because the eligible pool shrinks as the separation cap tightens, so the
overlapping share is strongly cap-dependent: $10.0\%$ at the 30-day cap, $4.6\%$ at 60 days,
$2.3\%$ at 120 days, $1.3\%$ at 240 days, and $0.7\%$ under unrestricted sampling. That
gradient runs in the same direction as the reported decline in changing-state information,
so it is a confound unusually well aligned with the hypothesis and cannot be dismissed on
the grounds that the absolute contamination is small.

We therefore repeated the sweep with every target-overlapping positive excluded from the
candidate pool, holding the panel, anchors, negatives, training budget, evaluation and seeds
fixed. Both builds were run in a single process so the comparison does not cross script
versions.

\begin{table}[!ht]
\centering
\caption{Span sweep with and without target-overlapping positive pairs. Brackets are 95\%
bootstrap intervals over units; five seeds per cell.}
\label{tab:target-overlap}
\small
\resizebox{\linewidth}{!}{%
\begin{tabular}{lcccc}
\toprule
& \multicolumn{2}{c}{Single-resident} & \multicolumn{2}{c}{Households} \\
\cmidrule(lr){2-3}\cmidrule(lr){4-5}
Cap & Original & Overlap-free & Original & Overlap-free \\
\midrule
30 days  & $+0.0664$ $[+0.0463,+0.0893]$ & $+0.0697$ $[+0.0496,+0.0919]$ & $+0.0396$ $[+0.0232,+0.0608]$ & $+0.0381$ $[+0.0229,+0.0578]$ \\
60 days  & $+0.0631$ $[+0.0444,+0.0836]$ & $+0.0614$ $[+0.0425,+0.0823]$ & $+0.0299$ $[+0.0151,+0.0495]$ & $+0.0320$ $[+0.0176,+0.0510]$ \\
120 days & $+0.0474$ $[+0.0328,+0.0637]$ & $+0.0511$ $[+0.0355,+0.0681]$ & $+0.0282$ $[+0.0138,+0.0476]$ & $+0.0269$ $[+0.0126,+0.0458]$ \\
240 days & $+0.0405$ $[+0.0270,+0.0550]$ & $+0.0413$ $[+0.0273,+0.0570]$ & $+0.0256$ $[+0.0107,+0.0459]$ & $+0.0211$ $[+0.0084,+0.0385]$ \\
Unrestricted & $+0.0243$ $[+0.0162,+0.0334]$ & $+0.0259$ $[+0.0172,+0.0353]$ & $+0.0138$ $[+0.0051,+0.0256]$ & $+0.0142$ $[+0.0056,+0.0259]$ \\
\midrule
Change & $-63\%$ & $-63\%$ & $-65\%$ & $-63\%$ \\
\bottomrule
\end{tabular}}
\end{table}

Removing these positives leaves the result essentially unchanged
(Table~\ref{tab:target-overlap}). Although the overlapping fraction decreases from $10.0\%$
at the shortest span to $0.7\%$ under unrestricted sampling, changing-state information
declines by $63\%$ in the single-resident cohort with and without the exclusion, and by
$65\%$ against $63\%$ in the household cohort. No individual cell moves by more than
$0.004$ nats per output dimension, every pair of intervals overlaps substantially, the
ordering across caps is preserved, and identity decodability is likewise unaffected
($0.619 \to 0.844$ originally against $0.627 \to 0.849$ after correction).

We report this as a robustness check rather than a correction, because the exclusion changes
no conclusion. We note two limits on its scope. First, the argument does not transfer
automatically to configurations with much shorter blocks: with daily windows the overlapping
share at the tightest cap reaches $50\%$, five times the worst case here, and that regime
requires its own control. Second, the analysis speaks only to blocks that contain the target
window; positives adjacent to but not overlapping the target remain in the pool by design,
since excluding them would confound the separation cap with the eligible pool size.

\paragraph{Positives drawn only from the training era.} A second eligibility question concerns the
test period rather than the target window. In the primary sweep, candidate positives are drawn
from a participant's whole record, so at wide caps a training anchor may be paired with a window
that lies after the train/test cut. No target is exposed, but the encoder then trains on
test-period inputs. Repeating the sweep with candidates confined to the training era leaves the
mechanism intact and moves one reported quantity. The tightest cap is unchanged, since a $2$-day
cap never reaches across the cut, and reproduces its reported value exactly ($0.1211$ and
$0.1235$, $1.16\times$ and $1.07\times$ the untrained floor). Under unrestricted pairing the
retained information is $0.08\times$ and $0.06\times$ the floor rather than $0.03\times$ and
$0.07\times$, so what survives at the broadest supervision span is between $6\%$ and $8\%$ of an
untrained encoder (Table~\ref{tab:soft-sampling}). The qualitative claim is unaffected: only the
tightest cap matches the floor, and every wider cap sits far below it.

\subsection{Separation versus pool breadth}
\label{app:pool-breadth}

A separation cap changes two things at once: how far a positive may sit from the anchor, and
how many candidates are eligible. The supervision-span sweep therefore does not isolate
separation. We cross the two factors, drawing $K$ before-anchor candidates from a band centered
on a target separation. Direction is fixed to before-anchor because it proved immaterial
($+0.0153$ single-resident, $-0.0107$ households, overlapping intervals).

\paragraph{The nominal design does not hold separation fixed at large $K$.} When $K$ is large
and the target separation short, too few candidates exist nearby and the band widens. Realized
separation therefore drifts with $K$: at a nominal four days it is $4.0$, $4.0$, $6.5$ and
$16.4$ days for $K = 1, 3, 10, 30$ (single-resident; $16.5$ at $K{=}30$ for households, where
the four- and ten-day targets both collapse to the same realized $16.5$ days). We report
realized separations throughout and restrict inference about $K$ to the nominal 30- and 60-day
rows, where the spread across $K$ is under two days.

\begin{table}[!ht]
\centering
\caption{Changing-state information by pool size $K$ and \emph{realized} mean separation
(parentheses), daily windows. Comparisons across $K$ are interpretable only at nominal 30 and
60 days; elsewhere band widening confounds $K$ with separation.}
\label{tab:pool-breadth}
\small
\resizebox{\linewidth}{!}{%
\begin{tabular}{lccccc}
\toprule
& 4 days & 10 days & 30 days & 60 days & 120 days \\
\midrule
\multicolumn{6}{l}{\emph{Single-resident}} \\
$K=1$  & $+0.1370$ ($4.0$) & $+0.0849$ ($10.0$) & $+0.0684$ ($30.0$) & $+0.0571$ ($58.6$) & $+0.0428$ ($97.4$) \\
$K=3$  & $+0.1258$ ($4.0$) & $+0.0993$ ($10.0$) & $+0.0812$ ($30.0$) & $+0.0773$ ($58.5$) & $+0.0673$ ($97.3$) \\
$K=10$ & $+0.0938$ ($6.5$) & $+0.0780$ ($9.5$)  & $+0.0823$ ($29.5$) & $+0.0613$ ($57.7$) & $+0.0737$ ($99.5$) \\
$K=30$ & $+0.0782$ ($16.4$)& $+0.0748$ ($16.4$) & $+0.0653$ ($29.2$) & $+0.0518$ ($56.6$) & $+0.0650$ ($103.9$) \\
\midrule
\multicolumn{6}{l}{\emph{Households}} \\
$K=1$  & $+0.1501$ ($4.0$) & $+0.1060$ ($10.0$) & $+0.0883$ ($30.0$) & $+0.0512$ ($58.7$) & $+0.0547$ ($99.0$) \\
$K=3$  & $+0.1628$ ($4.0$) & $+0.1368$ ($10.0$) & $+0.1035$ ($30.0$) & $+0.0582$ ($58.6$) & $+0.0757$ ($98.9$) \\
$K=10$ & $+0.0816$ ($6.5$) & $+0.0688$ ($9.5$)  & $+0.0490$ ($29.5$) & $+0.0414$ ($57.9$) & $+0.0643$ ($100.9$) \\
$K=30$ & $+0.0333$ ($16.5$)& $+0.0439$ ($16.5$) & $+0.0317$ ($29.2$) & $+0.0374$ ($56.8$) & $+0.0452$ ($104.9$) \\
\bottomrule
\end{tabular}}
\end{table}

\paragraph{The separation effect replicates.} At $K{=}1$, where realized separation matches
its target exactly, $\Delta I$ falls from four days to roughly 100 days by $69\%$
($0.1370 \to 0.0428$) and $64\%$ ($0.1501 \to 0.0547$). The decline holds in 3/3 seeds in both
cohorts, with paired differences of $-0.0942$ $[-0.1038, -0.0847]$ and $-0.0955$
$[-0.1053, -0.0856]$. Within-$K$ Spearman correlations against realized separation are
$-1.00$ and $-0.90$ at $K{=}1$ and $K{=}3$ in the two cohorts.

\paragraph{The pool-size effect does not replicate.} Restricting to the rows where separation
is matched across $K$, the single-resident cohort shows no effect of pool size
($\rho(\Delta I, K) = -0.20$ and $-0.40$; $-5\%$ and $-9\%$ from $K{=}1$ to $K{=}30$;
non-monotone), whereas the household cohort shows a substantial one
($\rho = -0.80$ and $-0.80$; $-64\%$ and $-27\%$). We therefore do not claim that pool breadth
is inert. What replicates is the separation effect; whether widening the pool additionally
harms the representation is cohort-dependent in our data, and we report it as unresolved.
Appendix~\ref{app:fixcard} settles the complementary direction with a five-seed control in
which every anchor receives exactly one positive at every separation: increasing temporal
separation is sufficient to reduce changing-state information even when positive cardinality
is fixed.

\begin{figure}[!ht]
\centering
\includegraphics[width=\linewidth]{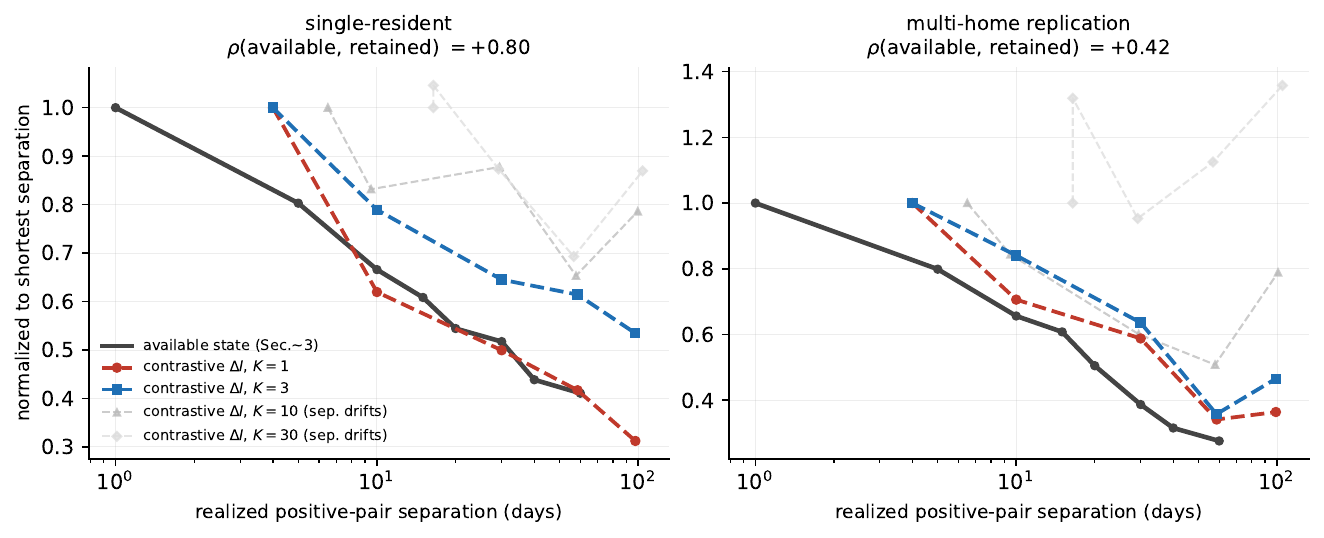}
\caption{Available state (Section~\ref{non-exchangeable}) and retained state
(Section~\ref{sec:supervision-span}) against realized positive-pair separation, each normalized to its own
shortest separation. The narrow-pool curves ($K{=}1$, $K{=}3$) track availability closely.
The $K{=}10$ and $K{=}30$ curves are greyed because band widening moves their starting
separation ($6.5$ and $16.4$ days rather than $4$), so their normalization base differs and
their apparent upturn past 60 days is an artifact of that. In the household
cohort the $K{=}30$ row has its first two cells at the same realized separation
($16.5$ days), which is what depresses that panel's pooled correspondence to $+0.42$.}
\label{fig:overlay}
\end{figure}

\paragraph{Retained state tracks available state.} Interpolating the availability curve of
Section~\ref{non-exchangeable} at each cell's realized separation gives a Spearman
correspondence with contrastive $\Delta I$ of $+0.80$ and $+0.42$ across all twenty cells, and
$+1.00$ and $+0.90$ within the narrow-pool rows; the pooled household value is depressed by the
large-$K$ rows, whose realized separations are compressed by band widening. These are descriptive,
since the twenty cells share anchors, encoders and an evaluation set. At $K{=}1$ the contrastive
curve tracks availability to about 60 days and then falls below it ($0.31$ against $0.44$ of its
short-separation value at 97 days), so the representation loses state somewhat faster than the
data does (Fig.~\ref{fig:overlay}).

\begin{table}[!ht]
\centering
\caption{Values plotted in Figure~\ref{fig:overlay}. Availability is the Section~\ref{non-exchangeable}
distance sweep; contrastive values are plotted at their realized separation, given in
parentheses. Each series is normalized to its own shortest separation.}
\label{tab:overlay-values}
\small
\resizebox{\linewidth}{!}{%
\begin{tabular}{llccccc}
\toprule
Cohort & Series & \multicolumn{5}{c}{normalized value at realized separation (days)} \\
\midrule
\multirow{5}{*}{Single-resident}
 & Availability & $1.000$ ($0$) & $0.666$ ($10$) & $0.517$ ($30$) & $0.411$ ($60$) & --- \\
 & $K=1$  & $1.000$ ($4.0$)  & $0.620$ ($10.0$) & $0.499$ ($30.0$) & $0.417$ ($58.6$) & $0.312$ ($97.4$) \\
 & $K=3$  & $1.000$ ($4.0$)  & $0.789$ ($10.0$) & $0.645$ ($30.0$) & $0.614$ ($58.5$) & $0.535$ ($97.3$) \\
 & $K=10$ & $1.000$ ($6.5$)  & $0.832$ ($9.5$)  & $0.878$ ($29.5$) & $0.653$ ($57.7$) & $0.786$ ($99.5$) \\
 & $K=30$ & $1.000$ ($16.4$) & $0.957$ ($16.4$) & $0.835$ ($29.2$) & $0.662$ ($56.6$) & $0.831$ ($103.9$) \\
\midrule
\multirow{5}{*}{Households}
 & Availability & $1.000$ ($0$) & $0.656$ ($10$) & $0.387$ ($30$) & $0.276$ ($60$) & --- \\
 & $K=1$  & $1.000$ ($4.0$)  & $0.706$ ($10.0$) & $0.588$ ($30.0$) & $0.341$ ($58.7$) & $0.364$ ($99.0$) \\
 & $K=3$  & $1.000$ ($4.0$)  & $0.841$ ($10.0$) & $0.636$ ($30.0$) & $0.358$ ($58.6$) & $0.465$ ($98.9$) \\
 & $K=10$ & $1.000$ ($6.5$)  & $0.844$ ($9.5$)  & $0.601$ ($29.5$) & $0.508$ ($57.9$) & $0.789$ ($100.9$) \\
 & $K=30$ & $1.000$ ($16.5$) & $1.319$ ($16.5$) & $0.953$ ($29.2$) & $1.125$ ($56.8$) & $1.358$ ($104.9$) \\
\bottomrule
\end{tabular}}
\end{table}

Table~\ref{tab:overlay-values} gives the plotted values. Raw (unnormalized) $\Delta I$ for
every cell appears in Table~\ref{tab:pool-breadth}; the availability series is the canonical
column of Table~\ref{tab:resid-sweep}.

The daily-window geometry uses two-day blocks, so absolute values are not comparable to the
five-day-window experiments elsewhere.

\section{Within-Person Discrimination}
\label{app:withinperson}

Introducing within-person negatives at full record span restores changing-state information to
approximately the short-record level while preserving strong identity decodability. The recovery
does not require temporally meaningful groups: random partitions work similarly, whereas
partitioning without within-person discrimination does not. The intervention therefore acts
through within-person discrimination rather than segmentation itself.

\subsection{Construction}
\label{app:segment-construction}

Only the contrastive label changes. In the unsegmented objective, the positive identity is the
participant, so every same-person observation is excluded from an anchor's negatives however far
apart in time the two lie. We instead assign each window a group and take the positive identity to
be the (participant, group) pair, so a participant's windows from \emph{other} groups enter the
denominator as negatives. An anchor's own positives are always excluded from its negatives,
whichever group they fall in.

Everything else is the nested-history construction of
Appendix~\ref{app:nested-construction} at the full record: the same anchors, the same locally
capped positives with realized separation near $20$ days, the same per-unit anchor budget,
schedule, encoder, optimizer and evaluation (Appendix~\ref{app:implementation}). Groups are
contiguous $30$-, $60$- or $120$-day spans of a participant's record, which yields $239$ to $987$
identities in place of $42$ or $54$.

\subsection{State recovery and identity preservation}
\label{app:segment-results}

At the full record, within-person discrimination recovers the entire record-span loss
(Table~\ref{tab:segment-identity}). Against the unsegmented objective, $\Delta I$ rises by
$+0.0529$ and $+0.0641$ with $30$-day groups, and every interval over the three group widths
excludes zero. Measured against the $30$-day record rather than the full-record baseline, the
single-resident arms are statistically indistinguishable from the short record ($+0.0063$
$[-0.0078,+0.0191]$ at $30$-day groups), and the household arms sit slightly above it ($+0.0153$
$[+0.0049,+0.0256]$), so more history with within-person negatives is at least as informative as
less history without them. Group width barely matters: $30$, $60$ and $120$ days give $0.1474$,
$0.1454$ and $0.1453$ in the single-resident cohort.

The recovery costs little identity. Identity accuracy falls from $0.613$ to $0.578$--$0.587$ and
from $0.757$ to $0.730$--$0.736$, remaining far above chance ($0.019$ and $0.024$), so these
encoders are still strongly person-identifiable. Expressed as a trade-off, each point of identity
accuracy given up buys $0.015$ to $0.019$ nats per dimension of changing-state information per
percentage point in the single-resident cohort, and $0.023$ to $0.030$ in households.

\subsection{Random-group and masking controls}
\label{app:segment-controls}

Two controls separate discrimination from segmentation.

\paragraph{Random groups recover as much as contiguous ones.} Assigning windows to groups at
random within each participant, preserving the number of identities and their sizes but destroying
any temporal meaning, gives $+0.0530$ and $+0.0580$ against the unsegmented objective, which
matches the contiguous arms. A group whose members are scattered across the record cannot break a
temporal chain, so the gain does not come from segmenting the trajectory into coherent periods. It
comes from the fact that a participant's distant windows now repel.

\paragraph{Masking cross-group pairs is an equivalence check.} Excluding cross-group pairs from
the loss entirely, neither positive nor negative, reproduces the unsegmented objective exactly
($\Delta I$ difference $+0.0000$, zero-width interval). This is a sanity check rather than an
independent result: the unsegmented objective already masks every same-person pair, so the two
losses are the same objective. It is worth stating because it pins down what the intervention
adds. In the unsegmented loss no term links a participant's distant windows, yet their
representations converge as the record grows (Appendix~\ref{app:nested-geometry}); only the
explicit repulsive term reverses that.

\paragraph{The intervention is specific to long records.} Applying the same relabeling at
$L{=}30$, with $15$-day groups, does not help: $-0.0090$ $[-0.0153,-0.0025]$ in the
single-resident cohort and $+0.0026$ $[-0.0023,+0.0074]$ in households, while identity falls from
$0.546$ to $0.509$ and from $0.668$ to $0.632$. Within-person negatives are therefore not a
generic improvement to the pretext task; they repair something that only arises once the record is
long.

\subsection{Representation geometry}
\label{app:segment-geometry}

The geometry moves back with the information (Table~\ref{tab:segment-geometry}). The
between-to-within ratio falls from $3.10$ to $2.76$--$2.79$ in the single-resident cohort and from
$4.74$ to $3.40$--$3.51$ in households, the latter returning to its $30$-day value of $3.55$. The
contraction of never-paired distant states reverses in the same way: the distance between a
participant's own windows more than $180$ days apart, relative to between-participant distance,
recovers from $0.576$ to $0.604$--$0.608$ against a $30$-day reference of $0.637$, and in
households from $0.458$ to $0.545$--$0.552$ against a reference of $0.539$.

The reversal is complete in households and partial in the single-resident cohort, where $\Delta I$
returns fully while the geometry recovers about half of its shift. Within-person discrimination
therefore does not simply undo the record-span effect; it changes what the encoder spends capacity
on, and the state readout benefits more than the coarse geometry summary suggests.

\begin{table}[!ht]
\centering
\caption{Within-person discrimination at the full record. Only the contrastive label changes:
the positive identity is (participant, group) rather than participant, so a participant's windows
from other groups become negatives while an anchor's own positives are always excluded. Positives,
their separation, the anchors, the schedule and the evaluation are those of
Appendix~\ref{app:nested-construction}. Contrasts are paired unit bootstraps against the
unsegmented full-record run and against the $30$-day record.}
\label{tab:segment-identity}
\small
\resizebox{\linewidth}{!}{%
\begin{tabular}{lccccc}
\toprule
Contrastive label & Identities & $\Delta I$ & Identity & vs.\ unsegmented & vs.\ $L{=}30$ \\
\midrule
\multicolumn{6}{l}{\emph{Single-resident}} \\
person identity (unsegmented) & $54$ & $0.0944$ & $0.613$ & $+0.0000$ $[+0.0000,+0.0000]$ & $-0.0466$ $[-0.0638,-0.0317]$ \\
cross-segment pairs masked & $524$ & $0.0944$ & $0.613$ & $+0.0000$ $[+0.0000,+0.0000]$ & $-0.0466$ $[-0.0638,-0.0317]$ \\
segments, 30 d & $987$ & $0.1474$ & $0.578$ & $+0.0529$ $[+0.0410,+0.0661]$ & $+0.0063$ $[-0.0078,+0.0191]$ \\
segments, 60 d & $524$ & $0.1454$ & $0.583$ & $+0.0510$ $[+0.0393,+0.0635]$ & $+0.0043$ $[-0.0096,+0.0170]$ \\
segments, 120 d & $300$ & $0.1453$ & $0.587$ & $+0.0509$ $[+0.0392,+0.0635]$ & $+0.0042$ $[-0.0096,+0.0172]$ \\
random groups, matched sizes & $664$ & $0.1475$ & $0.581$ & $+0.0530$ $[+0.0413,+0.0659]$ & $+0.0064$ $[-0.0074,+0.0191]$ \\
segments at $L{=}30$, 15 d & $1898$ & $0.1320$ & $0.509$ & --- & $-0.0090$ $[-0.0153,-0.0025]$ \\
\multicolumn{6}{l}{\emph{Households}} \\
person identity (unsegmented) & $42$ & $0.0539$ & $0.757$ & $+0.0000$ $[+0.0000,+0.0000]$ & $-0.0488$ $[-0.0641,-0.0356]$ \\
cross-segment pairs masked & $429$ & $0.0539$ & $0.757$ & $+0.0000$ $[+0.0000,+0.0000]$ & $-0.0488$ $[-0.0641,-0.0356]$ \\
segments, 30 d & $808$ & $0.1179$ & $0.730$ & $+0.0641$ $[+0.0519,+0.0778]$ & $+0.0153$ $[+0.0049,+0.0256]$ \\
segments, 60 d & $429$ & $0.1189$ & $0.735$ & $+0.0651$ $[+0.0527,+0.0789]$ & $+0.0163$ $[+0.0062,+0.0265]$ \\
segments, 120 d & $239$ & $0.1058$ & $0.736$ & $+0.0519$ $[+0.0423,+0.0623]$ & $+0.0032$ $[-0.0083,+0.0141]$ \\
random groups, matched sizes & $535$ & $0.1118$ & $0.735$ & $+0.0580$ $[+0.0473,+0.0697]$ & $+0.0092$ $[-0.0014,+0.0194]$ \\
segments at $L{=}30$, 15 d & $1552$ & $0.1051$ & $0.632$ & --- & $+0.0026$ $[-0.0023,+0.0074]$ \\
\bottomrule
\end{tabular}}
\end{table}
\begin{table}[!ht]
\centering
\caption{Geometry under within-person discrimination, $L_2$-normalized, on the fixed evaluation
anchors. ``Between/within'' is the ratio of between-participant centroid dispersion to
within-participant spread. The last column is the mean distance between a participant's own
windows more than $180$ days apart, divided by the mean distance to other participants' windows;
those pairs are never positives in any arm.}
\label{tab:segment-geometry}
\small
\begin{tabular}{lcccc}
\toprule
Arm & $\Delta I$ & Identity & Between/within & Own windows $>180$\,d \\
\midrule
\multicolumn{5}{l}{\emph{Single-resident}} \\
$L{=}30$ reference & $0.1410$ & $0.546$ & $2.397$ & $0.6366$ \\
full record, unsegmented & $0.0944$ & $0.613$ & $3.097$ & $0.5756$ \\
full record, segments 30 d & $0.1474$ & $0.578$ & $2.759$ & $0.6078$ \\
full record, segments 60 d & $0.1454$ & $0.583$ & $2.756$ & $0.6062$ \\
full record, segments 120 d & $0.1453$ & $0.587$ & $2.789$ & $0.6042$ \\
\multicolumn{5}{l}{\emph{Households}} \\
$L{=}30$ reference & $0.1025$ & $0.668$ & $3.548$ & $0.5394$ \\
full record, unsegmented & $0.0539$ & $0.757$ & $4.738$ & $0.4579$ \\
full record, segments 30 d & $0.1179$ & $0.730$ & $3.424$ & $0.5494$ \\
full record, segments 60 d & $0.1189$ & $0.735$ & $3.405$ & $0.5524$ \\
full record, segments 120 d & $0.1058$ & $0.736$ & $3.507$ & $0.5453$ \\
\bottomrule
\end{tabular}
\end{table}
\section{Prospective Replication in GLOBEM}
\label{app:globem-appendix}
\label{app:globem}

This appendix tests one mechanism prospectively: that broadening the temporal reach of
positive-pair supervision suppresses changing-state information. It does so on GLOBEM
\citep{xu2022globem}, a credentialed-access phone-sensing dataset from a different population
and sensing setting, under a protocol fixed by sha256 before the data it governs
was opened, with the hashes asserted at runtime. Broadening the cap reduces changing-state
information under both prespecified readouts, with $5/5$ seeds agreeing in sign.

The test is deliberately narrow. GLOBEM records are $60$--$97$ days, so the record-span
manipulation of Appendix~\ref{app:recordspan} cannot move supervision far enough to be
informative there, and we report it as null rather than as a failure to replicate. The
within-person discrimination result of Appendix~\ref{app:withinperson} is not tested here at all.
What transfers is the supervision-span mechanism itself.

\subsection{GLOBEM: selecting variables that can carry state}
\label{app:globem-w1}

The manipulation can only be tested on variables that carry temporally persistent state, and in
this dataset most do not. Ranking \texttt{:allday} columns by coverage, the obvious default,
yields a set with median within-person lag-1 autocorrelation $0.169$, median half-life $1.8$ days,
$20$ of $48$ variables effectively white, and a median of $2.49\%$ of baseline test error
removable from a two-day context. A manipulation of supervision span cannot damage persistent
state information in variables that carry almost none, and on that set the result is
uninformative in both directions: the trained encoder scores $+0.0080$ nats/dim against
PCA-$32$'s $+0.3727$, and the record-span arms are inconclusive. Coverage is orthogonal to
persistence, so the selection rule below ranks on persistence and predictability instead, fitted
on one wave and applied to another.

\subsection{GLOBEM: a prospective state-feature rule}
\label{app:globem-rule}

We therefore fixed a selection rule on Wave~1 that uses no SSL result, no learned
representation and no arm --- only a variance decomposition, a within-person autocorrelation and
a univariate linear $r^2$. None of these is computable from, or correlated by construction with,
the contrastive-versus-PCA contrast that the test evaluates. Per feature, on the rows where it is
present, participants with at least $60$ such days, windows calendar-contiguous:
\begin{quote}
\textbf{C1} within-participant variance $>0$ for $\ge 90\%$ of participants;\\
\textbf{C2} within-person lag-1 autocorrelation $\ge 0.30$;\\
\textbf{C3} half-life in $[1.5,14.0]$ days;\\
\textbf{C4} $r^2 \ge 0.10$.
\end{quote}
$r^2$ is the fraction of baseline test error removed, on Wave~1's own chronological split, by
ridge from $[B_{ij}, x_{p-1,j}, x_{p-2,j}]$ against $B_{ij}$ alone. It is univariate by
construction, so a feature's score does not depend on which others are in the set. Survivors are
ranked by $r^2$ and the top $48$ taken, to match the DETECT feature width; thresholds are not
relaxed if fewer survive.

Of 612 admissible \texttt{:allday} columns, 174 met the dtype and coverage
preconditions and 46 passed all four criteria. The resulting set is multi-channel
(24 Bluetooth, 20 location, 2 screen) where the coverage-ranked set had been pure
location, and it raises the regime from a median 2.49\% of error removable to
26.3\%.

The rule was frozen on Wave~1 and the test run on Wave~2, which had never been read. Waves~3
and~4 remain untouched and are not used under any outcome.

\subsection{The rule transfers}

Recomputed on Wave~2's 199 participants, reported and not used to gate anything:
lag-1 autocorrelation median $0.409$ with $41$ of $46$ features still meeting C2;
half-life median $2.2$ days with $46$ of $46$ still inside the C3 band; error removed
median $26.9\%$ with $43$ of $46$ still meeting C4. The rule was fitted on one cohort and the
variables it selected still instantiate the regime in an unseen one.

\subsection{Result}

Encoder, optimization and sampling follow Appendix~\ref{app:implementation}, with the encoder
width re-solved for this panel ($h=147$, $40{,}163$ parameters). All representations are
$L_2$-normalized before comparison, for the reason given in Appendix~\ref{app:tnc}. Estimands are
participant-weighted, and intervals resample participants and seeds jointly
(Appendix~\ref{app:bootstrap}).

\paragraph{Levels.} Every representation removes between $12.8\%$ and $21.6\%$ of
baseline test error except the person-contrastive encoder, which removes $0.14\%$
(Table~\ref{tab:globem-levels}). It is also the only identity-dominant one, at $0.779$
participant decodability against $0.159$ for masked reconstruction and $0.085$ for an
untrained encoder of the same shape, on identical anchors at identical width and budget. The
regime gate fails as it did on Wave~1: contrastive minus PCA-$32$ is $-0.1479$ $[-0.1775,-0.1212]$.

Because the selected features demonstrably carry short-horizon state
(Section~\ref{app:globem-rule}), this comparison is not subject to the objection that the
variables have no state to lose.

\paragraph{The direct supervision-span manipulation replicates.} Capping the maximum same-person
positive separation and widening it from $3.2$ to $24.6$ realized days reduces state
information under both prespecified readouts:

\begin{quote}
primary, conditioned $\Delta I$: $-0.0070$ $[-0.0121,-0.0033]$, $5/5$ seeds \\
secondary, error removed beyond $B_i$: $-0.0064$ $[-0.0120,-0.0026]$, $5/5$ seeds
\end{quote}

The readouts were required in advance to agree in sign, and they do. Across the grid, state
information falls while identity rises on the same axis (Table~\ref{tab:globem-tau}).

\paragraph{Limits.} First, the absolute state information captured by contrastive SSL is small
throughout. The decline runs from $0.80\%$ of baseline error
removed at the tightest cap to $0.14\%$ at unrestricted, against PCA-$32$'s $20.31\%$
on the same anchors. It is a large relative loss on a base that is a few percent of what a linear
compression extracts.

Second, the record-span arms do not replicate on GLOBEM: $-0.0013$ $[-0.0056,+0.0033]$ for record span with
supervision span free, $-0.0028$ $[-0.0076,+0.0023]$ with it pinned, and $+0.0002$ $[-0.0009,+0.0013]$ for the dissociation itself, all
intervals containing zero. The realized separations show why. GLOBEM records are $60$--$97$ days,
so growing the record moves mean positive separation only $14.4$ to $24.5$ days, a factor
of $1.7$, where the direct cap moves it a factor of $7.7$. This is the outcome our
prespecification anticipated for short records. Supervision span transfers; record span as a
proxy for it does not. Section~\ref{sec:supervision-span} draws that distinction, and GLOBEM tests it where
the two variables come apart.

\subsection{GLOBEM: single-positive control}
\label{app:globem-k1}

Widening the cap also widens the pool of admissible positives, from a minimum of $4$ candidates at
the tightest cap to $32$ at the widest, so separation and cardinality move together as they do on
DETECT (Appendix~\ref{app:fixcard}). Repeating the sweep with exactly one positive per anchor, the
eligible candidate whose lag is closest to $\tau$, holds cardinality fixed. Changing-state
information still falls with separation: $+0.0197$, $+0.0141$, $+0.0091$ and $+0.0075$ at $\tau =
4$, $8$, $16$ and $32$ days, a $62\%$ decline with one positive throughout.

The unrestricted cell of this control is not comparable to the others and we report it separately.
With a single positive, ``unrestricted'' selects the \emph{farthest} eligible candidate rather
than sampling uniformly, giving a realized separation of $51.2$ days and $\Delta I = +0.0154$,
above the $\tau=32$ cell. One distant positive is therefore less damaging here than four
moderately distant ones, which indicates that pool breadth contributes in this cohort as well as
separation.

\subsection{Qualifications}

\textbf{The person baseline is nearly inert on Wave~2.} $\mathrm{NLL}(y \mid 1) = 1.4770$
against $\mathrm{NLL}(y \mid B_i) = 1.4799$, so conditioning on the participant's
training-period mean is marginally \emph{worse} than an intercept. With a within-person variance
share of $0.851$ and $21$ of $46$ features vendor person-normalized, there is little in the
person mean to go beyond. This makes $\Delta I$ cleaner to read but means the phrase
``incremental beyond $B_i$'' carries less here than in DETECT.

\textbf{Pool breadth is confounded with separation.} Minimum admissible positives rise
$4 / 5 / 7 / 20 / 32$ across the grid, so a wider cap also supplies a more varied positive pool. DETECT's
design carries the same confound (Appendix~\ref{app:pool-breadth}), so the comparison is fair, but the
arm cannot separate ``positives further away'' from ``positives more varied''.

\textbf{Feature redundancy.} The selected block has an effective rank of $12.3$ of $46$, with
$22$ of $1{,}035$ pairs above $|r| = 0.95$; the mechanical rule admitted several near-duplicate
Bluetooth variants. The set is not edited after the fact, but the effective width is closer to
$12$ than to $46$.

\begin{table}[t]
\centering\small
\caption{GLOBEM Wave~2, prospective state-feature set. All representations $L_2$-normalized;
$r^2$ is the percentage of baseline test error removed beyond $B_i$; identity is $199$-way
participant decodability (chance $0.0050$). $8038$ anchors, $199$ participants, $2719$ test rows.}
\label{tab:globem-levels}
\begin{tabular}{lrrr}
\toprule
Representation & $\Delta I$ & $r^2$ (\%) & Identity \\
\midrule
Raw block ($92$-d) & $+0.1666$ & $21.57$ & --- \\
PCA-$32$ & $+0.1502$ & $20.31$ & --- \\
Masked reconstruction & $+0.1360$ & $18.59$ & $0.159$ \\
Random projection-$32$ & $+0.1213$ & $16.86$ & --- \\
Predictive & $+0.1222$ & $13.36$ & --- \\
Untrained encoder-$32$ & $+0.0910$ & $12.79$ & $0.085$ \\
Person-contrastive & $+0.0023$ & $0.14$ & $0.779$ \\
\bottomrule
\end{tabular}
\end{table}

\begin{table}[t]
\centering\small
\caption{GLOBEM Wave~2 supervision-span sweep. State information falls and identity rises on the
same axis. Separations are realized, not nominal.}
\label{tab:globem-tau}
\begin{tabular}{lrrr}
\toprule
Cap (realized sep., d) & $\Delta I$ & $r^2$ (\%) & Identity \\
\midrule
$3.2$ & $+0.0093$ & $0.80$ & $0.703$ \\
$5.2$ & $+0.0066$ & $0.42$ & --- \\
$8.8$ & $+0.0078$ & $0.48$ & $0.724$ \\
$15.2$ & $+0.0045$ & $0.33$ & --- \\
$24.6$ & $+0.0023$ & $0.14$ & $0.779$ \\
\bottomrule
\end{tabular}
\end{table}
\section{Robustness Across Models and Readouts}
\label{app:robustness}

The analyses in this appendix use the fixed-anchor build of
Appendix~\ref{app:record-span-builds}, and all are scored through the same anchors, splits,
person baselines $B_i$, targets, ridge grid and participant bootstrap as the runs they are
compared against. Confidence intervals are unit-clustered bootstraps; where two arms are
compared the bootstrap is \emph{paired}, with the same resample applied to both.

\subsection{Motion-only replication of the scale effect}
\label{app:motion-only-scale}
We next repeat the principal record-span and supervision-span experiments
after rebuilding the panel using only the 30 motion-derived features. This
provides a stricter test of whether changing sensor availability over long
records drives the scale effect. The state-to-identity shift persists. Under
person-level contrastive learning, increasing record span from 30 days to the
full history reduces changing-state information from \(0.0695\) to \(0.0382\)
nats/dim (\(-45\%\)) in the single-resident cohort, while identity accuracy
increases from \(0.475\) to \(0.674\). The household cohort shows the same
direction: changing-state information falls by \(31\%\) while identity accuracy
increases by \(0.132\). Predictive representations remain stable or improve
(\(+21\%\) and \(+4\%\)), while masked reconstruction changes little
(\(-9\%\) in both cohorts), preserving the objective-specific ordering observed
with the full feature set.

The supervision-span intervention isolates the mechanism more directly. With
the full record fixed, broadening same-person supervision from approximately
10 days to unrestricted pairing reduces changing-state information from
\(0.0755\) to \(0.0378\) nats/dim (\(-50\%\)) in the single-resident cohort and
by \(53\%\) in the household cohort, while identity decodability increases
by \(0.241\) and \(0.117\), respectively. In contrast, when positive-pair
separation is held near 21 days, changing-state information changes by only
\(-2\%\) and \(+3\%\), with identity accuracy changing by at most \(0.002\).
Thus, the scale-dependent representational shift persists in features that
cannot encode sensor availability, and in this construction the larger decline
tracks the temporal separation of positive pairs. The comparison is run on the
fixed-anchor build, in which the encoder already has the whole record at every
$L$ (Appendix~\ref{app:record-span-builds}), so it speaks to supervision reach
and not to how much history is available.

\subsection{Nonlinear predictive decoder}
\label{app:mlp-decoder}

Changing state is read out with a ridge decoder while identity is read out with
a logistic probe. The two readouts differ in capacity, so the divergence between
them could in principle describe the decoders rather than the representations.
This appendix re-estimates $\Delta I$ with a nonlinear decoder, changing nothing
else.

\paragraph{Protocol parity.}
The decoder is a two-hidden-layer MLP. Parity with the ridge path is the point of
the comparison, so: its hyperparameters (hidden width, weight decay) are selected
on the \emph{same} validation split on which the ridge penalty is selected, by the
same criterion of validation mean squared error; the predictive variance is taken
from validation residuals exactly as in the ridge path; and the two arms of every
comparison, $B_i$ alone and $[B_i, z]$, receive independent selection, so the
richer feature set cannot win by being handed a better-tuned decoder. Inputs are
standardized on fitting rows only. Where the estimator is cross-validated, both
decoder families are run inside one shared \texttt{GroupKFold} with the same
group-disjoint inner selection split, so they differ in the decoder family and not
in how their hyperparameters were chosen. Panel, anchors, splits, baseline,
representations, encoder training budget and seeds are unchanged throughout.

\paragraph{Determinism check.}
The ridge column is recomputed in the same process rather than copied. It
reproduces every one of the forty reported record-span cells exactly (maximum
absolute deviation $0$), which is what licenses reading the difference between
columns as an effect of the decoder.

\paragraph{Record span.}
Table~\ref{tab:mlp-recordspan} reports the sweep under both decoders. The decline
in changing-state information with record span persists in both cohorts:
$-72\%$ and $-79\%$ for person-contrastive learning, against $-80\%$ and $-82\%$
under ridge. The objective-specific ordering is unchanged, with predictive
representations rising ($+20\%$ and $+18\%$) and masked reconstruction flat
($-2\%$ and $-1\%$). The nonlinear decoder is not a uniform rescaling: it raises
contrastive $\Delta I$ by roughly half at the shortest record and slightly lowers
predictive and masked values, so the objectives are read differently rather than
all shifting together.

\begin{table}[!ht]
\centering
\caption{Record-span sweep under ridge and MLP decoders. Values are $\Delta I$ in
nats per output dimension; the final column is the change from a 30-day record to
the full record.}
\label{tab:mlp-recordspan}
\small
\begin{tabular}{lllccccc r}
\toprule
Cohort & Objective & Decoder & 30d & 60d & 120d & 240d & Full & 30d$\to$full \\
\midrule
\multirow{8}{*}{Single}
 & \multirow{2}{*}{Person-contrastive} & Ridge & .0971 & .0697 & .0565 & .0460 & .0198 & $-80\%$ \\
 &                                     & MLP   & .1433 & .1352 & .0809 & .0619 & .0398 & $-72\%$ \\
 & \multirow{2}{*}{\quad sep.-matched} & Ridge & .0896 & .1014 & .0827 & .1129 & .1004 & $+12\%$ \\
 &                                     & MLP   & .1390 & .1408 & .1181 & .1308 & .1271 & $-9\%$ \\
 & \multirow{2}{*}{Predictive}         & Ridge & .1983 & .2129 & .2136 & .2237 & .2299 & $+16\%$ \\
 &                                     & MLP   & .1752 & .1747 & .1821 & .1984 & .2094 & $+20\%$ \\
 & \multirow{2}{*}{Masked}             & Ridge & .2040 & .2042 & .1980 & .2030 & .2049 & $+0\%$ \\
 &                                     & MLP   & .1819 & .1777 & .1791 & .1790 & .1787 & $-2\%$ \\
\midrule
\multirow{8}{*}{Households}
 & \multirow{2}{*}{Person-contrastive} & Ridge & .0958 & .0744 & .0422 & .0222 & .0171 & $-82\%$ \\
 &                                     & MLP   & .1217 & .0923 & .0622 & .0506 & .0258 & $-79\%$ \\
 & \multirow{2}{*}{\quad sep.-matched} & Ridge & .0876 & .0862 & .0868 & .0755 & .0634 & $-28\%$ \\
 &                                     & MLP   & .1235 & .1122 & .1044 & .0938 & .1154 & $-7\%$ \\
 & \multirow{2}{*}{Predictive}         & Ridge & .1770 & .1746 & .1779 & .1900 & .1906 & $+8\%$ \\
 &                                     & MLP   & .1439 & .1438 & .1576 & .1661 & .1701 & $+18\%$ \\
 & \multirow{2}{*}{Masked}             & Ridge & .1806 & .1845 & .1885 & .1859 & .1864 & $+3\%$ \\
 &                                     & MLP   & .1484 & .1472 & .1420 & .1471 & .1463 & $-1\%$ \\
\bottomrule
\end{tabular}
\end{table}

\paragraph{Supervision span.}
The same substitution on the supervision-span axis gives the same answer. On the
next-five-day behavioral state, ridge and MLP respectively read $0.0611$ and
$0.0580$ at $\tau{=}2$ in the single-resident cohort, falling to $-0.0003$ and
$0.0062$ under unrestricted pairing; the household cohort falls from $0.0667$ and
$0.0443$ to $0.0128$ and $0.0110$. Both decoders also place $\tau{=}2$ at or below
the untrained floor of $0.0935$ and $0.0879$ (single) and $0.0655$ and $0.0760$
(households), so nonlinear capacity does not recover an advantage for the trained
encoder over random initialization on this target.

\paragraph{Decoder capacity.}
The grid above selected its largest configuration in every cell, which leaves open
whether the residual decline reflects a capacity ceiling. Extending the grid to
hidden width $512$ at the two extreme record spans changes the magnitude but not
the direction: person-contrastive reads $-69\%$ against $-79\%$, and predictive
$+12\%$ against $+18\%$. The wide grid does select the $512$-unit configuration,
so the additional capacity is used rather than merely offered.

\paragraph{The identity side.}
The asymmetry that motivates this appendix does not survive inspection from the
other direction either. Probing identity with an MLP of the same family and
capacity, rather than the linear probe, \emph{lowers} identity accuracy at every
anchor-relative distance in both cohorts: at zero distance on the raw history
block it reads $0.527$ against $0.629$ (single-resident) and $0.694$ against
$0.736$ (households), and the gap persists out to 180 days.
Nonlinear capacity therefore helps the state readout and hurts the identity
readout, and the divergence between them is present under either choice.

\paragraph{Limitation.}
Besides the next-five-day state, we also probed three non-nested temporal
contrasts, built causally from windows strictly after the anchor as differences of
box averages at increasing timescales, so that each emphasizes one timescale
rather than containing the faster ones: fast, medium and slow. On the fast and
medium contrasts the MLP decoder returns values at or indistinguishable from zero
for \emph{every} representation, including the untrained control, meaning it
collapses to the baseline on those targets. Those two targets are uninformative
under nonlinear decoding and we do not read them as evidence in either direction,
and no claim in this paper rests on them. The slow contrast and the
next-five-day state behave normally.

\subsection{Sequence-aware encoder}
\label{app:gru}
\label{app:gru-distance}
\label{app:gru-length}

The primary history representation is an MLP over a flattened block of windows,
which is permutation-sensitive only through position in the input vector. As
Section~\ref{non-exchangeable} claims that temporal location and temporal order
both matter, we repeat the three central analyses with a recurrent encoder whose
inductive bias is explicitly sequential. All arms use a GRU with a last-state
readout at matched parameter count ($40{,}200$ parameters), the same anchors,
splits, seeds, and downstream ridge decoder as the primary analyses.

\begin{table}[!ht]
\centering
\caption{History-source results with a GRU last-state encoder. Values are $\Delta I$
in nats per output dimension, mean $\pm$ standard deviation over five seeds.
Contrasts show the paired difference with a 95\% bootstrap interval over units.}
\label{tab:gru-source}
\small
\begin{tabular}{lcc}
\toprule
& Single-resident (52 units) & Households (42 units) \\
\midrule
Recent   & $0.3046 \pm 0.0037$ & $0.2443 \pm 0.0034$ \\
Shifted  & $0.0313 \pm 0.0214$ & $0.0145 \pm 0.0061$ \\
Random   & $0.0362 \pm 0.0140$ & $0.0418 \pm 0.0028$ \\
Distant  & $0.0085 \pm 0.0140$ & $-0.0003 \pm 0.0033$ \\
\midrule
Recent $-$ shifted & $+0.2732$ $[+0.2319,+0.3195]$, 5/5 & $+0.2298$ $[+0.1880,+0.2730]$, 5/5 \\
Recent $-$ random  & $+0.2684$ $[+0.2263,+0.3159]$, 5/5 & $+0.2026$ $[+0.1638,+0.2416]$, 5/5 \\
Recent $-$ distant & $+0.2961$ $[+0.2510,+0.3469]$, 5/5 & $+0.2447$ $[+0.1956,+0.2959]$, 5/5 \\
\bottomrule
\end{tabular}
\end{table}

The recent--non-recent separation is preserved and slightly enlarged relative to
the MLP (Table~\ref{tab:gru-source}). In the single-resident cohort the three
contrasts grow from $+0.2371$, $+0.2444$ and $+0.2767$ under the MLP to $+0.2732$,
$+0.2684$ and $+0.2961$; all six comparisons hold across 5/5 seeds in both cohorts
with bootstrap intervals excluding zero. Permuted-target controls remain at most
$0.03$ nats per dimension in magnitude. As with the MLP, the ordering among the
three non-recent arms does not replicate across cohorts, so we again claim only
the separation between recent and non-recent history.

\begin{table}[!ht]
\centering
\caption{Temporal-distance sweep with a GRU last-state encoder. A fixed 20-day
block is moved progressively further from the anchor with the target held at
$t+5$ days. Brackets are 95\% bootstrap intervals over units.}
\label{tab:gru-distance}
\small
\begin{tabular}{lcc}
\toprule
Gap from present & Single-resident & Households \\
\midrule
0 days  & $0.3071$ $[0.2562,0.3625]$ & $0.2429$ $[0.1936,0.2935]$ \\
5 days  & $0.2473$ $[0.2025,0.2949]$ & $0.1946$ $[0.1511,0.2391]$ \\
10 days & $0.1970$ $[0.1552,0.2427]$ & $0.1549$ $[0.1146,0.2008]$ \\
15 days & $0.1764$ $[0.1356,0.2198]$ & $0.1477$ $[0.1053,0.1940]$ \\
20 days & $0.1595$ $[0.1237,0.1983]$ & $0.1206$ $[0.0764,0.1678]$ \\
30 days & $0.1414$ $[0.1049,0.1790]$ & $0.1044$ $[0.0703,0.1419]$ \\
40 days & $0.1127$ $[0.0794,0.1456]$ & $0.0838$ $[0.0543,0.1163]$ \\
60 days & $0.1078$ $[0.0750,0.1400]$ & $0.0623$ $[0.0384,0.0872]$ \\
\bottomrule
\end{tabular}
\end{table}

Predictive information decays monotonically with distance under the GRU in both
cohorts (Table~\ref{tab:gru-distance}). Over 60 days it falls by 65\% and 74\%,
against 59\% and 72\% for the MLP, so the decay is if anything slightly steeper
for the sequence-aware encoder. The distance effect is therefore not an artifact
of the primary representation.

Section~\ref{non-exchangeable} also reports that predictive information is not confined to the
latest window, and that claim turns out to depend on the readout rather than the encoder family.
Extending recent history from one window to eight helps the GRU in both cohorts, with every
interval excluding zero, while a plain mean and a mean-pooled Transformer \emph{lose} information
by $N=8$ in the single-resident cohort. Training the same GRU and changing only how its hidden
sequence is reduced isolates the cause: the last-state readout improves over $N=1$ in $5/5$ seeds
in each cohort, the mean-pooled readout degrades in $0/5$, and the two diverge in sign at $N=8$ in
both cohorts. Averaging over a history window discards the recency structure that makes the window
informative, which is why the dilution scales with the number of windows averaged.

\subsection{Temporal order at daily resolution}
\label{app:order-daily}

The temporal-order comparison of Section~\ref{non-exchangeable} uses five-day
windows, so the order it manipulates is the order of five-day aggregates. A
reader may reasonably ask whether the effect is a property of the behavior or of
that aggregation. We repeat the comparison on the daily panel, changing the
window length and nothing else: the same builder, the same three arms, the same
parameter-matched encoders, the same per-anchor fixed permutation, the same
estimator, and the same five seeds. The train/test buffer is scaled from eight
to forty windows so that it remains forty calendar days.

Two matchings are reported, because at daily resolution they come apart and
reporting only one would be a choice rather than a result. At $K{=}4$ the
comparison is matched on \emph{window count} to the main-text $K{=}4$ row, giving
four days of context through an identical architecture. At $K{=}20$ it is matched
on \emph{calendar span}, giving the same twenty days of context the main-text row
receives. Absolute $\Delta I$ is not comparable across resolutions, since the
number and length of windows both change; the ordered$-$shuffled contrast is.

Table~\ref{tab:order-daily} reports the result. The advantage of correct
temporal order is larger at daily resolution than at five days under either
matching, by a factor of $1.6$ to $3$, and holds on 5/5 seeds in both cohorts
with bootstrap intervals excluding zero. Shuffled and set remain
indistinguishable throughout ($-0.0059$ to $+0.0036$, all intervals spanning
zero), so the gain is attributable to order rather than to encoder class, as in
the five-day analysis. The main-text five-day effect therefore understates the
value of temporal order rather than depending on the aggregation that produced
it.

\begin{table}[!ht]
\centering
\caption{Temporal order at daily resolution. Values are $\Delta I$ in nats per
output dimension; brackets give participant- or household-level bootstrap 95\%
confidence intervals for the ordered$-$shuffled difference. $K{=}4$ is matched on
window count to the main-text five-day row and $K{=}20$ on calendar span. The
final column repeats the main-text five-day value for reference.}
\label{tab:order-daily}
\small
\resizebox{\linewidth}{!}{%
\begin{tabular}{llccccc}
\toprule
Cohort & Context & Ordered & Shuffled & Set
& Ordered $-$ Shuffled & Main-text 5d \\
\midrule
Single-resident
& $K{=}4$ (4d)
& 0.3016 & 0.2381 & 0.2439
& \textbf{+0.0635} [0.0545, 0.0721] & +0.0408 \\

Single-resident
& $K{=}20$ (20d)
& 0.2602 & 0.1605 & 0.1569
& \textbf{+0.0997} [0.0824, 0.1183] & +0.0408 \\

Households
& $K{=}4$ (4d)
& 0.3167 & 0.2481 & 0.2539
& \textbf{+0.0686} [0.0486, 0.0882] & +0.0228 \\

Households
& $K{=}20$ (20d)
& 0.2598 & 0.1364 & 0.1423
& \textbf{+0.1234} [0.0975, 0.1519] & +0.0228 \\
\bottomrule
\end{tabular}}
\end{table}

Anchor counts are 28{,}497 and 26{,}130 for the single-resident cohort at
$K{=}4$ and $K{=}20$ over 56 participants, and 23{,}262 and 21{,}130 for the
household cohort over 42 households.

\subsection{Representation dimension and capacity}
\label{app:capacity}

Comparisons between representation-learning objectives can be confounded if
one model has substantially greater representational or parameter capacity.
We therefore match representation dimensionality and search encoder widths to
obtain comparable realized parameter counts across experimental arms.

Table~\ref{tab:capacity-matching} reports the resulting architecture and
parameter count for every learned-representation comparison.

\begin{table*}[!ht]
\centering
\caption{Realized model capacity by experimental arm. For the sequence encoders of
Appendix~\ref{app:gru-length}, width is the recurrent or model dimension and is matched to
the same $40$k budget at each context length $N$; the Transformer is a single encoder layer with
$4$ heads, learned positional embeddings, feed-forward width $2d_{\text{model}}$, no dropout, and
mean pooling over positions.}
\label{tab:capacity-matching}
\small
\resizebox{\linewidth}{!}{%
\begin{tabular}{llrrr}
\toprule
Experiment & Arm & Representation dim. & Encoder width & Parameters \\
\midrule
History source & All four arms & 32 & 95 & 33{,}791 \\
Temporal order & Ordered / shuffled & 32 & 95 & 33{,}791 \\
Temporal order & Set (deep sets) & 32 & 88 & 33{,}908 \\
Temporal order, $K{=}8$ & Ordered / shuffled & 32 & 71 & 34{,}111 \\
\midrule
Sequence encoders, $N{=}1/2/4/8$ & MLP & 48 & 151 / 134 / 108 / 74 & 39{,}999 / 39{,}920 / 40{,}200 / 39{,}992 \\
Sequence encoders, $N{=}1/2/4/8$ & GRU (1 layer) & 48 & 84 & 40{,}200 \\
Sequence encoders, $N{=}1/2/4/8$ & Transformer & 48 & 64 / 60 / 60 / 60 & 42{,}144 / 37{,}800 / 37{,}920 / 38{,}160 \\
\bottomrule
\end{tabular}}
\end{table*}

Capacity matching holds the representation dimension fixed across arms within an
experiment. A separate question is whether the state/identity dissociation itself
depends on that dimension: a 32-unit bottleneck might simply be too small to carry
both, forcing a trade-off that a wider representation would not face.

We repeat the record-span sweep at representation dimensions $16$, $32$, $64$ and
$128$. The encoder width is re-solved at each dimension so the realized parameter
count stays within $1\%$ of $40{,}000$ ($40{,}031$, $39{,}907$, $40{,}159$ and
$39{,}800$), which means total capacity is held fixed and only the bottleneck
varies. Anchors, splits, seeds and the downstream decoder are unchanged.

Table~\ref{tab:dim-sweep} reports the uniform-positive arm. The dissociation is
present at every dimension in both cohorts: changing-state information falls from
the 30-day record to the full record while identity accuracy rises. Absolute
$\Delta I$ increases with dimension, as expected from a wider bottleneck, but the
decline and the identity rise do not weaken, so neither is an artifact of a
constrained representation.

\begin{table}[!ht]
\centering
\caption{Record-span sweep at four representation dimensions, uniform positives,
parameter count matched to within $1\%$. $\Delta I$ is in nats per output
dimension; identity is multinomial decoding accuracy.}
\label{tab:dim-sweep}
\small
\begin{tabular}{llcccc}
\toprule
& & \multicolumn{2}{c}{30-day record} & \multicolumn{2}{c}{Full record} \\
\cmidrule(lr){3-4}\cmidrule(lr){5-6}
Cohort & Dim. & $\Delta I$ & Identity & $\Delta I$ & Identity \\
\midrule
\multirow{4}{*}{Single}
 & $16$  & $0.0639$ & $0.608$ & $0.0214$ & $0.842$ \\
 & $32$  & $0.0664$ & $0.619$ & $0.0243$ & $0.844$ \\
 & $64$  & $0.0731$ & $0.633$ & $0.0300$ & $0.841$ \\
 & $128$ & $0.0966$ & $0.613$ & $0.0376$ & $0.844$ \\
\midrule
\multirow{4}{*}{Households}
 & $16$  & $0.0304$ & $0.757$ & $0.0112$ & $0.861$ \\
 & $32$  & $0.0396$ & $0.759$ & $0.0138$ & $0.863$ \\
 & $64$  & $0.0543$ & $0.763$ & $0.0188$ & $0.861$ \\
 & $128$ & $0.0567$ & $0.760$ & $0.0308$ & $0.860$ \\
\bottomrule
\end{tabular}
\end{table}

\subsection{A temporally aware baseline}
\label{app:tnc}

Temporal Neighborhood Coding \citep{tonekaboni2021tnc} is the established method that restricts
temporal positives explicitly: positives are drawn from a neighborhood whose width is estimated
per signal by an augmented Dickey-Fuller stationarity test, and non-neighbors enter weighted by
$w$ because they may themselves be positives. We run it on the same build, anchors, budget and
evaluation, and add an ablation that keeps TNC's positive-selection rule under our own InfoNCE
loss, so the method's two contributions can be separated.

\begin{table}[!ht]
\centering
\caption{Raw-score TNC comparison on the \emph{five-day-window} build,
shown for completeness. Because representation norms differ across methods,
these values are not used for the final cross-method interpretation; the
scale-matched $L_2$-normalized analysis follows below. The last column is a
paired unit bootstrap against arm 1. Table~\ref{tab:tnc-daily} repeats the
raw-score comparison on the daily-window build used in
Figure~\ref{fig:4.2}.}
\label{tab:tnc}
\small
\resizebox{\linewidth}{!}{%
\begin{tabular}{lccccc}
\toprule
& \multicolumn{2}{c}{Single-resident} & \multicolumn{2}{c}{Households} & \\
\cmidrule(lr){2-3}\cmidrule(lr){4-5}
Arm & $\Delta I$ & identity & $\Delta I$ & identity & vs.\ unrestricted \\
\midrule
1 Unrestricted person-contrastive & $+0.0243$ & $0.844$ & $+0.0138$ & $0.863$ & --- \\
2 Hand-matched 30-day cap         & $+0.0830$ & $0.625$ & $+0.0458$ & $0.757$ & $+.0587$ / $+.0320$ \\
3 TNC                             & $+0.1752$ & $0.257$ & $+0.1612$ & $0.378$ & $+.1510$ / $+.1474$ \\
4 InfoNCE $+$ ADF neighborhood    & $+0.1109$ & $0.106$ & $+0.1005$ & $0.133$ & $+.0867$ / $+.0867$ \\
\midrule
\quad PCA, 32-d (reference)       & $+0.2674$ & $0.549$ & $+0.2233$ & $0.705$ & \\
\quad Untrained, 32-d (reference) & $+0.0995$ & $0.342$ & $+0.1087$ & $0.445$ & \\
\bottomrule
\end{tabular}}
\end{table}

At raw scale, every contrast in the last column excludes zero at five seeds
of five. Because these methods produce representations with substantially
different norms, however, we treat these values as descriptive and base the
cross-method conclusions on the scale-matched analysis below.

\paragraph{Which conclusions are build-dependent.} Figure~\ref{fig:4.2} uses the daily
build throughout, so we repeat the comparison there (Table~\ref{tab:tnc-daily}). TNC itself is
almost unchanged across geometries ($0.1752 \to 0.1716$ and $0.1612 \to 0.1711$); what moves are
the endpoints against which it is measured, and two secondary conclusions move with them.

\begin{table}[!ht]
\centering
\caption{Raw-score comparison on the daily-window build, matching
Figure~\ref{fig:4.2}. Because representation norms differ across
methods, final cross-method conclusions use the $L_2$-normalized analysis
below. The hand-matched arm and identity probes were not rerun under this
geometry, so those cells are unavailable.}
\label{tab:tnc-daily}
\small
\begin{tabular}{lcc}
\toprule
Arm & Single-resident & Households \\
\midrule
Unrestricted person-contrastive & $+0.0492$ & $+0.0223$ \\
TNC                             & $+0.1716$ & $+0.1711$ \\
InfoNCE $+$ ADF neighborhood    & $+0.1366$ & $+0.1825$ \\
\midrule
\quad Raw block, 96-d (reference)   & $+0.2738$ & $+0.3177$ \\
\quad PCA, 32-d (reference)         & $+0.2187$ & $+0.2648$ \\
\quad Random projection, 32-d       & $+0.1536$ & $+0.2170$ \\
\quad Untrained, 32-d (reference)   & $+0.1028$ & $+0.1493$ \\
\bottomrule
\end{tabular}
\end{table}

First, the recovery is large under both geometries. The raw-scored fractions of the
unrestricted-to-PCA gap in the run above are $62\%$ and $70\%$ on the five-day build and $72\%$
and $61\%$ on the daily one. We quote none of these as the recovery: the paragraphs below show
that the fraction moves both with the window geometry and with the draw of neighborhood
positives, and that the scoring itself needs correcting first.

\paragraph{A scale artifact, and what it costs.} InfoNCE normalizes its embeddings inside the
loss, so it constrains direction only and leaves $\lVert z \rVert$ free; TNC's bilinear score does
not. The two arms therefore arrive at the decoder on different scales --- measured norms are
$0.53$--$2.58$ for the InfoNCE-trained arms against $4.2$--$4.4$ for TNC (and $8.6$--$9.3$ for
PCA-$32$) --- and a ridge decoder can exploit magnitude. This was found in the external
replication (Appendix~\ref{app:globem}) and applies here. We therefore score all four
build\,$\times$\,cohort cells after $L_2$-normalizing every representation. Two of the three
comparisons this experiment supports at raw scale do not survive that correction.

\emph{Recovery.} TNC exceeds unrestricted person-contrastive training
in all four build\,$\times$\,cohort cells under both readouts, with every interval excluding zero
on $5/5$ seeds. Normalized, it recovers $63\%$, $47\%$, $50\%$ and $64\%$ of the
unrestricted-to-PCA gap.

We do not quote a precise recovery percentage. The re-scored run draws its own ADF neighborhood
positives and does not reproduce the original TNC arm on the daily build ($52\%$ against $72\%$
raw in single-resident, $68\%$ against $61\%$ in households), while the unrestricted and PCA
endpoints reproduce to within $0.007$; the movement is in the TNC arm alone. Across both runs,
geometries and cohorts the recovery spans roughly $47\%$ to $72\%$, so we report that TNC recovers
between about a half and two thirds of what unrestricted supervision destroys and treat the exact
fraction as unstable. PCA-$32$ still exceeds TNC in all four cells under both readouts, so
that limit is unchanged.

\emph{Comparison against the untrained reference.} Raw, TNC exceeds a
same-architecture untrained encoder in $4/4$ cells. Normalized, it does so in $1/4$: $+0.0249$
$[+0.0080,+0.0427]$ on
five-day single-resident, against $-0.0065$ $[-0.0394,+0.0235]$, $-0.0087$ $[-0.0339,+0.0165]$
and $+0.0156$ $[-0.0085,+0.0443]$ elsewhere, all intervals containing zero.

\emph{Decomposition of the two components.} At raw scale, TNC's positive-selection rule ---
restricting positives to an estimated temporal neighborhood --- appears to carry most of the
recovery. Under scale matching it carries none:
InfoNCE-plus-ADF minus untrained is $-0.0058$ $[-0.0291,+0.0158]$, $-0.0035$
$[-0.0270,+0.0211]$, $+0.0169$ $[-0.0104,+0.0449]$ and $+0.0082$ $[-0.0246,+0.0372]$, an
interval containing zero in all
four cells. The neighborhood rule alone buys nothing over an untrained encoder of the same shape.
TNC minus the ablation is significant in only $1/4$ cells and changes sign across them, so the
discriminator's contribution remains unresolved.

Taken together: TNC rescues \emph{contrastive learning}, recovering roughly half
to two thirds of what unrestricted person-contrastive training destroys, robustly across four
cells and both readouts. It does not extract state information that an untrained encoder of the
same shape does not already carry, and neither of its two components does so alone; the latter is
visible only once representation scale is matched.

\paragraph{Limits that hold under both builds.} TNC does not close the raw-to-learned gap.
On $L_2$-normalized representations with hierarchical intervals, PCA-$32$d exceeds it by
$+0.0727$ $[+0.0525,+0.0947]$ and $+0.0962$ $[+0.0719,+0.1254]$ on five-day, and by $+0.0928$
$[+0.0690,+0.1180]$ and $+0.0700$ $[+0.0460,+0.0932]$ on daily --- all four intervals excluding
zero,
$5/5$ seeds. And TNC's
ADF criterion selects the shortest admissible neighborhood
for every unit, which is the floor of our candidate grid imposed by the non-overlap
constraint ($10$ days at five-day windows, $2$ days at daily). The supportable reading is that
stationarity fails beyond that floor, not that the floor is an estimated optimum. That it lands
far below the $245$-day mean separation of unrestricted sampling, and near the hand-chosen
$30$-day cap, is independent corroboration of the diagnosis rather than a tuned result.

\subsection{Capacity-matched representation-free baselines}
\label{app:capacity-matched}

Appendix~\ref{app:repfree} feeds the raw history block directly to the decoder and compares it
against $32$-d learned representations, a comparison that is not capacity-matched: the block is
$192$-d there, and $96$-d on the fixed-anchor build used here. We remove the confound by reducing
the block to $32$ dimensions two ways, by PCA fit on training anchors only and by a fixed Gaussian
random projection, and scoring both through the identical decoder.

\begin{table}[!ht]
\centering
\caption{Capacity-matched baselines, ridge decoder. PCA is fit on training anchors only and the
training mean and components are applied unchanged to test.}
\label{tab:capacity-matched}
\small
\begin{tabular}{lcccc}
\toprule
& \multicolumn{2}{c}{Single-resident} & \multicolumn{2}{c}{Households} \\
\cmidrule(lr){2-3}\cmidrule(lr){4-5}
Representation & $\Delta I$ & identity & $\Delta I$ & identity \\
\midrule
Raw block, 96-d                  & $+0.3013$ & $0.614$ & $+0.2308$ & $0.738$ \\
PCA, 32-d (training-only)        & $+0.2674$ & $0.549$ & $+0.2233$ & $0.705$ \\
Random projection, 32-d          & $+0.2034$ & $0.462$ & $+0.1947$ & $0.616$ \\
Untrained encoder, 32-d          & $+0.0995$ & $0.342$ & $+0.1087$ & $0.445$ \\
Learned, short span              & $+0.0664$ & $0.619$ & $+0.0396$ & $0.759$ \\
Learned, unrestricted            & $+0.0243$ & $0.844$ & $+0.0138$ & $0.863$ \\
\bottomrule
\end{tabular}
\end{table}

PCA to $32$ dimensions retains $81.7\%$ and $84.1\%$ of block variance and $89\%$ and $97\%$ of
the raw block's $\Delta I$. Dimensionality therefore explains almost none of the gap, and the
caveat attached to Appendix~\ref{app:repfree} does not survive: under a paired bootstrap,
PCA-$32$d exceeds the strong learned representation by $+0.2095$ $[+0.1763,+0.2516]$ and
$+0.1751$ $[+0.1242,+0.2242]$.

Two secondary observations. The untrained floor is architecture-specific: an untrained
three-layer MLP scores $0.0995$ where a plain linear random projection at the same $32$
dimensions scores $0.2034$, so the ReLU stack itself discards information at initialization and
the appropriate learning-free reference is PCA. And the gap is not an artifact of a linear
readout. Replacing ridge with an MLP decoder, hyperparameters selected on the same validation
split with independent selection per arm, moves every arm in the expected direction but leaves
the contrast at $+0.1631$ $[+0.1356,+0.1948]$ and $+0.1646$ $[+0.1370,+0.1956]$. The MLP is not
uniformly stronger, since it lowers the raw block from $0.3013$ to $0.2332$, so the true gap
likely lies between the two decoders rather than at the nonlinear end.

\subsection{Contrastive implementation and pretext difficulty}
\label{app:contrastive}

Encoder architecture, optimization, positive and negative sampling, and the held-out
contrastive metrics follow Appendix~\ref{app:implementation}; this subsection reports only
the difficulty manipulations.

A natural alternative explanation for the span effect is that the pretext task saturates as
records grow: if person discrimination becomes trivial, the objective would stop extracting
useful structure and $\Delta I$ would fall for reasons unrelated to pair separation. We test
this by manipulating task difficulty directly.

Rather than changing the temperature, which alters the optimization geometry itself, we
change the negatives, in two independent ways.

\paragraph{Harder negatives.} In the \emph{hard} condition each anchor's $M_{-}$ negatives
are the nearest other-unit history blocks in raw feature space under $L_2$, mined once
before training and then fixed; same-unit candidates are masked out so the negatives remain
valid. This raises difficulty without perturbing the optimizer. The manipulation works as
intended: mean anchor-to-negative distance falls from $13.85$ to $8.25$ in the
single-resident cohort and from $13.79$ to $8.42$ in the household cohort.

\paragraph{A larger, dynamically resampled pool.} A related objection is that eight
negatives sampled once at construction is simply too impoverished a pool, so that the
objective degenerates for reasons unrelated to span. In the \emph{in-batch} condition every
other-unit anchor in the $128$-anchor minibatch acts as a negative, remasked at every step,
so the pool is both far larger and resampled dynamically as batches are reshuffled each
epoch. Same-unit members are masked to $-\infty$, which also removes the diagonal, so no
negative is ever a same-person block. The realized pool is $63.3$ and $51.8$ negatives per
anchor rather than the nominal $127$, because a batch of $128$ contains several anchors from
the same unit; the shortfall is consistent with $54$ and $42$ units and confirms that the
masking is active.

\begin{table}[!ht]
\centering
\caption{Contrastive difficulty crossed with longitudinal span. Pretext metrics are computed
on held-out validation anchors. Brackets are 95\% bootstrap intervals over units; ``seeds''
counts how many of five show a decline from 30 days to the full record.}
\label{tab:pretext-difficulty}
\small
\resizebox{\linewidth}{!}{%
\begin{tabular}{llccccc}
\toprule
Cohort & Negatives & Span & InfoNCE loss & Top-1 & $\Delta I$ & Change \\
\midrule
Single-resident & Easy (uniform) & 30d   & $2.2061$ & $0.595$ & $+0.0664$ $[+0.0463,+0.0893]$ & \\
Single-resident & Easy (uniform) & Full  & $2.7368$ & $0.353$ & $+0.0243$ $[+0.0162,+0.0334]$ & $-63\%$, 5/5 \\
Single-resident & Hard (mined)   & 30d   & $2.8729$ & $0.262$ & $+0.0392$ $[+0.0228,+0.0578]$ & \\
Single-resident & Hard (mined)   & Full  & $3.5039$ & $0.197$ & $+0.0234$ $[+0.0147,+0.0337]$ & $-40\%$, 5/5 \\
\midrule
Households & Easy (uniform) & 30d  & $2.1341$ & $0.625$ & $+0.0396$ $[+0.0232,+0.0608]$ & \\
Households & Easy (uniform) & Full & $2.4543$ & $0.443$ & $+0.0138$ $[+0.0051,+0.0256]$ & $-65\%$, 5/5 \\
Households & Hard (mined)   & 30d  & $2.6289$ & $0.370$ & $+0.0163$ $[+0.0058,+0.0297]$ & \\
Households & Hard (mined)   & Full & $3.0506$ & $0.308$ & $+0.0116$ $[+0.0031,+0.0230]$ & $-29\%$, 4/5 \\
\bottomrule
\end{tabular}}
\end{table}

\begin{table}[!ht]
\centering
\caption{Larger, dynamically resampled negative pool. The fixed-8 rows repeat the standard
condition as a paired reference within the same run. Pretext metrics are computed on held-out
validation anchors in batches of $128$ and averaged across batches, which is the batching the
in-batch pool requires; Table~\ref{tab:pretext-difficulty} scores the whole validation set in a
single pass. The fixed-8 checkpoints are the same in both tables, which is why $\Delta I$
matches exactly while the pretext columns differ slightly. Brackets are 95\% bootstrap
intervals over units.}
\label{tab:inbatch-negatives}
\small
\resizebox{\linewidth}{!}{%
\begin{tabular}{llccccc}
\toprule
Cohort & Negatives & Span & Pool size & InfoNCE loss & Top-1 & $\Delta I$ [95\% CI] \\
\midrule
Single-resident & Fixed-8  & 30d  & $8.0$  & $2.1453$ & $0.622$ & $+0.0664$ $[+0.0463,+0.0893]$ \\
Single-resident & Fixed-8  & Full & $8.0$  & $2.6355$ & $0.407$ & $+0.0243$ $[+0.0162,+0.0334]$ \\
Single-resident & In-batch & 30d  & $63.3$ & $2.4125$ & $0.582$ & $+0.0334$ $[+0.0238,+0.0440]$ \\
Single-resident & In-batch & Full & $63.3$ & $3.1163$ & $0.430$ & $+0.0146$ $[+0.0100,+0.0199]$ \\
\midrule
Households & Fixed-8  & 30d  & $8.0$  & $2.0827$ & $0.653$ & $+0.0396$ $[+0.0232,+0.0608]$ \\
Households & Fixed-8  & Full & $8.0$  & $2.3975$ & $0.471$ & $+0.0138$ $[+0.0051,+0.0256]$ \\
Households & In-batch & 30d  & $51.8$ & $2.3105$ & $0.611$ & $+0.0197$ $[+0.0101,+0.0318]$ \\
Households & In-batch & Full & $51.8$ & $2.7693$ & $0.516$ & $+0.0103$ $[+0.0030,+0.0201]$ \\
\bottomrule
\end{tabular}}
\end{table}

Three results follow (Tables~\ref{tab:pretext-difficulty} and~\ref{tab:inbatch-negatives}).
First, the pretext task does not
saturate as records grow; it becomes \emph{harder}. Under the standard sampling condition,
extending the record raises held-out InfoNCE loss from $2.21$ to $2.74$ and from $2.13$ to
$2.45$, and lowers top-1 accuracy from $0.595$ to $0.353$ and from $0.625$ to $0.443$. This
is the opposite of what a saturation account predicts, and it is consistent with the
pair-separation mechanism: positives drawn from across a full record are harder to align
precisely because they span different behavioral states.

Second, making the task harder does not prevent the degradation. Under mined negatives
$\Delta I$ still declines from 30 days to the full record, by $40\%$ (5/5 seeds) and $29\%$
(4/5 seeds).

Third, enlarging the negative pool does not prevent it either. With in-batch negatives the
decline is $56\%$ and $48\%$, at 5/5 seeds in both cohorts, against $63\%$ and $65\%$ for the
standard pool measured in the same runs. Pool size attenuates the effect slightly but comes
nowhere near explaining it. The pretext metrics again move against saturation: under
in-batch negatives, extending the record raises held-out InfoNCE from $2.41$ to $3.12$ and
from $2.31$ to $2.77$.

In both manipulations the absolute levels are lower at \emph{both} spans, since harder
negatives and larger pools each depress $\Delta I$ overall; the levels are therefore not
comparable across conditions and the quantity to read is the within-condition change. Across
three independent negative-sampling regimes -- uniform, mined-hard, and in-batch -- the
30-day-to-full decline holds at 5/5 seeds in every cohort except mined-hard households
(4/5), and in every regime the pretext task becomes harder rather than easier as records
grow. We therefore find no support for the saturation explanation.

\subsection{Global representation similarity}
\label{app:cka}

We test whether the predictive differences identified in
Section~\ref{non-exchangeable} are reflected by standard measures of
representation similarity. We compute linear Centered Kernel Alignment
(CKA) between the recent-history encoder and each alternative
\citep{kornblith2019similarity}. As independently trained instances
of the same encoder are not perfectly identical, we normalize CKA by the
similarity between two recent-history encoders trained with different
random seeds.

\begin{table}[t]
\centering
\caption{Normalized linear CKA and predictive information retained relative to the
recent-history encoder. CKA is normalized by the similarity between two recent-history
encoders trained with different random seeds (raw ceilings $0.876$ and $0.874$). Both cohorts
use the Cholesky estimator and the deduplicated household panel.}
\label{tab:cka-predictive}
\small
\begin{tabular}{lcccc}
\toprule
& \multicolumn{2}{c}{Single-resident} & \multicolumn{2}{c}{Households} \\
\cmidrule(lr){2-3}\cmidrule(lr){4-5}
History & CKA / ceiling & Information & CKA / ceiling & Information \\
\midrule
Recent  & 1.000 & 100\% & 1.000 & 100\% \\
Shifted & 0.536 & 9\%   & 0.592 & 7\%   \\
Random  & 0.571 & 9\%   & 0.610 & 10\%  \\
Distant & 0.571 & 5\%   & 0.597 & 2\%   \\
\bottomrule
\end{tabular}
\end{table}
Despite large differences in future-relevant information, the representations remain
substantially similar under CKA. Every non-recent arm retains between $2\%$ and $10\%$ of the
recent encoder's predictive information while sitting at $54$--$61\%$ of the recent--recent
CKA ceiling; the distant-history encoder retains $5\%$ and $2\%$ of the information at $57\%$
and $60\%$ of the ceiling. CKA also assigns near-identical values to the three non-recent
representations despite their differing predictive information. The two cohorts agree.

\paragraph{Reconstruction control.}
A representation need not learn future-predictive structure to achieve
high similarity. An autoencoder trained only to reconstruct its input
reaches 0.845 and 0.916 of the recent--recent CKA ceiling in the
single- and household cohorts, respectively. Thus, substantial
global similarity can arise from structure shared by the inputs and
model class rather than from information relevant to future behavior.

\paragraph{Where does the shared structure lie?}
We next separate directions that are strongly shared between
representations from the remaining representation. Removing the four
most strongly shared canonical directions from a 5-day encoder leaves
its predictive information essentially unchanged (0.2837 versus
0.2831 nats/dim for the full representation), whereas the shared
component alone provides only 0.1285 nats/dim. Conversely, adding the
remaining, less-shared component to a 40-day encoder recovers
approximately 90\% of the predictive gap between the two encoders.

These analyses show that the directions contributing most strongly to
global representation similarity need not be those carrying the
temporal information relevant for prediction. CKA is therefore useful
for characterizing global representational overlap, but it does not
establish that two longitudinal representations preserve the same
future-relevant information.
\section{Additional Analyses and Scope}
\label{app:scope}

\subsection{Held-out channel evaluation}
\label{app:heldout}

Every result so far evaluates representations by predicting the same 48-dimensional
behavioral vector that the self-supervised objectives are built from. A natural concern is
that the span effect is specific to that target. We therefore repeat the 30-day versus
full-record comparison against a behavioral channel that no encoder ever sees.

We hold out the \emph{time in bed} channel and its three derived statistics: the log daily
total, the overnight fraction, and the hourly-distribution entropy. These three dimensions
are removed from the encoder inputs, from the self-supervised objectives' own prediction
targets, and from the person baseline $B_i$. Encoders therefore operate on 45 dimensions and
have no access to the target at any point in training. Evaluation uses the same ridge
decoder conditioned on $B_i$, the same anchors, splits, seeds and unit bootstrap as the
record-span experiment, with the training pool fixed at 269 and 208 anchors so that longer
records confer no additional gradient steps.

\paragraph{Scope of the claim.} We describe this as a held-out feature evaluation rather
than cross-channel prediction, and it is not external clinical validation. Time in bed is
another ambient stream from the same homes, sharing the
household, the sensor installation and the missingness process with the inputs. It is a
target the objectives never optimized for, which is what the comparison requires, but it is
not an outcome measured independently of the sensing system.

\begin{table}[!ht]
\centering
\caption{Held-out channel evaluation. $\Delta I$ is measured on the time-in-bed statistics,
which are excluded from all encoder inputs, from the self-supervised targets, and from
$B_i$. Brackets are 95\% bootstrap intervals over units; identity accuracy is reported on
the same frozen representations.}
\label{tab:heldout}
\small
\resizebox{\linewidth}{!}{%
\begin{tabular}{llcccc}
\toprule
Cohort & Objective & 30-day $\Delta I$ & Full-record $\Delta I$ & Change & Identity (30d $\to$ full) \\
\midrule
Single-resident & Person-contrastive         & $+0.2030$ $[+0.1494,+0.2718]$ & $-0.0115$ $[-0.0335,+0.0136]$ & $-106\%$ & $0.467 \to 0.738$ \\
Single-resident & \quad separation-matched   & $+0.2406$ $[+0.1772,+0.3225]$ & $+0.3208$ $[+0.2474,+0.4010]$ & $+33\%$  & $0.471 \to 0.560$ \\
Single-resident & Predictive                 & $+0.5652$ $[+0.4644,+0.6693]$ & $+0.6381$ $[+0.5081,+0.7900]$ & $+13\%$  & $0.500 \to 0.500$ \\
Single-resident & Masked reconstruction      & $+0.6829$ $[+0.5578,+0.8281]$ & $+0.6715$ $[+0.5422,+0.8227]$ & $-2\%$   & $0.433 \to 0.433$ \\
\midrule
Households & Person-contrastive         & $+0.2177$ $[+0.1736,+0.2677]$ & $-0.0018$ $[-0.0095,+0.0049]$ & $-101\%$ & $0.621 \to 0.827$ \\
Households & \quad separation-matched   & $+0.2039$ $[+0.1627,+0.2493]$ & $+0.1850$ $[+0.1343,+0.2416]$ & $-9\%$   & $0.626 \to 0.740$ \\
Households & Predictive                 & $+0.4769$ $[+0.3631,+0.6059]$ & $+0.4985$ $[+0.3792,+0.6418]$ & $+5\%$   & $0.643 \to 0.636$ \\
Households & Masked reconstruction      & $+0.5629$ $[+0.4448,+0.6982]$ & $+0.5465$ $[+0.4317,+0.6794]$ & $-3\%$   & $0.566 \to 0.527$ \\
\bottomrule
\end{tabular}}
\end{table}

The scale-dependent degradation reproduces on the held-out channel, and in a stronger form
than on the within-objective target (Table~\ref{tab:heldout}). Under person-level
contrastive learning, extending the record removes essentially all usable information about
time in bed: $\Delta I$ falls from $+0.2030$ to $-0.0115$ and from $+0.2177$ to $-0.0018$,
so a full-record representation predicts the held-out channel no better than the person
baseline alone. Both full-record intervals include zero.

Separation-matched positives prevent this. In the single-resident cohort the matched arm
\emph{improves} with span, from $+0.2406$ to $+0.3208$, and in the household cohort it
loses only $9\%$. Predictive and masked representations are stable or improving throughout
($+13\%$, $+5\%$, $-2\%$, $-3\%$). Identity moves in the opposite direction to state, as
elsewhere: $0.467 \to 0.738$ and $0.621 \to 0.827$ under uniform sampling, against
$0.471 \to 0.560$ and $0.626 \to 0.740$ when pair separation is held fixed, while the
predictive and masked representations barely move.

As the target was withheld from the inputs, from the self-supervised objectives and
from the baseline, this cannot be explained by the representation and the evaluation target
sharing an optimization criterion. Subject to the scope caveat above, it shows the effect is
a property of what the representation encodes rather than of the particular reconstruction
target used elsewhere in the paper.

\subsection{Cohort-size analysis}
\label{app:cohort-size}

The span experiment of Section~\ref{sec:longer-records} is run on 54 single-resident
participants and 42 households. As person-level contrastive learning draws
its negatives from other identities, both the difficulty of the pretext task and
the size of the negative pool depend on how many people are in the cohort. We
therefore repeat the experiment while varying only the number of training
identities.

We reuse the Section~\ref{sec:longer-records} experiment unchanged: same contrastive
objective, same encoder and parameter budget, same anchor construction, same
pair sampling, same training budget, and the same downstream evaluation. The
only intervention is which participants are in the panel.

Subsets are drawn from the units that survive the eligibility filter, and the
panel is then rebuilt from scratch for each subset. Feature standardization, the
person baselines $B_i$, the anchor list, the negative pool, and the identity
label space are all recomputed within that subset, so each replicate is a
genuinely smaller study rather than a post-hoc slice of a larger one. This
matters here: a post-hoc slice would retain the negative pool of the full cohort
and would therefore not test the quantity of interest.

We use three levels, $N \in \{15, 30, \text{full}\}$, where full is 54
participants and 42 households. At $N=15$ and $N=30$ we draw six independent
random subsets; at the full level there is only one possible subset, so that
point has no across-subset variability and its uncertainty reflects seed
variation alone. Each cell is trained with three seeds. We report the paired
within-subset change from a 30-day record to the full record, since absolute
$\Delta I$ varies substantially across small participant draws and only the
paired difference is comparable across levels.

\begin{table}[!ht]
\centering
\caption{Effect of cohort size on the span experiment. Entries are the paired
change from 30-day to full-record training, $\Delta I_{\mathrm{full}} -
\Delta I_{30\mathrm{d}}$ and the corresponding change in identity accuracy,
mean $\pm$ standard deviation across participant subsets. The full level admits
only one subset, so no across-subset deviation is available.
Identity is multiclass accuracy, so chance is $1/N$ ($0.067$, $0.033$ and $0.019$ or $0.024$ at
the full level). No cell is near the ceiling: 30-day accuracies run $0.744$--$0.614$ and
$0.801$--$0.755$. Chance-adjusting the change, $(\mathrm{acc}-1/N)/(1-1/N)$, leaves it
essentially unchanged ($+0.134$, $+0.176$, $+0.235$ and $+0.075$, $+0.084$, $+0.111$ for uniform
positives), so its growth with cohort size is not a chance-level or ceiling artifact.}
\label{tab:cohort-size}
\small
\resizebox{\linewidth}{!}{%
\begin{tabular}{llcccc}
\toprule
& & \multicolumn{2}{c}{Uniform positives} & \multicolumn{2}{c}{Separation-matched} \\
\cmidrule(lr){3-4}\cmidrule(lr){5-6}
Cohort & $N$ & $\Delta(\Delta I)$ & $\Delta$ identity & $\Delta(\Delta I)$ & $\Delta$ identity \\
\midrule
Single-resident & 15 & $-0.0452 \pm 0.0273$ & $+0.125 \pm 0.031$ & $-0.0051 \pm 0.0054$ & $-0.001 \pm 0.010$ \\
Single-resident & 30 & $-0.0404 \pm 0.0214$ & $+0.171 \pm 0.027$ & $+0.0001 \pm 0.0056$ & $+0.003 \pm 0.006$ \\
Single-resident & 54 (full) & $-0.0428$ & $+0.231$ & $+0.0136$ & $+0.004$ \\
\midrule
Households & 15 & $-0.0394 \pm 0.0160$ & $+0.070 \pm 0.020$ & $+0.0058 \pm 0.0122$ & $-0.003 \pm 0.005$ \\
Households & 30 & $-0.0185 \pm 0.0046$ & $+0.081 \pm 0.009$ & $-0.0020 \pm 0.0029$ & $+0.005 \pm 0.005$ \\
Households & 42 (full) & $-0.0268$ & $+0.108$ & $+0.0059$ & $-0.001$ \\
\bottomrule
\end{tabular}}
\end{table}

The two components of the effect behave differently
(Table~\ref{tab:cohort-size}).

The loss of state information is flat across cohort size and the identity gain is not. Under
uniform sampling the single-resident cohort loses $-0.045$, $-0.040$ and $-0.043$ nats per
dimension at $N=15$, $30$ and $54$, and the household cohort shows no trend ($-0.039$, $-0.019$,
$-0.027$); over the same levels identity accuracy rises by $+0.125$, $+0.171$ and $+0.231$ and by
$+0.070$, $+0.081$ and $+0.108$, monotone in $N$ in both cohorts. That is the direction expected
if a larger negative pool makes person discrimination harder and absorbs more capacity. With
separation-matched positives the change in $\Delta I$ lies within $\pm0.014$ of zero in all six
cells and identity within $\pm0.005$, so the mechanism holds with as few as $15$ training
identities. A small study should therefore expect the state-information collapse at close to full
magnitude and roughly half the identity increase; of the two, the state result is the more
portable. This experiment uses two spans rather than five and three seeds per cell, so that six
subsets at each of two cohort sizes remain affordable.

\subsection{Alternative temporal objectives}
\label{app:alt-objectives}
\label{app:learned-weight-diagnostic}
\label{app:bucketize-audit}
\label{app:horizon-weighting}
\label{app:stable-local}
\label{app:horizon-mixture}
\label{app:retrieval}

Several modifications motivated by temporal non-exchangeability did not produce a robust
improvement across cohorts, and we record them so that the empirical finding that temporal
location matters is not read as the stronger claim that any temporal modification to the
objective recovers that information. Weighting the predictive loss by prediction horizon changed
$\Delta I$ by at most $0.01$ nats per dimension and did not replicate in sign across cohorts.
Factorizing the representation into a stable component from long history and a local component
from the recent block localized identity into the stable half but did not improve prediction.
Conditioning the encoder on the prediction horizon closed most of a mixture deficit in the
single-resident cohort ($0.0518$ to $0.0189$) and nothing in households, where the deficit it
targets is absent ($0.0137$ against $0.0140$). Retrieval architectures that select or aggregate
history conditional on the prediction context, with matched window counts and parameter budgets,
did not beat the fixed recent-history summary. In each case the direction of the effect was
either inconsistent across cohorts or within the interval of the unmodified objective.

\subsection{Diagnostics and implementation disclosures}
\label{app:negative-results}

To understand the behavior of learned temporal weighting, we examine
contrastive loss as a function of temporal separation $\Delta t$. At
initialization, per-bin InfoNCE loss increases monotonically with temporal
distance, from approximately
\[
6.042
\quad\text{to}\quad
6.429.
\]
Thus, temporally distant pairs are initially more difficult to align under
the contrastive objective.

After training, however, the per-$\Delta t$ loss curve is nearly flat, with
only a 1.4\% spread across bins. The learned temporal weights are strongly
negatively correlated with the initial alignment difficulty,
\[
r=-0.995.
\]

These diagnostics indicate that the learned temporal weights track \emph{contrastive
alignment difficulty} rather than measured informativeness. The decay the objective learns is a
statement about which pairs are easy to align, not about which histories carry information about
future behavior. The two happen to point the same way here, which is why a learned monotone
weighting resembles a calibrated one without being derived from the same quantity. The learned
weighting should therefore not be read as recovering the temporal informativeness curve measured
by $\Delta I$.

\begin{table}[!ht]
\centering
\caption{InfoNCE loss by temporal-separation bin before and after training,
for the seven effective bins (Appendix~\ref{app:bucketize-audit}). Loss is
averaged over all pairs falling in each bin; learned weights are normalized to
the nearest bin.}
\label{tab:weight-diagnostic}
\small
\begin{tabular}{lccc}
\toprule
$\Delta t$ bin & Initial loss & Final loss & Learned weight \\
\midrule
2 & 6.0422 & 0.0703 & 0.884 \\
3 & 6.1339 & 0.0704 & 0.732 \\
4--5 & 6.2152 & 0.0711 & 0.578 \\
6--8 & 6.2789 & 0.0712 & 0.424 \\
9--13 & 6.3430 & 0.0708 & 0.272 \\
14--20 & 6.3631 & 0.0704 & 0.169 \\
21+ & 6.4287 & 0.0713 & 0.084 \\
\bottomrule
\end{tabular}
\end{table}

An implementation audit identified an off-by-one error in the temporal
bucketization used by the weighting model. Although the configuration
specified eight temporal bins, the bucketization rule made one bin
unreachable, yielding seven effective bins.

We report all temporal-weighting diagnostics using the effective binning
actually seen by the model and do not interpret the unreachable bin as an
observed condition. This implementation detail does not alter the broader
finding that learned temporal weights track contrastive alignment difficulty
rather than measured predictive informativeness.
\section{Implementation Details}
\label{app:implementation}

Unless a section states otherwise, every learned representation in this paper uses the
configuration below, held fixed across spans, cohorts, representation widths and
interventions.

\subsection{Model architectures}
\label{app:impl-arch}

The encoder is a three-layer MLP over the flattened history block with ReLU activations,
projecting to a $32$-dimensional representation; hidden width is chosen so that the realized
parameter count is as close as possible to $40{,}000$, and is re-solved whenever the
representation width changes so that capacity is matched rather than confounded with
dimensionality. Representations are $L_2$-normalized before the similarity computation.

\subsection{Positive and negative sampling}
\label{app:impl-sampling}

Similarities are scaled by a temperature of $\tau = 0.1$. Each anchor receives $M_{+}$
positives and $M_{-} = 8$ negatives; $M_{+} = 4$ in the span experiments and $M_{+} = 6$ in
the pair-pool experiments of Appendix~\ref{app:pool-breadth}. Positives are
other windows from the \emph{same} unit, sampled uniformly from the eligible pool within the
available record span, except in the separation-matched condition where the pool is
restricted so that the pair-separation distribution matches its 30-day reference. Negatives
are anchors drawn uniformly from \emph{other} units, resampled once at panel construction
and then held fixed, so that all arms within a comparison see the same negatives. The loss
is InfoNCE with a multi-positive denominator: for each anchor the positive logits and all
negative logits enter a single \texttt{logsumexp}, and the mean is taken over positives.

\subsection{Optimization and hyperparameters}
\label{app:impl-optim}

Optimization uses Adam at a learning rate of $10^{-3}$ with minibatches of $128$ anchors,
for $60$ epochs over a training pool that is held to a fixed number of anchors at every
record span, so that longer records confer no additional gradient steps. Batches with fewer
than $16$ anchors are skipped. We note that Adam is invariant to a constant rescaling of the
loss, which is why the unnormalized and globally-normalized temporal weightings of
Appendix~\ref{app:learned-weight-diagnostic} agree while per-anchor normalization does not.

Held-out contrastive metrics are computed on the validation anchors, which are disjoint from
both the training pool and the test anchors used for $\Delta I$. The reported InfoNCE loss is
the same quantity minimized during training, evaluated under \texttt{no\_grad} after the
final epoch. Top-1 accuracy is the fraction of validation anchors whose \emph{lowest-scoring}
positive still outranks its \emph{highest-scoring} negative, i.e.\ a strict criterion under
which every positive must beat every negative.

\subsection{Compute and software}
\label{app:impl-compute}

All experiments run on CPU in double precision for the estimator and single precision for the
encoders, using PyTorch for the encoders and scikit-learn for the ridge decoders and identity
probes. A full record-span or supervision-span sweep at the base encoder is a few minutes per
cohort and seed; the $4\times$ encoder of Appendix~\ref{app:strong-contrastive} is roughly an
order of magnitude more.

\subsection{Strong-contrastive encoder}
\label{app:strong-contrastive}

If the person-contrastive learner were simply under-powered, the supervision-span effect would
be an artifact of a weak encoder rather than a property of the objective. We therefore make the
learner unambiguously competent and repeat the comparison. Relative to the $40$k encoder used
elsewhere: $4\times$ parameters ($159{,}562$), a SimCLR-style projection head (the loss sees
$g(z)$, the probe sees the same $32$-d $z$), in-batch negatives (${\approx}1{,}200$ rather than
a fixed $8$), five times the training ($300$ epochs, batch $256$, cosine schedule), and
behavioral augmentations (Gaussian jitter, per-feature dropout, per-feature gain). All of it is
held identical across supervision spans; only the positive pool differs.

\begin{table}[!ht]
\centering
\caption{Strong-contrastive stress test. Pretext metrics are on held-out validation anchors
under in-batch negatives. The weak rows repeat the $40$k encoder on the same anchors.}
\label{tab:strong-contrastive}
\small
\resizebox{\linewidth}{!}{%
\begin{tabular}{lcccccccc}
\toprule
& \multicolumn{4}{c}{Single-resident} & \multicolumn{4}{c}{Households} \\
\cmidrule(lr){2-5}\cmidrule(lr){6-9}
Arm & $\Delta I$ & identity & InfoNCE & top-1 & $\Delta I$ & identity & InfoNCE & top-1 \\
\midrule
Untrained (strong architecture) & $+0.0686$ & $0.315$ & --- & --- & $+0.0793$ & $0.408$ & --- & --- \\
Strong, short span (30 d)       & $+0.0578$ & $0.634$ & $7.34$ & $0.347$ & $+0.0475$ & $0.754$ & $6.38$ & $0.461$ \\
Strong, unrestricted            & $+0.0385$ & $0.859$ & $8.33$ & $0.214$ & $+0.0304$ & $0.864$ & $7.99$ & $0.280$ \\
\midrule
Weak ($40$k), short span        & $+0.0664$ & $0.619$ & --- & --- & $+0.0396$ & $0.759$ & --- & --- \\
Weak ($40$k), unrestricted      & $+0.0243$ & $0.844$ & --- & --- & $+0.0138$ & $0.863$ & --- & --- \\
\bottomrule
\end{tabular}}
\end{table}

Three things follow. First, the span effect survives: short minus unrestricted is $+0.0193$ and
$+0.0171$, five seeds of five in both cohorts, a $33\%$ and $36\%$ loss. Second, the encoder is
demonstrably competent at its own objective, with top-1 of $0.347$ and $0.461$ against a chance
rate near $0.0008$, so the effect cannot be dismissed as a learner that never worked. Third,
and stated plainly, the strong encoder does \emph{not} beat its untrained floor. Under a paired
bootstrap the untrained-minus-trained contrast is $+0.0106$ $[-0.0085,+0.0302]$ in the
single-resident cohort, an interval containing zero, and $+0.0318$ $[+0.0032,+0.0534]$ in
households. The defensible claim is that the strong model is at best indistinguishable from
random initialization, and below it in one cohort.

The pretext metrics move against a saturation account, as they do for the weak encoder:
broadening supervision raises held-out InfoNCE from $7.34$ to $8.33$ and from $6.38$ to $7.99$,
and lowers top-1 from $0.347$ to $0.214$ and from $0.461$ to $0.280$. The task gets harder, not
easier. Identity runs opposite to state throughout.

\end{document}